\documentclass[10pt,twocolumn,letterpaper]{article}

\usepackage[pagenumbers]{cvpr} %
\usepackage[dvipsnames]{xcolor}

\usepackage{arydshln}  %
\usepackage{tikz}  %
\usepackage{makecell}  %
\usepackage{nicematrix}  %
\usepackage{algpseudocode}
\usepackage[linesnumbered,ruled,vlined]{algorithm2e}
\usepackage{layouts}
\usepackage{colortbl}
\usepackage{fp}
\usepackage{pgffor}
\usepackage{pgf}
\usepackage{dblfloatfix}
\usepackage{bm}

\usepackage{multirow}
\makeatletter
\newcommand*{\rom}[1]{\expandafter\@slowromancap\romannumeral #1@}
\makeatother

\definecolor{tabfirst}{rgb}{1, 0.7, 0.7} %
\definecolor{tabsecond}{rgb}{1, 0.85, 0.7} %
\definecolor{tabthird}{rgb}{1, 1, 0.7} %

\usepackage{svg}
\usepackage{booktabs}  %
\usepackage{float}     %
\usepackage{array}     %
\usepackage{caption}   %
\usepackage{graphicx}  %

\definecolor{first}{HTML}{8ABEE6} 
\definecolor{second}{HTML}{B8D2E6} 
\definecolor{third}{HTML}{DAE1E6} 
\definecolor{invalid}{HTML}{E6B8B8} 

\makeatletter \newcommand{\myhypertarget}[2]{\Hy@raisedlink{\hypertarget{#1}{#2}}} \makeatother

\makeatletter
\newcommand{\customlabel}[2]{%
   \protected@write \@auxout {}{\string \newlabel {#1}{{#2}{\thepage}{#2}{#1}{}} }%
   \myhypertarget{#1}{}%
}
\makeatother

\definecolor{cvprblue}{rgb}{0.21,0.49,0.74}
\usepackage[pagebackref,breaklinks,colorlinks,allcolors=cvprblue]{hyperref}

\def\paperID{11906} %
\def\confName{CVPR}
\def\confYear{2025}

\title{SILO: Solving Inverse Problems with Latent Operators}

\author{Ron Raphaeli, Sean Man,  Michael Elad\\
Technion - Israel Institute of Technology, Haifa, Israel
\\
{\tt\small \{ronraphaeli,sean.man,elad\}@cs.technion.ac.il}
}

\begin{document}
\maketitle
\begin{abstract}
Consistent improvement of image priors over the years has led to the development of better inverse problem solvers.
Diffusion models are the newcomers to this arena, posing the strongest known prior to date. 
Recently, such models operating in a latent space have become increasingly predominant due to their efficiency.
In recent works, these models have been applied to solve inverse problems.
Working in the latent space typically requires multiple applications of an Autoencoder during the restoration process, which leads to both computational and restoration quality challenges.
In this work, we propose a new approach for handling inverse problems with latent diffusion models, where a learned degradation function operates within the latent space, emulating a known image space degradation.
Usage of the learned operator reduces the dependency on the Autoencoder to only the initial and final steps of the restoration process, facilitating faster sampling and superior restoration quality.
We demonstrate the effectiveness of our method on a variety of image restoration tasks and datasets, achieving significant improvements over prior art.
\end{abstract}

\begin{figure}[ht]
  \centering
 
 \includegraphics[width=\linewidth]{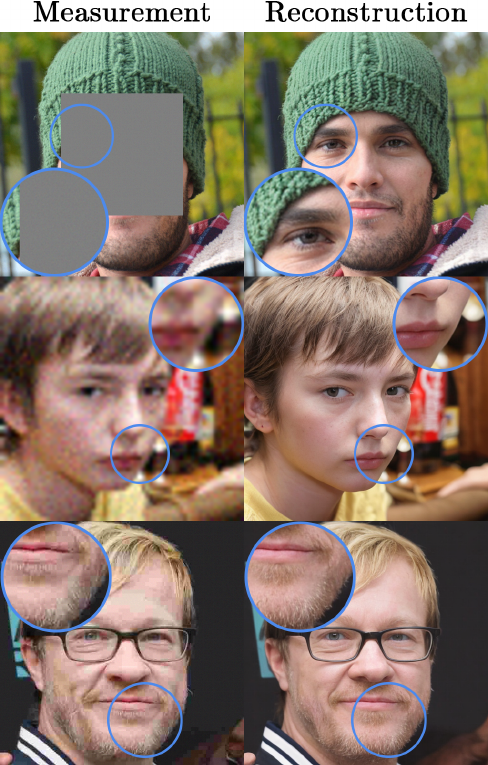}
\caption{Measurements and their corresponding reconstructions using our proposed latent inverse solver, SILO (\cref{alg:reconstruction}).}
\label{fig:first} 
\end{figure}

\begin{figure*}[ht]
  \centering
 
 \includegraphics[width=0.8\linewidth]{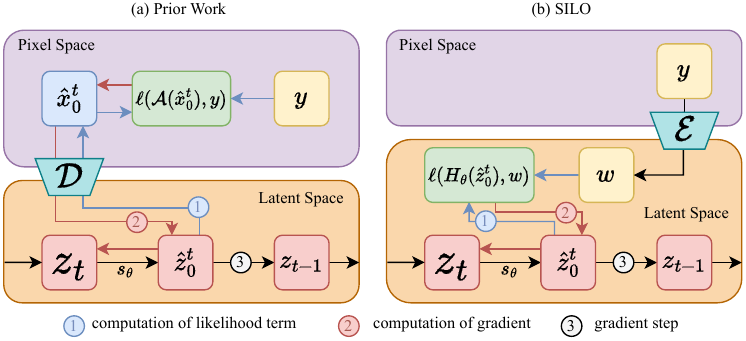}
\caption{\textbf{Computational schemes of prior work and SILO.}
(a): Prior work enforces consistency to the measurement in pixel space, resulting in differentiation through the decoder.
(b): SILO keeps all calculations in the latent space. %
This allows faster reconstructions while improving their perceptual quality compared to prior work, as seen in \cref{fig:ffhq comparison,tab:results_of_sr4_and_gb,tab:results_of_sr8_and_ip}.}
\label{fig:hero} 

\end{figure*}

\section{Introduction}

Methods for solving inverse problems aim to recover an unknown signal from its degraded measurements. Assuming that an image $x$ has been drawn from the probability density function $p(x)$, in this work we consider a degradation operator 
$\mathcal{A}(\cdot)$ which forms the given observation $y\sim\mathcal{N}(\mathcal{A}(x),\sigma_y^2I) $. Over the past several decades, countless techniques have been developed to tackle such inverse problems, adopting various directions.
Owing to these methods, it is possible to sharpen and denoise images \cite{TrainableNonlinearReactionchen2017,ImageDenoisingSparsedabov2007,ImageDenoisingSparseelad2006,ImageDenoisingDeepelad2023,WeightedNuclearNormgu2014,ResidualDenseUNetgurrola-ramos2021}, accelerate CT scans \cite{AdaptiveCompressedSensingelata2025,SolvingInverseProblemssong2021}, and get information on the atmosphere's composition \cite{InverseMethodsAtmosphericrodgers2000}.

Most inverse problems are ill-posed, not having a unique solution, and the majority of the solutions do not correspond with natural images. An appealing approach to address these challenges is to harness methods that can sample from the prior $p(x)$, and adapt them to sample from the posterior distribution $x\sim p(x|y)$ \cite{DiffusionPosteriorSamplingchung2022,ZeroShotImageRestorationwang2022,DenoisingDiffusionRestorationkawar2022,HighPerceptualQualityohayon2021,HighPerceptualQualityJPEGman2023}.
If sampled correctly, we recover a signal that is simultaneously in the distribution of real-world signals while being consistent with the measurement $y$. %

Diffusion models \cite{GenerativeModelingEstimatingsong2019a,ScoreBasedGenerativeModelingsong2020,DenoisingDiffusionProbabilisticho2020,DiffusionModelsBeatdhariwal2021,DeepUnsupervisedLearningsohl-dickstein2015,ImprovingImageGenerationbetker} learn the score function of the prior, allowing to sample from it faithfully.
These models are responsible for the significant advancement in generative methods in recent years, enabling the generation of imaginary photos of scenes that do not exist \cite{DiffusionModelsBeatdhariwal2021,HighResolutionImageSynthesisrombach2022,PhotorealisticTextImageDiffusionsaharia2022}.
Unfortunately, sampling images with a diffusion model is relatively slow.
Latent diffusion models (LDM) ~\cite{HighResolutionImageSynthesisrombach2022} aim to resolve this issue by moving the diffusion process to a latent space of a lower dimension, thereby enabling faster score computation.
At the end of the sampling, a decoder is used to convert the latent vector into an image in the pixel space. This decoder is part of a pre-trained Autoencoder \cite{AutoEncodingVariationalBayeskingma2022} that LDMs employ. 

Works that aim to solve inverse problems using latent diffusion, such as PSLD~\cite{SolvingLinearInverserout2023} and ReSample~\cite{SolvingInverseProblemssong2023} benefit from these models' speed and the fact they are usually trained on larger and possibly more diverse datasets.
The latter fact allows the model to handle general natural images instead of focusing on specific domains (e.g., face images). 
Alongside these clear benefits of leaning on LDM, the challenge is that while the diffusion process occurs fully in the latent space, the information needed to enforce consistency to the observation lies in the measurement (pixel-based) space.
The prevailing solution involves repeatedly using and differentiating through the Autoencoder during sampling. This leads to weaknesses such as slow sampling as well as blurry and noisy reconstructions.
Prior works that address inverse problems via LDMs tried to mitigate these symptoms by proposing different regularizations~\cite{SolvingLinearInverserout2023,SolvingInverseProblemssong2023,FirstOrderTweedieSolvingrout2024,PrompttuningLatentDiffusionchung2023,RegularizationTextsLatentkim2024}, achieving mixed results.

This work eliminates the repeated use of the Autoencoder, transferring the entire restoration process to the latent space.
To do so, we suggest learning a latent operator as a small neural network that emulates known degradations, and migrate them from the image-space to the latent-space.
By eliminating the need to use the decoder for consistency guidance.
The Autoencoder is used only twice, regardless of the number of diffusion steps.
In this way, the guidance and sampling steps are done in the latent space (\cref{fig:hero}), and the Autoencoder is used to encode the observation and decode the final restoration to the image space.
Our key contributions are the following:

\begin{enumerate}

\item We propose a novel approach for solving inverse problems with LDMs without using the Autoencoder during the diffusion process.

\item This approach accelerates the restoration process by eliminating the multiple uses of the Autoencoder and its differentiation.

\item Our method achieves state-of-the-art results in various inverse problems and datasets.

\end{enumerate}

\section{Background}

\subsection{Inverse problems}
Inverse problems generally refer to the task of recovering a clean and probable signal that correlates to observed degraded measurement.
Specifically, our problem can be described as
\begin{equation}\label{eq:general_inverse_problem}
    y = \mathcal{A}(x) + v
\end{equation}
where $x\in \mathbb{R}^d$ is the unknown clean image, $ \mathcal{A}(\cdot):\mathbb{R}^d\rightarrow\mathbb{R}^n$ is a degradation operator (\eg Gaussian blur) $v$ is noise sampled from a normal distribution, $\mathcal{N}(0,\sigma_y^2I)$, with variance $\sigma_y^2$ and $y\in\mathbb{R}^n$ is the observed measurement. 

Due to the problem's ill-posed nature, some assumptions must be made. In our case of recovering degraded images, we assume prior knowledge on the behavior of real-world (natural) images.
Over several decades, many methods have been proposed to incorporate our understanding of the properties of images to solve inverse problems. %
Nowadays, methods that leverage deep learning achieve state-of-the-art results \cite{RestoreFormerHighQualityBlindwang2022,SwinIRImageRestorationliang2021,ESRGANEnhancedSuperResolutionwang2018,FlexibleBlindJPEGjiang2021,GaussianDenoiserResidualzhang2017,TrainableNonlinearReactionchen2017,VariationalPerspectiveSolvingmardani2023}.
The connection between inverse problems and machine learning is natural, as we can aim to learn the prior of natural images using the enormous amount of available data.
Given a robust prior, one can recover clean images from degraded ones using techniques such as plug-and-play \cite{LittleEngineThatromano2017,PlugPlayADMMImagechan2017,PlugPlayPriorsModelvenkatakrishnan2013,DenoisingDiffusionModelszhu2023}, score-based \cite{DiffusionPosteriorSamplingchung2022,DenoisingDiffusionRestorationkawar2022,VariationalPerspectiveSolvingmardani2023,DenoisingDiffusionModelszhu2023,DecomposedDiffusionSamplerchung2023}, sparsity-based \cite{KSVDAlgorithmDesigningaharon2006,SparsityBasedPoissongiryes2012,SingleImageScaleUsingzeyde2012}, and more.

\subsection{Diffusion models}
Diffusion models sample from $p(x)$ by applying a series of activations of pretrained minimum mean squared error (MMSE) denoisers \cite{ImageDenoisingDeepelad2023}. 
Specifically, following DDPM's notations~\cite{DenoisingDiffusionProbabilisticho2020}, a forward process is first defined as the creation of a sequence of noisy images $x_t$ for timesteps \{$t=0,1,\ldots,T$\}  via
\begin{equation}\label{eq:ddpm_xt}
    x_t\sim \mathcal{N}(x_t;\sqrt{\bar{\alpha_t}}x_0,(1-\bar{\alpha_t})I),
\end{equation}
where $x_0$ is a clean image, $\bar{\alpha_t}=\Pi_{s=1}^t\alpha_s$,  $\alpha_t:=1-\beta_t$ and $\{\beta_s\}^T_{s=1}$ is a chosen variance schedule.
Generating images with diffusion models is achieved by reversing the above forward process. This is done via the recursive relation \cite{ScoreBasedGenerativeModelingsong2020}
\begin{equation}\label{eq:DDPM_score}
    x_{t-1}=\frac{1}{\sqrt{\alpha_t}}\left(x_t+\beta_t\nabla_{x_t} \ln{p(x_t)}\right)+\sqrt{\beta_t}u,
\end{equation}
where $u\sim\mathcal{N}(0,I)$ is a noise perturbation.
The term $s_\theta (x_t) := \nabla_{x_t} \ln{p(x_t)}$ is known as the score function~\cite{ScoreBasedGenerativeModelingsong2020,GenerativeModelingEstimatingsong2019,ConnectionScoreMatchingvincent2011}, 
obtained via Tweedie's formula \cite{TweediesFormulaSelectionefron2011}: 
\begin{equation}\label{tweedie}
    \nabla_{x_t} \ln{p(x_t)}=\frac{1}{1-\bar{\alpha_t}}(-x_t+\sqrt{\bar{\alpha_t}}\mathbb{E}_{x_t\sim p(x_t)}[x_0|x_t]).
\end{equation}
The expression $\mathbb{E}_{x_t\sim p(x_t)}[x_0|x_t]$ is nothing but the MMSE estimate of $x_0$ given $x_t$, i.e. an image denoiser that aims to remove white additive Gaussian noise while striving to get the smallest $L_2$ error.
This denoiser is formed as a learned neural network, being the only portion of the process that relies on training.
The common practice is to train a network $\epsilon_\theta(x_t,t)$, which predicts the noise added to create $x_t$, conditioned on the timestep $t$. Thus, given $\epsilon_\theta(x_t,t)$, the synthesis process amounts to the update step \cite{DenoisingDiffusionProbabilisticho2020}  
\begin{equation}\label{eq:ddpm_xt-1}
    x_{t-1}=\frac{1}{\sqrt{\alpha_t}}\left(x_t-\frac{\beta_t}{\sqrt{1-\bar{\alpha_t}}}\epsilon_\theta(x_t,t)\right)+\sqrt{\beta_t}u,
\end{equation}
initialized from $x_T\sim\mathcal{N}(0,I)$ and iteratively progresses towards $x_0$, a sample from $p(x)$.

\subsection{Latent diffusion}
Sampling $x\sim p(x)$ with a diffusion model is tedious, as it requires multitude ($50-1000$, depending on the approximation scheme) of activations of the denoiser network, following the iterative manner of equation \cref{eq:ddpm_xt-1}.
To mitigate this problem, it was suggested to perform the diffusion process in a lower dimensional latent space \cite{HighResolutionImageSynthesisrombach2022}.
In this approach, an Autoencoder is used to encode images to the latent space and decode latent vectors (latents for short) back to the image space as follows: 
\begin{equation}\label{eq:Autoencoder}
z=\mathcal{E}(x),\quad x\approx\mathcal{D}(z)\space,
\end{equation}
where $x$ is an image, $z$ is a latent, $\mathcal{E}(\cdot):\mathbb{R}^d\rightarrow\mathbb{R}^k $ is the encoder %
and $\mathcal{D}(\cdot):\mathbb{R}^k\rightarrow\mathbb{R}^d$ is the decoder. %
In our case, $k \ll d$ implies that the diffusion process occurs in the lower dimension latent space, and each network pass is thus much faster, leading to more efficient training and sampling. 
Since the diffusion process is performed on latents $z$, instead of images $x$, the creation of noisy latents is adapted from \cref{eq:ddpm_xt}, transforming to
\begin{equation}\label{eq:ddpm_zt}
    z_t\sim \mathcal{N}(z_t;\sqrt{\bar{\alpha_t}}z_0,(1-\bar{\alpha_t})I),
\end{equation}
where $z_0$ is a clean latent.
In the latent domain, the reverse process takes the form of \cref{eq:ddpm_xt-1}, and translates to
\begin{equation}\label{eq:ddpm_zt-1}
    z_{t-1}=\frac{1}{\sqrt{\alpha_t}}\left(z_t-\frac{\beta_t}{\sqrt{1-\bar{\alpha_t}}}\epsilon_\theta(z_t,t)\right)+\sqrt{\beta_t}u_z.
\end{equation}

\section{LDM for inverse problems: related work}
To sample from the posterior, $x\sim p(x|y)$, one need to use the conditional score $\nabla_{x_t} \ln{p(x_t|y})$.
When migrating to the latent space, this score becomes $\nabla_{z_t} \ln{p(z_t|y})$.
Following Bayes rule we get
\begin{equation}
    \nabla_{z_t} \ln{p(z_t|y})=\nabla_{z_t} \ln{p(z_t})+\nabla_{z_t} \ln{p(y|z_t}).
\end{equation}
The prior score function (first term in RHS) is obtained from the latent diffusion model.
On the other hand, the score likelihood term (second term in RHS) involve calculations in the image space, since the degradation operator $\mathcal{A}$ is applied to images.
For example, DPS~\cite{DiffusionPosteriorSamplingchung2022}, which uses image space diffusion, relies on the approximation
\begin{equation}\label{eq:DPS}
    \nabla_{x_t} \ln{p(y|x_t)} \approx -\frac{1}{\sigma_y^2}\nabla_{x_t} \|y-\mathcal{A}(\hat{x}^t_0)\|_2^2 ,
\end{equation}
where $\hat{x}^t_0 = \mathbb{E}_{x_t\sim p(x_t)}[x_0|x_t]$.
This expression implies that calculating the update term within each step of the diffusion process requires a differentiation of both the degradation operator and the denoiser. 
If this is to be adjusted to operate in the latent space, the naive approach would be to decode the latent at each step and calculate the score likelihood approximation as follows:
\begin{equation}\label{eq:LDPS}
    \nabla_{z_t} \ln{p(y|z_t)} \approx -\frac{1}{\sigma_y^2}\nabla_{z_t} \|y-\mathcal{A}(\mathcal{D}(\hat{z}^t_0))\|_2^2 ,
\end{equation} where,
\begin{equation} \label{eq:mmse_denoise_z}
    \hat{z}^t_0 = \mathbb{E}_{z_t\sim p(z_t)}[z_0|z_t] = \frac{\left( z_t - \sqrt{1 - \bar{\alpha}_t} \epsilon_\theta (z_t,t) \right)}{\sqrt{\bar{\alpha}_t}}.
\end{equation}
This pulls the reconstruction to be consistent with the measurement in the pixel space, resulting in differentiating with respect to the latent $z_t$ through the decoder.
In practice, this approach leads to blurry and noisy reconstructions due to the reliance on the Autoencoder and its gradients~\cite{SolvingInverseProblemssong2023}. 
Recall that the Autoencoder is trained to decode clean and real latents $z_0$.
However, especially during the early stages of the restoration, the latents, $\hat{z}^t_0$, differ from that condition, and thus we decode latents that are out-of-distribution to the training data of the Autoencoder.
Moreover, backpropagating through a large neural network can produce noisy gradients, corrupting the information needed for the reconstruction. %
Previous work approached these difficulties in different ways, all heavily relying on the decoder in one way or another. In LDPS~\cite{SolvingInverseProblemssong2023}, every diffusion step is followed by \cref{eq:LDPS}.
GML-DPS~\cite{SolvingLinearInverserout2023} adds a projection step,
\begin{equation}
    \label{eq:gml}
    \nabla_{z_t} \|\hat{z}^t_0-\mathcal{E}(\mathcal{D}(\hat{z}^t_0))\|_2^2 ,
\end{equation}
that pulls the latent vector towards a fixed point of the encoder-decoder. PSLD~\cite{SolvingLinearInverserout2023} performs orthogonal projection, which changes the step in \cref{eq:gml} to
\begin{equation}
    \label{eq:psld}
    \nabla_{z_t} \|\hat{z}^t_0-\mathcal{E}(\mathcal{A}^Ty+(I-\mathcal{A}^T\mathcal{A})\mathcal{D}(\hat{z}^t_0))\|_2^2.
\end{equation}
Taking a somewhat different approach, ReSample \cite{SolvingInverseProblemssong2023} solves the optimization problem 
\begin{equation}
    \label{eq:resample_optimization}
   \hat{z}^*(y) \in \underset{z}{\arg\min}{\|y-\mathcal{A}(\mathcal{D}(z)\|_2^2}
\end{equation}
during reconstruction, and adds noise to the solution to create the next latent for the diffusion process.
STSL~\cite{FirstOrderTweedieSolvingrout2024} stabilizes and accelerates the conditional sampling process by combining LDPS steps with Tweedie's second-order approximation.
A separate line of work regularizes the restoration process by harnessing the fact that LDMs are commonly text-conditioned.
In P2L \cite{PrompttuningLatentDiffusionchung2023}, the text embedding is updated during reconstruction, providing another way to promote consistency.
Similarly, TReg~\cite{RegularizationTextsLatentkim2024} updates the negative text embedding to minimize the similarity between it and the CLIP encoding \cite{LearningTransferableVisualradford2021} of the restored image.

\section{Proposed method}\label{sec:method}

Since the diffusion process and the measurement operator, $\mathcal{A}$, operate in different domains, there is no way to avoid using the Autoencoder completely.
There are two plausible ways to use it:
(1) We could use the decoder during the diffusion process to decode latents and use the known operator $\mathcal{A}$ in the pixel space.
Alternatively, (2) we could bring the measurement and degradation operator to the latent space.
The first paradigm has been employed in all previous works in the field.
In the Appendix, we demonstrate two flaws in using the decoder and differentiating through it.
These flaws manifest as artifacts, as seen in \cref{sec:experiments} and as reported in ReSample (Appendix D in \cite{SolvingInverseProblemssong2023}).

In the second suggested paradigm, which is the one we propose in this work, the measurement, $y$, is encoded to the latent space, naively by $w=\mathcal{E}(y)$.
Additionally, a learned operator, $H_\theta$, needs to mimic the degradation operator $\mathcal{A}$ while operating entirely in the latent domain. This suggestion gives rise to four critical questions:

\noindent \customlabel{q1}{\textbf{Q1}}\textbf{Q1:} Clearly, $y$ is not a natural high-quality image. Why would it be allowed to apply the encoder $\mathcal{E}$ onto it?

\noindent \customlabel{q2}{\textbf{Q2}}\textbf{Q2:} What are our requirements from the operator $H_\theta$, and how do we promote consistency to the measurement?

\noindent \customlabel{q3}{\textbf{Q3}}\textbf{Q3:} How should the operator $H_\theta$ be learned?

\noindent \customlabel{q4}{\textbf{Q4}}\textbf{Q4:} How can this operator aid in restoration?

\noindent In the following subsections, we will explore and answer these questions in detail.

\subsection{Encoding the measurement}\label{sec:encoding_the_measurement}

\begingroup
\newcommand{\centered}[1]{\begin{tabular}{l} #1 \end{tabular}}
\def\totalwidth{0.5}
\def\numdegredations{2}
\def\nummetrics{4}
\FPeval{\totalcolumns}{round(\numdegredations*\nummetrics ,0)}
\FPeval{\colwidth}{\totalwidth/\totalcolumns}

\begin{table}[b]
    \centering
    \setlength{\tabcolsep}{4pt}
    \small
    \begin{tabular}{l*{\totalcolumns}{c}}
    
    Degradation & $x, f(x)$ & $y_{nl}, f(y_{nl})$ & $y, f(y)$ & $y_{nl}, f(y)$ \\
    \midrule
    Gaussian blur        &\multirow{2}{*}{\rule[-2ex]{0.4pt}{5ex}}& 46.03 & 37.57 & 40.93 \\
    SR $\times8$      && 45.72 & 43.25 & 40.31 \\
    SR $\times4$      &31.23& 40.81 & 37.90 & 38.22 \\
    Inpaint   &\multirow{2}{*}{\rule[-2ex]{0.4pt}{5ex}}& 32.23 & 31.23 & 31.89 \\
    JPEG && 32.04 & 30.81 & 31.36 \\
    \end{tabular}
    
    \caption{
    PSNR values for four types of image pairs with various degradations.
    The settings are consistent with those in \cref{sec:Degradations}.
    We see the loss of information when encoding clean and degraded images to the latent space (full analysis in \cref{sec:encoding_the_measurement}).
    }
    \label{tab:encode_decode_ablation}
\end{table}

Referring to \ref{q1},
we acknowledge that computing $w$ for a degraded image requires using the encoder on an image that is out-of-distribution relative to its training data\footnote{While our choice is to use $w=\mathcal{E}(y)$, future work may consider better ways to compute $w$, possibly considering learned alternatives.}.
To gain confidence that this is a valid step to take, we first investigate how much the encoding-decoding process changes an image.
Specifically, we compare the PSNR values of images and their encoded-decoded versions by applying $f(\cdot)=\mathcal{D}(\mathcal{E}(\cdot))$, average over 1000 images using various degradations and include the results in \cref{tab:encode_decode_ablation}. The full experiment settings are described in \cref{sec:experiments}.
From the results, we see that for noiseless degraded images, $y_{nl} = \mathcal{A}(x) $, the PSNR with their decoded-encoded counterpart, $f(y_{nl})$, is higher than for natural images.
In addition, a denoising effect occurs when applying $f$ on noisy measurements, $y$, as for the majority of the tested degradations, we observe $\text{PSNR}(y_{nl},f(y)) > \text{PSNR}(y,f(y))$.
This reassures us that applying the encoder on degraded images should not be a major bottleneck in our method.

\subsection{Score likelihood in latent space}
Moving to \ref{q2}, it is necessary to understand how the approximation of the score likelihood in \cref{eq:DPS} can be transformed to allow sampling in the latent space alone. 
Consider a measurement $y^*$ created from an unknown signal $x^*$ using \cref{eq:general_inverse_problem}.
Assuming that the Autoencoder enables a  near-perfect reconstruction on degraded images, $\mathcal{D}(\mathcal{E}(y)) \cong y$, 
we can write
\begin{equation}\label{eq:bound_step_1}
    \| y^*-\mathcal{A}(x)\|^2_2 \cong \| \mathcal{D}(\mathcal{E}(y^*)) -\mathcal{D}(\mathcal{E}(\mathcal{A}(x)))\|^2_2.
\end{equation}
To provide an answer to \ref{q2}, suppose we have access to a trained operator $H_\theta (z) = \mathcal{E}(\mathcal{A}(x))$, 
where $z=\mathcal{E}(x)$. This operator aims to mimic the degradation $\mathcal{A}(x)$ while operating fully in the latent domain. 
Then, \cref{eq:bound_step_1} can be rewritten as: 
\begin{eqnarray}\label{eq:bound_step_2}
    \| y^*-\mathcal{A}(x)\|^2_2 & \cong &  \| \mathcal{D}(\mathcal{E}(y^*)) -\mathcal{D}(\mathcal{E}(\mathcal{A}(x)))\|^2_2 \nonumber \\  & \cong & \| \mathcal{D}(\mathcal{E}(y^*)) -\mathcal{D}(H_\theta (z))\|^2_2.
\end{eqnarray}
Since $\mathcal{D}$ is differentiable, it is also Lipschitz continuous,
therefore, a constant $C$ exists\footnote{A redesign of the Autoencoder to lower the value of $C$ may have a positive impact on the overall performance of LDM in general, and for our proposed inverse problem solver in particular. This is left for future work.} such that
\begin{equation}\label{eq:bound_step_3}
\| \mathcal{D}(\mathcal{E}(y^*)) -\mathcal{D}(H_\theta (z))\|^2_2 \leq C \| \mathcal{E}(y^*) - H_\theta (z)\|^2_2.
\end{equation}
Ultimately, since the RHS in \cref{eq:bound_step_3} bounds the LHF in \cref{eq:bound_step_1}, we can minimize it as a proxy to \cref{eq:DPS}, yielding, 
\begin{equation}\label{eq:SILO_step}
    \nabla_{z_t} \ln{p(y|z_t)} \approx -\frac{Const}{\sigma_y^2}\nabla_{z_t} \|w-H_\theta(\hat{z}^t_0)\|_2^2 ,
\end{equation}
where $w=\mathcal{E}(y^*)$. This motivates us to design such $H_\theta$, that will allow for reconstructions using \cref{eq:SILO_step}.

\begingroup
\def\totalwidth{1}
\def\numdegredations{2}
\def\nummetrics{4}
\FPeval{\totalcolumns}{round(\numdegredations*\nummetrics +1,0)}
\FPeval{\colwidth}{\totalwidth/\totalcolumns}

\newcommand{\rankhl}[4]{%
  \ifnum\numexpr\pdfstrcmp{#1}{#2}=0
    \cellcolor{first}#1%
  \else\ifnum\numexpr\pdfstrcmp{#1}{#3}=0
    \cellcolor{second}#1%
  \else\ifnum\numexpr\pdfstrcmp{#1}{#4}=0
    \cellcolor{third}#1%
  \else
    #1%
  \fi\fi\fi
}

\newcommand{\rankhlrev}[4]{%
  \ifnum\numexpr\pdfstrcmp{#1}{#4}=0
    \cellcolor{first}#1%
  \else\ifnum\numexpr\pdfstrcmp{#1}{#3}=0
    \cellcolor{second}#1%
  \else\ifnum\numexpr\pdfstrcmp{#1}{#2}=0
    \cellcolor{third}#1%
  \else
    #1%
  \fi\fi\fi
}

\edef\firstTime{148}\edef\secondTime{149}\edef\thirdTime{331}
\edef\firstPSNRa{28.74}\edef\secondPSNRa{28.63}\edef\thirdPSNRa{28.00}
\edef\firstCPSNRa{50.66}\edef\secondCPSNRa{44.18}\edef\thirdCPSNRa{44.07}
\edef\firstLPIPSa{0.253}\edef\secondLPIPSa{0.236}\edef\thirdLPIPSa{0.222}
\edef\firstFIDa{30.33}\edef\secondFIDa{29.61}\edef\thirdFIDa{28.34}
\edef\firstKIDa{10.74}\edef\secondKIDa{9.68}\edef\thirdKIDa{8.21}

\edef\firstPSNRb{29.34}\edef\secondPSNRb{29.06}\edef\thirdPSNRb{28.23}
\edef\firstCPSNRb{42.74}\edef\secondCPSNRb{38.76}\edef\thirdCPSNRb{36.69}
\edef\firstLPIPSb{0.247}\edef\secondLPIPSb{0.200}\edef\thirdLPIPSb{0.182}
\edef\firstFIDb{29.63}\edef\secondFIDb{26.51}\edef\thirdFIDb{23.82}
\edef\firstKIDb{9.05}\edef\secondKIDb{7.34}\edef\thirdKIDb{5.10}

\begin{table*}[!b]
    \centering
    \setlength{\tabcolsep}{4pt}
    \small
    \begin{tabular}{l *{\totalcolumns}{c}}
        \multicolumn{2}{c}{} &
	    \multicolumn{\nummetrics}{c}{Gaussian blur} &
	    \multicolumn{\nummetrics}{c}{Super-Resolution $\times4$} 
	    \\
        
        \cmidrule(lr){3-6}
        \cmidrule(lr){7-10}
        
        Method &
        Time [sec]&
        PSNR &
        LPIPS &
        FID &
        KID &
        PSNR &
        LPIPS &
        FID&
        KID 
        \\

        \hline
        
        Ours (RV)& 
        \rankhl{149}{\firstTime}{\secondTime}{\thirdTime}&
        \rankhl{26.70}{\firstPSNRa}{\secondPSNRa}{\thirdPSNRa}&
        \rankhlrev{0.222}{\firstLPIPSa}{\secondLPIPSa}{\thirdLPIPSa}&
        \rankhlrev{28.34}{\firstFIDa}{\secondFIDa}{\thirdFIDa}&
        \rankhlrev{8.21}{\firstKIDa}{\secondKIDa}{\thirdKIDa}&
        
        \rankhl{27.03}{\firstPSNRb}{\secondPSNRb}{\thirdPSNRb}&
        \rankhlrev{0.182}{\firstLPIPSb}{\secondLPIPSb}{\thirdLPIPSb}&
        \rankhlrev{23.82}{\firstFIDb}{\secondFIDb}{\thirdFIDb}&
        \rankhlrev{5.10}{\firstKIDb}{\secondKIDb}{\thirdKIDb}
        \\

        Ours (SD)& 
        \rankhl{148}{\firstTime}{\secondTime}{\thirdTime}&
        \rankhl{26.55}{\firstPSNRa}{\secondPSNRa}{\thirdPSNRa}&
        \rankhlrev{0.236}{\firstLPIPSa}{\secondLPIPSa}{\thirdLPIPSa}&
        \rankhlrev{30.33}{\firstFIDa}{\secondFIDa}{\thirdFIDa}&
        \rankhlrev{9.68}{\firstKIDa}{\secondKIDa}{\thirdKIDa}&
        
        \rankhl{26.95}{\firstPSNRb}{\secondPSNRb}{\thirdPSNRb}&
        \rankhlrev{0.200}{\firstLPIPSb}{\secondLPIPSb}{\thirdLPIPSb}&
        \rankhlrev{26.51}{\firstFIDb}{\secondFIDb}{\thirdFIDb}&
        \rankhlrev{7.34}{\firstKIDb}{\secondKIDb}{\thirdKIDb}
        \\

        ReSample& 
        \rankhl{1418}{\firstTime}{\secondTime}{\thirdTime}&
        \rankhl{27.92}{\firstPSNRa}{\secondPSNRa}{\thirdPSNRa}&
        \rankhlrev{0.253}{\firstLPIPSa}{\secondLPIPSa}{\thirdLPIPSa}&
        \rankhlrev{29.61}{\firstFIDa}{\secondFIDa}{\thirdFIDa}&
        \rankhlrev{10.74}{\firstKIDa}{\secondKIDa}{\thirdKIDa}&
        
        \rankhl{24.62}{\firstPSNRb}{\secondPSNRb}{\thirdPSNRb}&
        \rankhlrev{0.433}{\firstLPIPSb}{\secondLPIPSb}{\thirdLPIPSb}&
        \rankhlrev{45.02}{\firstFIDb}{\secondFIDb}{\thirdFIDb}&
        \rankhlrev{25.50}{\firstKIDb}{\secondKIDb}{\thirdKIDb}
        \\
        PSLD& 
        \rankhl{390}{\firstTime}{\secondTime}{\thirdTime}&
        \rankhl{28.63}{\firstPSNRa}{\secondPSNRa}{\thirdPSNRa}&
        \rankhlrev{0.288}{\firstLPIPSa}{\secondLPIPSa}{\thirdLPIPSa}&
        \rankhlrev{38.44}{\firstFIDa}{\secondFIDa}{\thirdFIDa}&
        \rankhlrev{12.23}{\firstKIDa}{\secondKIDa}{\thirdKIDa}&
        \rankhl{28.23}{\firstPSNRb}{\secondPSNRb}{\thirdPSNRb}&
        \rankhlrev{0.249}{\firstLPIPSb}{\secondLPIPSb}{\thirdLPIPSb}&
        \rankhlrev{29.63}{\firstFIDb}{\secondFIDb}{\thirdFIDb}&
        \rankhlrev{10.11}{\firstKIDb}{\secondKIDb}{\thirdKIDb}
        \\

        GML-DPS& 
        \rankhl{389}{\firstTime}{\secondTime}{\thirdTime}&
        \rankhl{28.74}{\firstPSNRa}{\secondPSNRa}{\thirdPSNRa}&
        \rankhlrev{0.309}{\firstLPIPSa}{\secondLPIPSa}{\thirdLPIPSa}&
        \rankhlrev{42.68}{\firstFIDa}{\secondFIDa}{\thirdFIDa}&
        \rankhlrev{16.58}{\firstKIDa}{\secondKIDa}{\thirdKIDa}&
        \rankhl{29.34}{\firstPSNRb}{\secondPSNRb}{\thirdPSNRb}&
        \rankhlrev{0.247}{\firstLPIPSb}{\secondLPIPSb}{\thirdLPIPSb}&
        \rankhlrev{30.71}{\firstFIDb}{\secondFIDb}{\thirdFIDb}&
        \rankhlrev{9.05}{\firstKIDb}{\secondKIDb}{\thirdKIDb}
        \\
        
        LDPS& 
        \rankhl{331}{\firstTime}{\secondTime}{\thirdTime}&
        \rankhl{28.00}{\firstPSNRa}{\secondPSNRa}{\thirdPSNRa}&
        \rankhlrev{0.327}{\firstLPIPSa}{\secondLPIPSa}{\thirdLPIPSa}&
        \rankhlrev{47.38}{\firstFIDa}{\secondFIDa}{\thirdFIDa}&
        \rankhlrev{19.95}{\firstKIDa}{\secondKIDa}{\thirdKIDa}&
        \rankhl{29.06}{\firstPSNRb}{\secondPSNRb}{\thirdPSNRb}&
        \rankhlrev{0.281}{\firstLPIPSb}{\secondLPIPSb}{\thirdLPIPSb}&
        \rankhlrev{34.44}{\firstFIDb}{\secondFIDb}{\thirdFIDb}&
        \rankhlrev{11.69}{\firstKIDb}{\secondKIDb}{\thirdKIDb}
        \\

    \end{tabular}
    \caption{Comparison of inverse problem solvers using latent diffusion on the FFHQ dataset.
    We mark the \colorbox{first}{first}, \colorbox{second}{second}, and \colorbox{third}{third} best-performing methods in each metric.
    }
    \label{tab:results_of_sr4_and_gb}
\end{table*}
\endgroup

\subsection{The latent degradation  operator}

According to the assumptions that led to \cref{eq:SILO_step}, $H_\theta$ should be trained in a way that approximates $H_\theta (z) \approx \mathcal{E}(y)$.
Thus, answering \ref{q3}, we train the operator $H_\theta$ with the following loss:
\begin{equation}\label{eq:loss}
    \mathcal{L}=\mathbb{E}_{
    \substack{
    x\sim p(x) \\
    y|x\sim\mathcal{N}(\mathcal{A}(x),\sigma_y^2I) \\
    t\sim \mathcal{U}_{[0,T]}}}
    [\|H_\theta(\hat{z}^t_0,t)-\mathcal{E}(y)\|_1].
\end{equation}
The expectation sweeps through ideal images $x\sim p(x)$, and creates $y$ from them according to \cref{eq:general_inverse_problem}.  $\hat{z}^t_0$ is created from the same images $x$ after encoding, adding noise and denoising (\cref{eq:Autoencoder,eq:ddpm_zt,eq:mmse_denoise_z}).
The computation graph for the training of $H_\theta$ is presented in \Cref{fig:training}.
We refer to $H_\theta$ as a learned degradation operator since for $t=0$, $H_\theta$ gets a clean latent, $\mathcal{E}(x)$, and outputs the encoding of the degraded image, $\mathcal{E}(y)$.
Additional information on the design and nature of $H_\theta$ is provided in \cref{sec:Degradations} and Appendix.

\begin{figure}[tb]
  \centering
 
 \includegraphics[width=\linewidth]{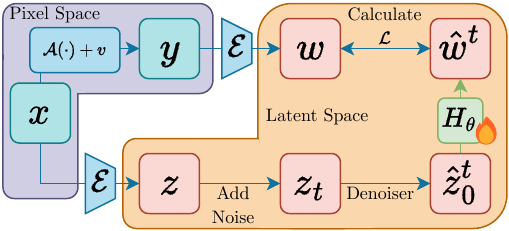}
 
\caption{\textbf{Training scheme of the latent operator,} $\mathbf{H_\theta}$.
In training, gradients flow from $\mathcal{L}$ to update the parameters of $H_\theta$. Note that no gradients pass through the pixel space.
$H_\theta$ learns to mimic the effect of the degradation operator in the latent space, allowing us to use SILO to solve inverse problems using LDMs.}
\label{fig:training} 
\end{figure}

\subsection{Reconstruction}

We are left with \ref{q4}, questioning the way to deploy the trained operator $H_\theta$ in solving inverse problems. \cref{alg:reconstruction} describes the proposed recovery scheme, termed \textbf{SILO} (\textbf{S}olving \textbf{I}nverse Problems with \textbf{L}atent \textbf{O}perators).
Following the ablation study in DPS (Appendix C in \cite{DiffusionPosteriorSamplingchung2022}), we use a term similar to 
\cref{eq:SILO_step} in step 9 of the algorithm, taking a gradient of the square root of the RHS in \cref{eq:bound_step_3}.
Note that in \cref{alg:reconstruction}, the decoder and encoder are each used once.
Furthermore, gradients are never calculated through the decoder or encoder during sampling; they are only calculated through the denoiser and the learned operator, $H_\theta$ in the latent space.

\begin{algorithm}[hbt]
    \caption{SILO: Reconstruction Algorithm}\label{alg:reconstruction}
\KwData{measurement $y$, encoder $\mathcal{E}$, decoder $\mathcal{D}$, latent diffusion model $\epsilon_\theta$, trained degradation operator $H_\theta$, text condition $\mathcal{C}$, consistency scale $\eta$, noise schedule $\{\beta_t\}_{t=0}^T$}
\KwResult{A reconstruction $\hat{x}$}

$z_T \sim \mathcal{N}(0, I)$\;
Encoding: $w = \text{clamp}(\mathcal{E}(y)$,-4,4)\;
\For{$t = T$ to $1$}{
    
    $\hat{\epsilon} \leftarrow \epsilon_\theta (z_t,t,\mathcal{C}) $\;
    
    $\hat{z}^t_0 \leftarrow \frac{1}{\sqrt{\bar{\alpha}_t}} \left( z_t - \sqrt{1 - \bar{\alpha}_t} \hat{\epsilon} \right)$\;
    
    $n \sim \mathcal{N}(0, I)$\;
    
    $z'_{t-1} \leftarrow \frac{\sqrt{\alpha_t}(1 - \bar{\alpha}_{t-1})}{1 - \bar{\alpha}_t} z_t + \frac{\sqrt{\bar{\alpha}_{t-1}} \beta_t}{1 - \bar{\alpha}_t} \hat{z}^t_0 + \sqrt{\frac{1 - \bar{\alpha}_{t-1}}{1 - \bar{\alpha}_{t}}\beta_t} n$\;
    
    $ \hat{w}^t \leftarrow H_\theta (\hat{z}^t_0,t)$\;
    
    $z_{t-1} \leftarrow z'_{t-1} - \eta \nabla_{z_t} \| w - \hat{w}^t \|_2$\;
}
Decoding: $\hat{x} = \mathcal{D}(z_0)$\;
\Return $\hat{x}$\;
\end{algorithm}

\section{Experiments}\label{sec:experiments}

In this section, we present experiments comparing our method, SILO, with other latent diffusion-based methods that are reproducible using public implementations.
The methods in our comparison include LDPS \cite{SolvingInverseProblemssong2023}, GML-DPS \cite{SolvingLinearInverserout2023},
PSLD \cite{SolvingLinearInverserout2023} and ReSample \cite{SolvingInverseProblemssong2023}.
We begin by describing the metrics used in our evaluation, provide details on the tested degradations, and then present the experimental results.
Implementation details are described in the Appendix.

\subsection{Metrics}\label{sec:metrics}
Relying solely on distortion or perceptual metrics can be misleading.
Low distortion does not imply a realistic image, and high perceptual quality does not imply low discrepancy with the ground-truth image \cite{PerceptionDistortionTradeoffblau2018}.
Thus, we include a comprehensive set of metrics in our experiments to provide a thorough comparison of SILO with other methods. These include the following: 

\noindent {\bf Distortion.}
We report PSNR %
and LPIPS\footnote{As recommended by \cite{PerceptualSimilarityzhang2018}, we use AlexNet \cite{ImageNetClassificationDeepkrizhevsky2012} for evaluation, as it is preferred over VGG \cite{VeryDeepConvolutionalsimonyan2015}.
Unlike prior work, we present $\text{LPIPS}_\text{Alex}$ in \cref{sec:results} and include both versions in Appendix for completeness.} \cite{UnreasonableEffectivenessDeepzhang2018}
values between the restorations and groudn-truth images for each degradation separately, averaged over the test set.

\noindent {\bf Perception.}
The perception metrics we provide are Fréchet inception distance (\textbf{FID}) \cite{GANsTrainedTwoheusel2017}, and kernel inception distance (\textbf{KID}) \cite{DemystifyingMMDGANsbinkowski2018} multiplied by a factor of $10^3$.
These reported values assess the distance between $p_x$ (the real image distribution from the test set) and $p_{\hat{x}}$.

\noindent {\bf Runtime.} We measure the algorithms' duration from start to end of the restoration process [seconds].
Computations are performed on an NVIDIA L40S GPU with full precision (FP32) and averaged over 100 images.
Runtime measurements refer to the Super-resolution $\times8$ task.

\subsection{Degradations, datasets and models }\label{sec:Degradations}

We evaluate SILO and the competing methods across a variety of common degradations:

\noindent \textbf{Gaussian blur.} The images are padded with reflection and convolved with a Gaussian kernel of size $61 \times 61$ with a standard deviation of 3.

\noindent \textbf{Super-resolution \boldmath$\times4$ or \boldmath$\times8$.} The image are downscaled by a factor of $4$ or $8$ with a bicubic kernel.

\noindent \textbf{Inpainting.} A box mask of size $256$ by $256$ pixels is applied to the center of the image.

\noindent \textbf{JPEG.} JPEG decompression with quality factor $10$ is applied to the image.

Linear degradations are applied for all methods, with an additional non-linear degradation for methods that support it.
Images are normalized to the range $[-1,1]$, processed through $\mathcal{A}$, and, unless otherwise specified, white Gaussian noise with $\sigma_y =0.01$ is added in accordance with \cref{eq:general_inverse_problem}.
This noise level is chosen for consistency with previous works using latent diffusion \cite{SolvingInverseProblemssong2023,PrompttuningLatentDiffusionchung2023,FirstOrderTweedieSolvingrout2024}.

For face restoration tasks, experiments are performed over the FFHQ dataset \cite{StyleBasedGeneratorArchitecturekarras2019}, scaled to $512\times512$.
The test set consists of the first $1,000$ images in the dataset, and the training set consists of the remaining images.
For general restorations, the training set is LSDIR-train \cite{LSDIRLargeScaleli2023}, and the test set is the first $1,000$ images of COCO-val2017 \cite{MicrosoftCOCOCommonlin2014}.

We use SD-v1.5~\cite{HighResolutionImageSynthesisrombach2022} and RV-v5.1~\cite{StablediffusionapiRealisticvision51Huggingsg161222} as our pretrained diffusion models. 
These models share the same architecture, but RV-v5.1 produces more realistic generations.
For the Autoencoder \cite{AutoEncodingVariationalBayeskingma2022}, we use the default model for SD-v1.5, which is also compatible with RV-v5.1.

For simplicity, we choose $H_\theta$ as Readout-Guidence (RG), the network suggested in~\cite{ReadoutGuidanceLearningluo2024}, with a minor modification to condition on the noise level, $\sigma_y$, in the measurement.
As described in the paper, this operator extracts features from the denoising network and learns how to combine and process them to produce the desired output. 
Notably, $H_\theta$ is also $t$-dependent by design, unlike $\mathcal{A}$,  which does not depend on the diffusion timestep.
We present a preliminary ablation on this matter in \cref{sec:results} and leave further investigation for future research.

\subsection{Results}\label{sec:results}

\begingroup
\newcolumntype{M}[1]{>{\centering\arraybackslash}m{#1}}
\newcommand{\vcentered}[1]{\begin{tabular}{@{}l@{}} #1 \end{tabular}}
\setlength{\tabcolsep}{0pt} %
\renewcommand{\arraystretch}{0} %

\def\columns{6}
\def\totalwidth{0.8}

\FPeval{\colwidth}{\totalwidth/\columns}
\FPeval{\imgwidth}{\totalwidth/\columns *\columns}

\FPeval{\colwidth}{clip(\totalwidth/\columns)}

\begin{figure*}[h!]
    \centering
    \begin{tabular}{*{\columns}{M{\colwidth\linewidth}}}

        \footnotesize{$x$} &
        \footnotesize{$y$} &
        \footnotesize{Ours (RV)} &
        \footnotesize{Ours (SD)} &
        \footnotesize{ReSample} &
        \footnotesize{PSLD} 
        \\
	    
        \rule{0pt}{0.8ex}\\

        \multicolumn{6}{c}{\centered{\includegraphics[width=\imgwidth\linewidth]{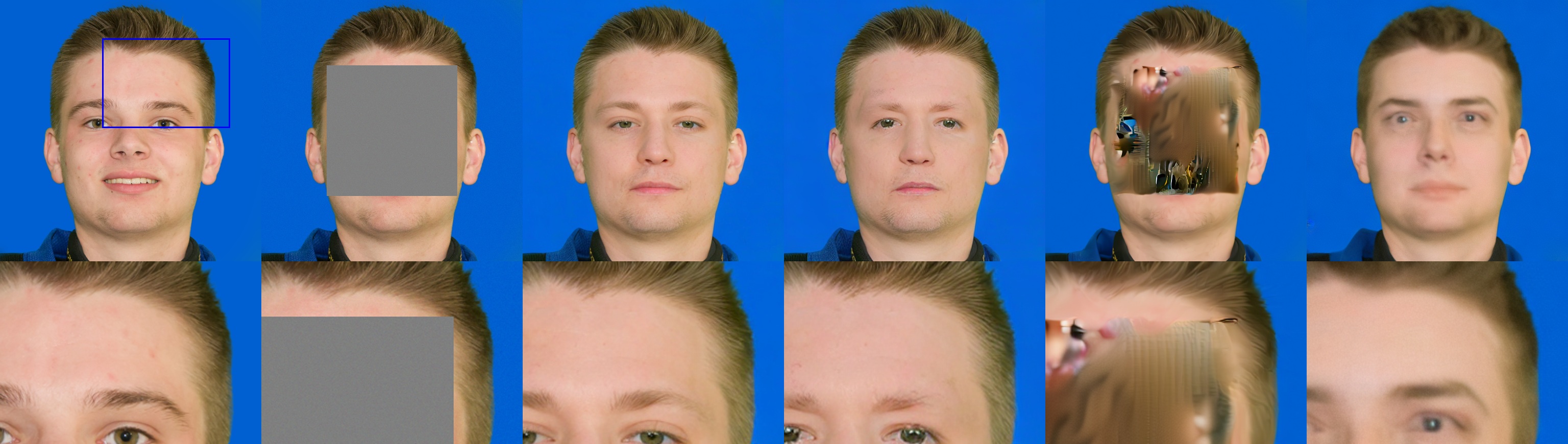}}} 
        \\
        \multicolumn{6}{c}{\centered{\includegraphics[width=\imgwidth\linewidth]{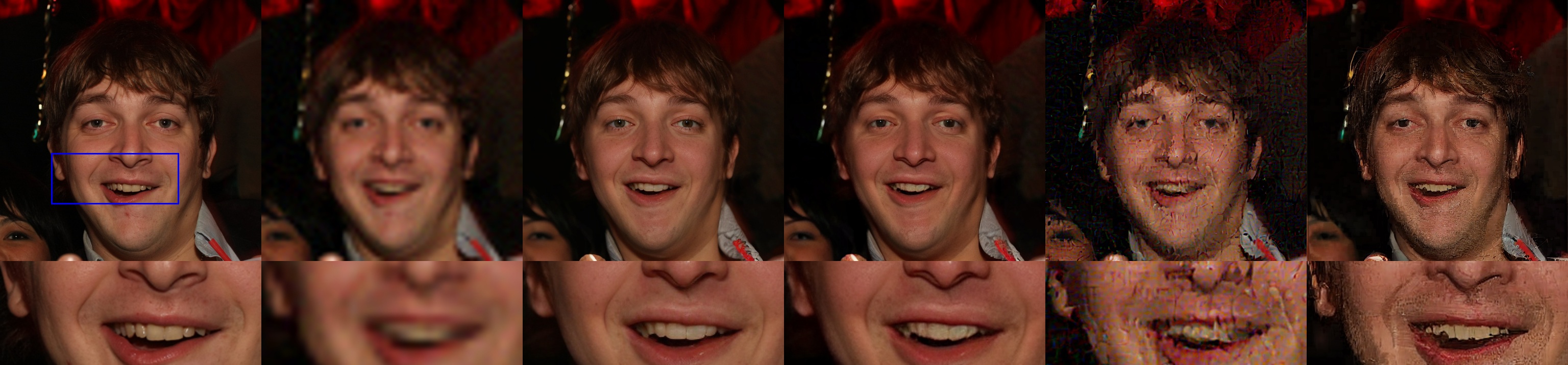}}} 
        \\
        \multicolumn{6}{c}{\centered{\includegraphics[width=\imgwidth \linewidth]{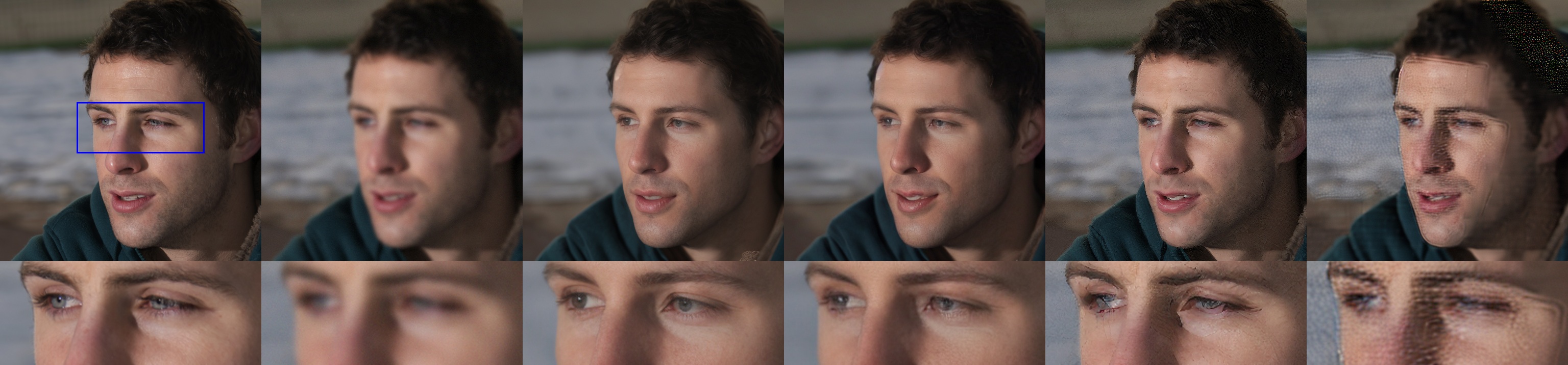}}} 
        \\

    \end{tabular}
    \caption{\textbf{Comparison of SILO to other methods.}
    From left to right: the clean image, $x$, the measurement $y$, the reconstructions using SILO (with RV and SD), ReSample, and PSLD.
    Each row contains the image and a zoom-in to show the differences better.
    From top to bottom the degredations are inpainting, super-resolution (8), and Gaussian blur. The settings are detailed in \cref{sec:Degradations}.}
    \label{fig:ffhq comparison}
\end{figure*}
\endgroup

\begingroup
\def\totalwidth{1}
\def\numdegredations{3}
\def\nummetrics{4}
\FPeval{\totalcolumns}{round(\numdegredations*\nummetrics +1 ,0)}
\FPeval{\colwidth}{\totalwidth/\totalcolumns}

\newcommand{\rankhl}[4]{%
  \ifnum\numexpr\pdfstrcmp{#1}{#2}=0
    \cellcolor{first}#1%
  \else\ifnum\numexpr\pdfstrcmp{#1}{#3}=0
    \cellcolor{second}#1%
  \else\ifnum\numexpr\pdfstrcmp{#1}{#4}=0
    \cellcolor{third}#1%
  \else
    #1%
  \fi\fi\fi
}

\newcommand{\rankhlrev}[4]{%
  \ifnum\numexpr\pdfstrcmp{#1}{#4}=0
    \cellcolor{first}#1%
  \else\ifnum\numexpr\pdfstrcmp{#1}{#3}=0
    \cellcolor{second}#1%
  \else\ifnum\numexpr\pdfstrcmp{#1}{#2}=0
    \cellcolor{third}#1%
  \else
    #1%
  \fi\fi\fi
}

\edef\firstTime{148}\edef\secondTime{149}\edef\thirdTime{331}
\edef\firstPSNRa{27.01}\edef\secondPSNRa{26.89}\edef\thirdPSNRa{26.28}
\edef\firstCPSNRa{41.23}\edef\secondCPSNRa{40.91}\edef\thirdCPSNRa{38.73}
\edef\firstLPIPSa{0.320}\edef\secondLPIPSa{0.253}\edef\thirdLPIPSa{0.226}
\edef\firstFIDa{38.50}\edef\secondFIDa{30.71}\edef\thirdFIDa{27.10}
\edef\firstKIDa{12.94}\edef\secondKIDa{8.47}\edef\thirdKIDa{5.23}

\edef\firstPSNRb{22.51}\edef\secondPSNRb{22.23}\edef\thirdPSNRb{20.64}
\edef\firstCPSNRb{36.11}\edef\secondCPSNRb{30.09}\edef\thirdCPSNRb{30.04}
\edef\firstLPIPSb{0.273}\edef\secondLPIPSb{0.151}\edef\thirdLPIPSb{0.139}
\edef\firstFIDb{49.56}\edef\secondFIDb{21.04}\edef\thirdFIDb{18.98}
\edef\firstKIDb{16.02}\edef\secondKIDb{4.32}\edef\thirdKIDb{1.80}

\edef\firstPSNRc{27.60}\edef\secondPSNRc{25.69}\edef\thirdPSNRc{25.52}
\edef\firstLPIPSc{0.268}\edef\secondLPIPSc{0.212}\edef\thirdLPIPSc{0.203}
\edef\firstFIDc{33.69}\edef\secondFIDc{27.30}\edef\thirdFIDc{25.48}
\edef\firstKIDc{7.54}\edef\secondKIDc{6.02}\edef\thirdKIDc{4.21}

\begin{table*}[tb]
    \centering
    \setlength{\tabcolsep}{4pt}
    \small
    \begin{tabular}{l *{\totalcolumns}{c}}
        \multicolumn{2}{c}{} &
	\multicolumn{\nummetrics}{c}{Super-Resolution $\times8$} &
	\multicolumn{\nummetrics}{c}{Inpainting} &
        \multicolumn{\nummetrics}{c}{JPEG}
	    \\
     
        \cmidrule(lr){3-6}
        \cmidrule(lr){7-10}
        \cmidrule(lr){11-14}
        
        Method &
        Time [sec]&
        PSNR &
        LPIPS &
        FID &
        KID &

        PSNR &
        LPIPS &
        FID &
        KID &
        
        PSNR &
        LPIPS &
        FID&
        KID 
        \\

        \hline
        
        Ours (RV)& 
        \rankhl{149}{\firstTime}{\secondTime}{\thirdTime}&
        \rankhl{26.28}{\firstPSNRa}{\secondPSNRa}{\thirdPSNRa}&
        \rankhlrev{0.226}{\firstLPIPSa}{\secondLPIPSa}{\thirdLPIPSa}&
        \rankhlrev{27.10}{\firstFIDa}{\secondFIDa}{\thirdFIDa}&
        \rankhlrev{5.23}{\firstKIDa}{\secondKIDa}{\thirdKIDa}&
        
        \rankhl{22.51}{\firstPSNRb}{\secondPSNRb}{\thirdPSNRb}&
        \rankhlrev{0.139}{\firstLPIPSb}{\secondLPIPSb}{\thirdLPIPSb}&
        \rankhlrev{18.98}{\firstFIDb}{\secondFIDb}{\thirdFIDb}&
        \rankhlrev{1.80}{\firstKIDb}{\secondKIDb}{\thirdKIDb} &

        \rankhl{25.52}{\firstPSNRc}{\secondPSNRc}{\thirdPSNRc}&
        \rankhlrev{0.203}{\firstLPIPSc}{\secondLPIPSc}{\thirdLPIPSc}&
        \rankhlrev{25.48}{\firstFIDc}{\secondFIDc}{\thirdFIDc}&
        \rankhlrev{4.21}{\firstKIDc}{\secondKIDc}{\thirdKIDc}
        \\

        Ours (SD)& 
        \rankhl{148}{\firstTime}{\secondTime}{\thirdTime}&
        \rankhl{26.13}{\firstPSNRa}{\secondPSNRa}{\thirdPSNRa}&
        \rankhlrev{0.253}{\firstLPIPSa}{\secondLPIPSa}{\thirdLPIPSa}&
        \rankhlrev{30.71}{\firstFIDa}{\secondFIDa}{\thirdFIDa}&
        \rankhlrev{8.47}{\firstKIDa}{\secondKIDa}{\thirdKIDa}&
        
        \rankhl{22.23}{\firstPSNRb}{\secondPSNRb}{\thirdPSNRb}&
        \rankhlrev{0.151}{\firstLPIPSb}{\secondLPIPSb}{\thirdLPIPSb}&
        \rankhlrev{21.04}{\firstFIDb}{\secondFIDb}{\thirdFIDb}&
        \rankhlrev{4.32}{\firstKIDb}{\secondKIDb}{\thirdKIDb}&

        \rankhl{25.40}{\firstPSNRc}{\secondPSNRc}{\thirdPSNRc}&
        \rankhlrev{0.212}{\firstLPIPSc}{\secondLPIPSc}{\thirdLPIPSc}&
        \rankhlrev{27.30}{\firstFIDc}{\secondFIDc}{\thirdFIDc}&
        \rankhlrev{6.02}{\firstKIDc}{\secondKIDc}{\thirdKIDc}
        \\

        ReSample& 
        \rankhl{1418}{\firstTime}{\secondTime}{\thirdTime}&
        \rankhl{22.80}{\firstPSNRa}{\secondPSNRa}{\thirdPSNRa}&
        \rankhlrev{0.575}{\firstLPIPSa}{\secondLPIPSa}{\thirdLPIPSa}&
        \rankhlrev{131.75}{\firstFIDa}{\secondFIDa}{\thirdFIDa}&
        \rankhlrev{118.57}{\firstKIDa}{\secondKIDa}{\thirdKIDa}&
        
        \rankhl{16.91}{\firstPSNRb}{\secondPSNRb}{\thirdPSNRb}&
        \rankhlrev{0.273}{\firstLPIPSb}{\secondLPIPSb}{\thirdLPIPSb}&
        \rankhlrev{146.08}{\firstFIDb}{\secondFIDb}{\thirdFIDb}&
        \rankhlrev{119.34}{\firstKIDb}{\secondKIDb}{\thirdKIDb}&

        \rankhl{25.69}{\firstPSNRc}{\secondPSNRc}{\thirdPSNRc}&
        \rankhlrev{0.456}{\firstLPIPSc}{\secondLPIPSc}{\thirdLPIPSc}&
        \rankhlrev{39.71}{\firstFIDc}{\secondFIDc}{\thirdFIDc}&
        \rankhlrev{20.17}{\firstKIDc}{\secondKIDc}{\thirdKIDc}
        \\
        PSLD& 
        \rankhl{390}{\firstTime}{\secondTime}{\thirdTime}&
        \rankhl{25.08}{\firstPSNRa}{\secondPSNRa}{\thirdPSNRa}&
        \rankhlrev{0.320}{\firstLPIPSa}{\secondLPIPSa}{\thirdLPIPSa}&
        \rankhlrev{41.58}{\firstFIDa}{\secondFIDa}{\thirdFIDa}&
        \rankhlrev{14.90}{\firstKIDa}{\secondKIDa}{\thirdKIDa}&
        
        \rankhl{20.58}{\firstPSNRb}{\secondPSNRb}{\thirdPSNRb}&
        \rankhlrev{0.357}{\firstLPIPSb}{\secondLPIPSb}{\thirdLPIPSb}&
        \rankhlrev{50.84}{\firstFIDb}{\secondFIDb}{\thirdFIDb}&
        \rankhlrev{17.23}{\firstKIDb}{\secondKIDb}{\thirdKIDb} &
        \multicolumn{4}{c}{\cellcolor{invalid}cannot compute for nonlinear $\mathcal{A}$}

        \\

        GML-DPS& 
        \rankhl{389}{\firstTime}{\secondTime}{\thirdTime}&
        \rankhl{27.01}{\firstPSNRa}{\secondPSNRa}{\thirdPSNRa}&
        \rankhlrev{0.327}{\firstLPIPSa}{\secondLPIPSa}{\thirdLPIPSa}&
        \rankhlrev{38.71}{\firstFIDa}{\secondFIDa}{\thirdFIDa}&
        \rankhlrev{12.99}{\firstKIDa}{\secondKIDa}{\thirdKIDa}&
        
        \rankhl{20.64}{\firstPSNRb}{\secondPSNRb}{\thirdPSNRb}&
        \rankhlrev{0.356}{\firstLPIPSb}{\secondLPIPSb}{\thirdLPIPSb}&
        \rankhlrev{49.89}{\firstFIDb}{\secondFIDb}{\thirdFIDb}&
        \rankhlrev{16.54}{\firstKIDb}{\secondKIDb}{\thirdKIDb}&

        \rankhl{27.60}{\firstPSNRc}{\secondPSNRc}{\thirdPSNRc}&
        \rankhlrev{0.268}{\firstLPIPSc}{\secondLPIPSc}{\thirdLPIPSc}&
        \rankhlrev{33.69}{\firstFIDc}{\secondFIDc}{\thirdFIDc}&
        \rankhlrev{7.54}{\firstKIDc}{\secondKIDc}{\thirdKIDc}
        \\
        
        LDPS& 
        \rankhl{331}{\firstTime}{\secondTime}{\thirdTime}&
        \rankhl{26.89}{\firstPSNRa}{\secondPSNRa}{\thirdPSNRa}&
        \rankhlrev{0.343}{\firstLPIPSa}{\secondLPIPSa}{\thirdLPIPSa}&
        \rankhlrev{38.50}{\firstFIDa}{\secondFIDa}{\thirdFIDa}&
        \rankhlrev{12.94}{\firstKIDa}{\secondKIDa}{\thirdKIDa}&
        
        \rankhl{20.58}{\firstPSNRb}{\secondPSNRb}{\thirdPSNRb}&
        \rankhlrev{0.368}{\firstLPIPSb}{\secondLPIPSb}{\thirdLPIPSb}&
        \rankhlrev{49.56}{\firstFIDb}{\secondFIDb}{\thirdFIDb}&
        \rankhlrev{16.02}{\firstKIDb}{\secondKIDb}{\thirdKIDb}&

        \rankhl{24.53}{\firstPSNRc}{\secondPSNRc}{\thirdPSNRc}&
        \rankhlrev{0.373}{\firstLPIPSc}{\secondLPIPSc}{\thirdLPIPSc}&
        \rankhlrev{53.21}{\firstFIDc}{\secondFIDc}{\thirdFIDc}&
        \rankhlrev{17.71}{\firstKIDc}{\secondKIDc}{\thirdKIDc}
        \\

    \end{tabular}
    \caption{Comparison of inverse problem solvers using latent diffusion on the FFHQ dataset.
    }
    \label{tab:results_of_sr8_and_ip}
\end{table*}
\endgroup

Quantitative results in \cref{tab:results_of_sr4_and_gb,tab:results_of_sr8_and_ip} show that our method consistently outperforms all other methods by FID, KID, and LPIPS while achieving a shorter reconstruction time by a factor of ${\sim3}$ compared to PSLD and ${\sim10}$ compared to ReSample. 
Qualitative comparisons in \cref{fig:ffhq comparison} reveal the differences in reconstruction results.
PSLD and ReSample %
lack high perceptual quality, as manifested by noticeable artifacts in their reconstructions. 
In contrast, SILO %
generates natural-looking images while maintaining consistent reconstructions, all while using the same diffusion prior as other methods.

We further demonstrate our method's advantages through additional experiments, showing (1) enhanced robustness to measurement noise, (2) improved performance when leveraging high-quality text conditions, (3) generalizability across diverse datasets, and (4) usage of a learned operator that is not conditioned on $t$.

\noindent {\bf Noisier measurements.}
Works using LDMs add a small amount of noise to the measurements.
This choice might relate to the observation that these methods struggle to maintain high perceptual quality at higher noise levels.
As shown in \cref{tab:ablation_noise}, and the second row in \cref{fig:first}, our method achieves reasonable results even at a higher noise level ($\sigma_y = 0.03$).

\begingroup
\def\totalwidth{1}
\def\numdegredations{1}
\def\nummetrics{4}
\FPeval{\totalcolumns}{round(\numdegredations*\nummetrics,0)}
\FPeval{\colwidth}{\totalwidth/\totalcolumns}

\newcommand{\rankhl}[4]{%
  \ifnum\numexpr\pdfstrcmp{#1}{#2}=0
    \cellcolor{first}#1%
  \else\ifnum\numexpr\pdfstrcmp{#1}{#3}=0
    \cellcolor{second}#1%
  \else\ifnum\numexpr\pdfstrcmp{#1}{#4}=0
    \cellcolor{third}#1%
  \else
    #1%
  \fi\fi\fi
}

\newcommand{\rankhlrev}[4]{%
  \ifnum\numexpr\pdfstrcmp{#1}{#4}=0
    \cellcolor{first}#1%
  \else\ifnum\numexpr\pdfstrcmp{#1}{#3}=0
    \cellcolor{second}#1%
  \else\ifnum\numexpr\pdfstrcmp{#1}{#2}=0
    \cellcolor{third}#1%
  \else
    #1%
  \fi\fi\fi
}

\edef\firstTime{0}\edef\secondTime{0}\edef\thirdTime{0}
\edef\firstPSNRa{26.20}\edef\secondPSNRa{26.19}\edef\thirdPSNRa{25.14}
\edef\firstCPSNRa{35.52}\edef\secondCPSNRa{34.87}\edef\thirdCPSNRa{34.30}
\edef\firstLPIPSa{0.359}\edef\secondLPIPSa{0.279}\edef\thirdLPIPSa{0.252}
\edef\firstFIDa{39.61}\edef\secondFIDa{32.77}\edef\thirdFIDa{30.28}
\edef\firstKIDa{12.80}\edef\secondKIDa{8.00}\edef\thirdKIDa{6.04}

\begin{table}[!ht]
    \centering
    \setlength{\tabcolsep}{4pt}
    \small
    \begin{tabular}{l *{\totalcolumns}{c}}
        
        Method &
        PSNR &
        LPIPS &
        FID &
        KID \cr
        \hline
        
        Ours (RV)& 
        \rankhl{25.27}{\firstPSNRa}{\secondPSNRa}{\thirdPSNRa}&
        \rankhlrev{0.252}{\firstLPIPSa}{\secondLPIPSa}{\thirdLPIPSa}&
        \rankhlrev{30.28}{\firstFIDa}{\secondFIDa}{\thirdFIDa}&
        \rankhlrev{6.04}{\firstKIDa}{\secondKIDa}{\thirdKIDa}
        \\

        Ours (SD)& 
        \rankhl{25.21}{\firstPSNRa}{\secondPSNRa}{\thirdPSNRa}&
        \rankhlrev{0.279}{\firstLPIPSa}{\secondLPIPSa}{\thirdLPIPSa}&
        \rankhlrev{32.77}{\firstFIDa}{\secondFIDa}{\thirdFIDa}&
        \rankhlrev{8.00}{\firstKIDa}{\secondKIDa}{\thirdKIDa}
        \\

        ReSample& 
        \rankhl{16.18}{\firstPSNRa}{\secondPSNRa}{\thirdPSNRa}&
        \rankhlrev{0.724}{\firstLPIPSa}{\secondLPIPSa}{\thirdLPIPSa}&
        \rankhlrev{235.8}{\firstFIDa}{\secondFIDa}{\thirdFIDa}&
        \rankhlrev{253.1}{\firstKIDa}{\secondKIDa}{\thirdKIDa}
        
        \\
        PSLD& 
        \rankhl{24.37}{\firstPSNRa}{\secondPSNRa}{\thirdPSNRa}&
        \rankhlrev{0.359}{\firstLPIPSa}{\secondLPIPSa}{\thirdLPIPSa}&
        \rankhlrev{61.99}{\firstFIDa}{\secondFIDa}{\thirdFIDa}&
        \rankhlrev{33.23}{\firstKIDa}{\secondKIDa}{\thirdKIDa}
        \\

        GML-DPS& 
        \rankhl{26.19}{\firstPSNRa}{\secondPSNRa}{\thirdPSNRa}&
        \rankhlrev{0.354}{\firstLPIPSa}{\secondLPIPSa}{\thirdLPIPSa}&
        \rankhlrev{41.06}{\firstFIDa}{\secondFIDa}{\thirdFIDa}&
        \rankhlrev{13.69}{\firstKIDa}{\secondKIDa}{\thirdKIDa}
        \\
        
        LDPS& 
        \rankhl{26.20}{\firstPSNRa}{\secondPSNRa}{\thirdPSNRa}&
        \rankhlrev{0.359}{\firstLPIPSa}{\secondLPIPSa}{\thirdLPIPSa}&
        \rankhlrev{39.61}{\firstFIDa}{\secondFIDa}{\thirdFIDa}&
        \rankhlrev{12.80}{\firstKIDa}{\secondKIDa}{\thirdKIDa}
        \\
        
    \end{tabular}
    \caption{Comparison of inverse problem solvers using LDMs on the FFHQ dataset, for SR $\times8$ with $\sigma_y=0.03$. }
    \label{tab:ablation_noise}
\end{table}
\endgroup

\noindent {\bf Priors and text conditioning.}
Since the diffusion priors are text-conditioned, we explore the relationship between an informative prompt, which includes details about the clean image, and the method's ability to reconstruct it as a sharp, clean image. 
In \cref{tab:sampling_ablation}, we compare the quality of reconstructions generated using RV-v5.1 and SD-v1.5.
For each model, we use \cref{alg:reconstruction} with varying prompts and classifier-free guidance (CFG) \cite{ClassifierFreeDiffusionGuidanceho2021}.
We test three types of prompts: a null prompt (an empty string), a generic prompt (``A high quality photo of a face''), and a set of high-quality (HQ) prompts, specific to each image.
The HQ prompts are generated using Qwen \cite{Qwen2VLEnhancingVisionLanguagewang2024}, a vision-language model.
Looking at \cref{tab:sampling_ablation}, we observe that better diffusion models, using detailed text conditions and sampling with CFG, can improve the reconstructions' perceptual quality.

\begingroup
\def\totalwidth{1}
\def\nummetrics{4}
\FPeval{\totalcolumns}{round(\nummetrics + 3,0)}
\FPeval{\colwidth}{\totalwidth/\totalcolumns}

\newcommand{\rankhl}[4]{%
  \ifnum\numexpr\pdfstrcmp{#1}{#2}=0
    \cellcolor{first}#1%
  \else\ifnum\numexpr\pdfstrcmp{#1}{#3}=0
    \cellcolor{second}#1%
  \else\ifnum\numexpr\pdfstrcmp{#1}{#4}=0
    \cellcolor{third}#1%
  \else
    #1%
  \fi\fi\fi
}

\newcommand{\rankhlrev}[4]{%
  \ifnum\numexpr\pdfstrcmp{#1}{#4}=0
    \cellcolor{first}#1%
  \else\ifnum\numexpr\pdfstrcmp{#1}{#3}=0
    \cellcolor{second}#1%
  \else\ifnum\numexpr\pdfstrcmp{#1}{#2}=0
    \cellcolor{third}#1%
  \else
    #1%
  \fi\fi\fi
}

\begin{table}[]
    \centering
    \setlength{\tabcolsep}{4pt}
    \small
    \begin{tabular}{*{\totalcolumns}{c}}
        Model & CFG & Prompt
        & PSNR & LPIPS  & FID & KID \\
        \hline
         \multirow{4}{*}{RV} & \multirow{2}{*}{4} & HQ      &  25.78  & 0.219  &  24.43 & 2.27 \\
         &  & generic              & 26.06  & 0.222    & 26.21 & 3.96 \\
         \cmidrule(lr){2-7}
         & \multirow{2}{*}{1} & generic               & 26.28  & 0.226    & 27.10 & 5.23 \\
         &  & null             & 26.35  & 0.230    & 27.63 & 5.83 \\
         \hline
        \multirow{4}{*}{SD}  & \multirow{2}{*}{4} & HQ       & 26.08  & 0.252    & 29.29 & 7.65 \\
        &  & generic            & 26.05  & 0.246  &  29.24 & 7.03 \\
        \cmidrule(lr){2-7}
         & \multirow{2}{*}{1} & generic                 & 26.13  & 0.253   & 30.71 & 8.47 \\
         &  & null                       & 26.18  & 0.263   & 32.55 & 10.3 \\
        \hline
    \end{tabular}
    \caption{Comparison of using SILO on FFHQ dataset, for SR $\times 8$, when different models, CFG levels, and text-conditions are used.}
    \label{tab:sampling_ablation}
\end{table}
\endgroup
\label{sec:hq_prompt_prior_ablation}

\noindent {\bf Results on COCO.} 
To show that SILO is not limited to a specific domain of images such as face images, we compare solutions of inpainting on the COCO dataset.
In \cref{tab:coco,fig:coco_fig}, we demonstrate that SILO outperforms competing methods in this challenging dataset as well, providing sharp and plausible reconstructions.
\begingroup
\newcolumntype{M}[1]{>{\centering\arraybackslash}m{#1}}
\newcommand{\vcentered}[1]{\begin{tabular}{@{}l@{}} #1 \end{tabular}}
\setlength{\tabcolsep}{0pt} %
\renewcommand{\arraystretch}{0} %

\def\columns{4}
\def\totalwidth{1}

\FPeval{\colwidth}{\totalwidth/\columns}
\FPeval{\imgwidth}{\totalwidth/\columns *\columns}

\FPeval{\colwidth}{clip(\totalwidth/\columns)}

\begin{figure}[tb]
    \centering
    \begin{tabular}{*{\columns}{M{\colwidth\linewidth}}}

        \footnotesize{$y$} &
        \footnotesize{Ours (SD)} &
        \footnotesize{ReSample} &
        \footnotesize{PSLD} 
        \\
	    
        \rule{0pt}{0.8ex}\\

        \multicolumn{4}{c}{\centered{\includegraphics[width=\imgwidth\linewidth]{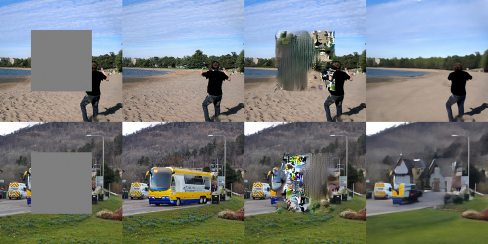}}} 

    \end{tabular}
    \caption{Restorations of masked images from the COCO dataset.}
    \label{fig:coco_fig}
\end{figure}
\endgroup

\begingroup
\def\totalwidth{1}
\def\numdegredations{1}
\def\nummetrics{4}
\FPeval{\totalcolumns}{round(\numdegredations*\nummetrics,0)}
\FPeval{\colwidth}{\totalwidth/\totalcolumns}

\newcommand{\rankhl}[4]{%
  \ifnum\numexpr\pdfstrcmp{#1}{#2}=0
    \cellcolor{first}#1%
  \else\ifnum\numexpr\pdfstrcmp{#1}{#3}=0
    \cellcolor{second}#1%
  \else\ifnum\numexpr\pdfstrcmp{#1}{#4}=0
    \cellcolor{third}#1%
  \else
    #1%
  \fi\fi\fi
}

\newcommand{\rankhlrev}[4]{%
  \ifnum\numexpr\pdfstrcmp{#1}{#4}=0
    \cellcolor{first}#1%
  \else\ifnum\numexpr\pdfstrcmp{#1}{#3}=0
    \cellcolor{second}#1%
  \else\ifnum\numexpr\pdfstrcmp{#1}{#2}=0
    \cellcolor{third}#1%
  \else
    #1%
  \fi\fi\fi
}

\edef\firstTime{0}\edef\secondTime{0}\edef\thirdTime{0}
\edef\firstPSNRa{18.51}\edef\secondPSNRa{18.30}\edef\thirdPSNRa{18.25}
\edef\firstCPSNRa{35.52}\edef\secondCPSNRa{34.87}\edef\thirdCPSNRa{34.30}
\edef\firstLPIPSa{0.297}\edef\secondLPIPSa{0.221}\edef\thirdLPIPSa{0.214}
\edef\firstFIDa{88.16}\edef\secondFIDa{48.96}\edef\thirdFIDa{45.59}
\edef\firstKIDa{21.99}\edef\secondKIDa{3.74}\edef\thirdKIDa{2.15}

\begin{table}[]
    \centering
    \setlength{\tabcolsep}{4pt}
    \small
    \begin{tabular}{l *{\totalcolumns}{c}}
        
        Method &
        PSNR &
        LPIPS &
        FID &
        KID \cr
        \hline
        
        Ours (RV)& 
        \rankhl{18.51}{\firstPSNRa}{\secondPSNRa}{\thirdPSNRa}&
        \rankhlrev{0.214}{\firstLPIPSa}{\secondLPIPSa}{\thirdLPIPSa}&
        \rankhlrev{48.96}{\firstFIDa}{\secondFIDa}{\thirdFIDa}&
        \rankhlrev{3.74}{\firstKIDa}{\secondKIDa}{\thirdKIDa}
        \\

        Ours (SD)& 
        \rankhl{18.30}{\firstPSNRa}{\secondPSNRa}{\thirdPSNRa}&
        \rankhlrev{0.221}{\firstLPIPSa}{\secondLPIPSa}{\thirdLPIPSa}&
        \rankhlrev{45.59}{\firstFIDa}{\secondFIDa}{\thirdFIDa}&
        \rankhlrev{2.15}{\firstKIDa}{\secondKIDa}{\thirdKIDa}
        \\

        ReSample& 
        \rankhl{16.53}{\firstPSNRa}{\secondPSNRa}{\thirdPSNRa}&
        \rankhlrev{0.297}{\firstLPIPSa}{\secondLPIPSa}{\thirdLPIPSa}&
        \rankhlrev{104.37}{\firstFIDa}{\secondFIDa}{\thirdFIDa}&
        \rankhlrev{54.16}{\firstKIDa}{\secondKIDa}{\thirdKIDa}
        
        \\
        PSLD& 
        \rankhl{18.24}{\firstPSNRa}{\secondPSNRa}{\thirdPSNRa}&
        \rankhlrev{0.454}{\firstLPIPSa}{\secondLPIPSa}{\thirdLPIPSa}&
        \rankhlrev{90.38}{\firstFIDa}{\secondFIDa}{\thirdFIDa}&
        \rankhlrev{24.40}{\firstKIDa}{\secondKIDa}{\thirdKIDa}
        \\

        GML-DPS& 
        \rankhl{18.25}{\firstPSNRa}{\secondPSNRa}{\thirdPSNRa}&
        \rankhlrev{0.453}{\firstLPIPSa}{\secondLPIPSa}{\thirdLPIPSa}&
        \rankhlrev{88.16}{\firstFIDa}{\secondFIDa}{\thirdFIDa}&
        \rankhlrev{21.99}{\firstKIDa}{\secondKIDa}{\thirdKIDa}
        \\
        
        LDPS& 
        \rankhl{18.26}{\firstPSNRa}{\secondPSNRa}{\thirdPSNRa}&
        \rankhlrev{0.474}{\firstLPIPSa}{\secondLPIPSa}{\thirdLPIPSa}&
        \rankhlrev{92.67}{\firstFIDa}{\secondFIDa}{\thirdFIDa}&
        \rankhlrev{24.98}{\firstKIDa}{\secondKIDa}{\thirdKIDa}
        \\
        
    \end{tabular}
    \caption{Comparison of inverse problem solvers using latent diffusion on 1,000 images from the COCO dataset with inpainting.}
    \label{tab:coco}
\end{table}
\endgroup

\noindent {\bf Learned operator ablation.}
The proposed $H_\theta$ is $t$-dependent.
This choice is not inherently tied to SILO or the assumptions that led to the development of \cref{alg:reconstruction}.
In \cref{tab:cnn} we show that even when $H_\theta$ is a small time-independent convolutional neural network (CNN) \cite{ImageNetClassificationDeepkrizhevsky2012,WhatBestMultistagejarrett2009,LargescaleLearningSVMhuang2006}, SILO provides good reconstructions and requires even less compute power. %

\begingroup
\def\totalwidth{1}
\def\numdegredations{1}
\def\nummetrics{4}
\FPeval{\totalcolumns}{round(\numdegredations*\nummetrics +1,0)}
\FPeval{\colwidth}{\totalwidth/\totalcolumns}

\newcommand{\rankhl}[4]{%
  \ifnum\numexpr\pdfstrcmp{#1}{#2}=0
    \cellcolor{first}#1%
  \else\ifnum\numexpr\pdfstrcmp{#1}{#3}=0
    \cellcolor{second}#1%
  \else\ifnum\numexpr\pdfstrcmp{#1}{#4}=0
    \cellcolor{third}#1%
  \else
    #1%
  \fi\fi\fi
}

\newcommand{\rankhlrev}[4]{%
  \ifnum\numexpr\pdfstrcmp{#1}{#4}=0
    \cellcolor{first}#1%
  \else\ifnum\numexpr\pdfstrcmp{#1}{#3}=0
    \cellcolor{second}#1%
  \else\ifnum\numexpr\pdfstrcmp{#1}{#2}=0
    \cellcolor{third}#1%
  \else
    #1%
  \fi\fi\fi
}

\edef\firstTime{109}\edef\secondTime{149}\edef\thirdTime{0}
\edef\firstPSNRa{26.34}\edef\secondPSNRa{26.28}\edef\thirdPSNRa{28.00}
\edef\firstCPSNRa{50.66}\edef\secondCPSNRa{44.18}\edef\thirdCPSNRa{44.07}
\edef\firstLPIPSa{0.253}\edef\secondLPIPSa{0.230}\edef\thirdLPIPSa{0.226}
\edef\firstFIDa{30.33}\edef\secondFIDa{27.10}\edef\thirdFIDa{26.88}
\edef\firstKIDa{10.74}\edef\secondKIDa{5.23}\edef\thirdKIDa{4.65}

\edef\firstPSNRb{29.34}\edef\secondPSNRb{29.06}\edef\thirdPSNRb{28.23}
\edef\firstCPSNRb{42.74}\edef\secondCPSNRb{38.76}\edef\thirdCPSNRb{36.69}
\edef\firstLPIPSb{0.247}\edef\secondLPIPSb{0.200}\edef\thirdLPIPSb{0.182}
\edef\firstFIDb{29.63}\edef\secondFIDb{26.51}\edef\thirdFIDb{23.82}
\edef\firstKIDb{9.05}\edef\secondKIDb{7.34}\edef\thirdKIDb{5.10}

\begin{table}[bt]
    \centering
    \setlength{\tabcolsep}{4pt}
    \small
    \begin{tabular}{l *{\totalcolumns}{c}}

        Method &
        Time [sec]&
        PSNR &
        LPIPS &
        FID &
        KID 
        \\
        \hline        
        Ours (RG-RV)& 
        \rankhl{149}{\firstTime}{\secondTime}{\thirdTime}&
        \rankhl{26.28}{\firstPSNRa}{\secondPSNRa}{\thirdPSNRa}&
        \rankhlrev{0.226}{\firstLPIPSa}{\secondLPIPSa}{\thirdLPIPSa}&
        \rankhlrev{27.10}{\firstFIDa}{\secondFIDa}{\thirdFIDa}&
        \rankhlrev{5.23}{\firstKIDa}{\secondKIDa}{\thirdKIDa}
        
        \\

        Ours (CNN-RV)& 
        \rankhl{109}{\firstTime}{\secondTime}{\thirdTime}&
        \rankhl{26.34}{\firstPSNRa}{\secondPSNRa}{\thirdPSNRa}&
        \rankhlrev{0.230}{\firstLPIPSa}{\secondLPIPSa}{\thirdLPIPSa}&
        \rankhlrev{26.88}{\firstFIDa}{\secondFIDa}{\thirdFIDa}&
        \rankhlrev{4.65}{\firstKIDa}{\secondKIDa}{\thirdKIDa}
        
        \\

    \end{tabular}
    \caption{Results of SILO using RG and a CNN that is not conditioned on $t$, on the FFHQ dataset, for SR $\times 8$. }
    \label{tab:cnn}
\end{table}
\endgroup

\section{Conclusion}

This work introduced a novel approach to solving inverse problems using LDMs.
Rather than enforcing consistency to the measurement in the pixel domain, our method operates entirely within the latent space, improving both reconstruction quality and sampling runtime.

\noindent \textbf{Limitations.} Since we use the Autoencoder of SD to create $w$, we expect successful reconstructions when the measurement somewhat resembles a natural image.
This behavior holds for most common degradations.
For cases that diverge from this assumption significantly (e.g., phase retrieval), alternative methods of generating $w$ from $y$ may be needed.
Additionally, SILO requires a preliminary stage of training $H_\theta$ to mimic $\mathcal{A}$.
This is done once, and then $H_\theta$ can be used for limitless restorations afterward. 

\noindent \textbf{Future work.} Our approach opens up several directions for future research.
Using alternative encoders or feature extractors to compute $w$ could enable SILO to handle a broader range of degradations and further enhance reconstruction quality.
Another potential extension is to design $H_\theta$ to mimic a parametric family of degradations, conditioned on the parameters of $\mathcal{A}$, resulting in more versatile latent operators.
Since our method bridges the gap in cases where the score likelihood and data score are computed in different domains, future work could build upon this concept, potentially extending SILO's contributions even further.

\clearpage
{
    \small
    \bibliographystyle{ieeenat_fullname}

\begin{thebibliography}{64}
\providecommand{\natexlab}[1]{#1}
\providecommand{\url}[1]{\texttt{#1}}
\expandafter\ifx\csname urlstyle\endcsname\relax
  \providecommand{\doi}[1]{doi: #1}\else
  \providecommand{\doi}{doi: \begingroup \urlstyle{rm}\Url}\fi

\bibitem[Aharon et~al.(2006)Aharon, Elad, and Bruckstein]{KSVDAlgorithmDesigningaharon2006}
M. Aharon, M. Elad, and A. Bruckstein.
\newblock K-{{SVD}}: {{An}} algorithm for designing overcomplete dictionaries for sparse representation.
\newblock \emph{IEEE Transactions on Signal Processing}, 54\penalty0 (11):\penalty0 4311--4322, 2006.

\bibitem[Betker et~al.()Betker, Goh, Jing, Brooks, Wang, Li, Ouyang, Zhuang, Lee, Guo, Manassra, Dhariwal, Chu, Jiao, and Ramesh]{ImprovingImageGenerationbetker}
James Betker, Gabriel Goh, Li Jing, Tim Brooks, Jianfeng Wang, Linjie Li, Long Ouyang, Juntang Zhuang, Joyce Lee, Yufei Guo, Wesam Manassra, Prafulla Dhariwal, Casey Chu, Yunxin Jiao, and Aditya Ramesh.
\newblock Improving {{Image Generation}} with {{Better Captions}}.

\bibitem[Bi{\'n}kowski et~al.(2018)Bi{\'n}kowski, Sutherland, Arbel, and Gretton]{DemystifyingMMDGANsbinkowski2018}
Miko{\l}aj Bi{\'n}kowski, Danica~J. Sutherland, Michael Arbel, and Arthur Gretton.
\newblock Demystifying {{MMD GANs}}.
\newblock In \emph{International {{Conference}} on {{Learning Representations}}}, 2018.

\bibitem[Blau and Michaeli(2018)]{PerceptionDistortionTradeoffblau2018}
Yochai Blau and Tomer Michaeli.
\newblock The {{Perception-Distortion Tradeoff}}.
\newblock In \emph{Proceedings of the {{IEEE Conference}} on {{Computer Vision}} and {{Pattern Recognition}}}, pages 6228--6237, 2018.

\bibitem[Chan et~al.(2017)Chan, Wang, and Elgendy]{PlugPlayADMMImagechan2017}
Stanley~H. Chan, Xiran Wang, and Omar~A. Elgendy.
\newblock Plug-and-{{Play ADMM}} for {{Image Restoration}}: {{Fixed-Point Convergence}} and {{Applications}}.
\newblock \emph{IEEE Transactions on Computational Imaging}, 3\penalty0 (1):\penalty0 84--98, 2017.

\bibitem[Chen and Pock(2017)]{TrainableNonlinearReactionchen2017}
Yunjin Chen and Thomas Pock.
\newblock Trainable {{Nonlinear Reaction Diffusion}}: {{A Flexible Framework}} for {{Fast}} and {{Effective Image Restoration}}.
\newblock \emph{IEEE Trans. Pattern Anal. Mach. Intell.}, 39\penalty0 (6):\penalty0 1256--1272, 2017.

\bibitem[Chung et~al.(2022)Chung, Kim, Mccann, Klasky, and Ye]{DiffusionPosteriorSamplingchung2022}
Hyungjin Chung, Jeongsol Kim, Michael~Thompson Mccann, Marc~Louis Klasky, and Jong~Chul Ye.
\newblock Diffusion {{Posterior Sampling}} for {{General Noisy Inverse Problems}}.
\newblock In \emph{The {{Eleventh International Conference}} on {{Learning Representations}}}, 2022.

\bibitem[Chung et~al.(2023{\natexlab{a}})Chung, Lee, and Ye]{DecomposedDiffusionSamplerchung2023}
Hyungjin Chung, Suhyeon Lee, and Jong~Chul Ye.
\newblock Decomposed {{Diffusion Sampler}} for {{Accelerating Large-Scale Inverse Problems}}.
\newblock In \emph{The {{Twelfth International Conference}} on {{Learning Representations}}}, 2023{\natexlab{a}}.

\bibitem[Chung et~al.(2023{\natexlab{b}})Chung, Ye, Milanfar, and Delbracio]{PrompttuningLatentDiffusionchung2023}
Hyungjin Chung, Jong~Chul Ye, Peyman Milanfar, and Mauricio Delbracio.
\newblock Prompt-tuning latent diffusion models for inverse problems, 2023{\natexlab{b}}.

\bibitem[Dabov et~al.(2007)Dabov, Foi, Katkovnik, and Egiazarian]{ImageDenoisingSparsedabov2007}
Kostadin Dabov, Alessandro Foi, Vladimir Katkovnik, and Karen Egiazarian.
\newblock Image {{Denoising}} by {{Sparse}} 3-{{D Transform-Domain Collaborative Filtering}}.
\newblock \emph{IEEE Transactions on Image Processing}, 16\penalty0 (8):\penalty0 2080--2095, 2007.

\bibitem[Dhariwal and Nichol(2021)]{DiffusionModelsBeatdhariwal2021}
Prafulla Dhariwal and Alexander~Quinn Nichol.
\newblock Diffusion {{Models Beat GANs}} on {{Image Synthesis}}.
\newblock In \emph{Advances in {{Neural Information Processing Systems}}}, 2021.

\bibitem[Efron(2011)]{TweediesFormulaSelectionefron2011}
Bradley Efron.
\newblock Tweedie's {{Formula}} and {{Selection Bias}}.
\newblock \emph{Journal of the American Statistical Association}, 106\penalty0 (496):\penalty0 1602--1614, 2011.

\bibitem[Elad and Aharon(2006)]{ImageDenoisingSparseelad2006}
Michael Elad and Michal Aharon.
\newblock Image {{Denoising Via Sparse}} and {{Redundant Representations Over Learned Dictionaries}}.
\newblock \emph{IEEE Transactions on Image Processing}, 15\penalty0 (12):\penalty0 3736--3745, 2006.

\bibitem[Elad et~al.(2023)Elad, Kawar, and Vaksman]{ImageDenoisingDeepelad2023}
Michael Elad, Bahjat Kawar, and Gregory Vaksman.
\newblock Image {{Denoising}}: {{The Deep Learning Revolution}} and {{Beyond}}---{{A Survey Paper}}.
\newblock \emph{SIAM Journal on Imaging Sciences}, 16\penalty0 (3):\penalty0 1594--1654, 2023.

\bibitem[Elata et~al.(2025)Elata, Michaeli, and Elad]{AdaptiveCompressedSensingelata2025}
Noam Elata, Tomer Michaeli, and Michael Elad.
\newblock Adaptive {{Compressed Sensing}} with~{{Diffusion-Based Posterior Sampling}}.
\newblock In \emph{Computer {{Vision}} -- {{ECCV}} 2024}, pages 290--308, Cham, 2025. Springer Nature Switzerland.

\bibitem[Giryes and Elad(2012)]{SparsityBasedPoissongiryes2012}
Raja Giryes and Michael Elad.
\newblock Sparsity based {{Poisson}} denoising.
\newblock In \emph{2012 {{IEEE}} 27th {{Convention}} of {{Electrical}} and {{Electronics Engineers}} in {{Israel}}}, pages 1--5, 2012.

\bibitem[Gu et~al.(2014)Gu, Zhang, Zuo, and Feng]{WeightedNuclearNormgu2014}
Shuhang Gu, Lei Zhang, Wangmeng Zuo, and Xiangchu Feng.
\newblock Weighted {{Nuclear Norm Minimization}} with {{Application}} to {{Image Denoising}}.
\newblock In \emph{2014 {{IEEE Conference}} on {{Computer Vision}} and {{Pattern Recognition}}}, pages 2862--2869, 2014.

\bibitem[{Gurrola-Ramos} et~al.(2021){Gurrola-Ramos}, Dalmau, and Alarc{\'o}n]{ResidualDenseUNetgurrola-ramos2021}
Javier {Gurrola-Ramos}, Oscar Dalmau, and Teresa~E. Alarc{\'o}n.
\newblock A {{Residual Dense U-Net Neural Network}} for {{Image Denoising}}.
\newblock \emph{IEEE Access}, 9:\penalty0 31742--31754, 2021.

\bibitem[Heusel et~al.(2017)Heusel, Ramsauer, Unterthiner, Nessler, and Hochreiter]{GANsTrainedTwoheusel2017}
Martin Heusel, Hubert Ramsauer, Thomas Unterthiner, Bernhard Nessler, and Sepp Hochreiter.
\newblock {{GANs Trained}} by a {{Two Time-Scale Update Rule Converge}} to a {{Local Nash Equilibrium}}.
\newblock In \emph{Advances in {{Neural Information Processing Systems}}}. Curran Associates, Inc., 2017.

\bibitem[Ho and Salimans(2021)]{ClassifierFreeDiffusionGuidanceho2021}
Jonathan Ho and Tim Salimans.
\newblock Classifier-{{Free Diffusion Guidance}}.
\newblock In \emph{{{NeurIPS}} 2021 {{Workshop}} on {{Deep Generative Models}} and {{Downstream Applications}}}, 2021.

\bibitem[Ho et~al.(2020)Ho, Jain, and Abbeel]{DenoisingDiffusionProbabilisticho2020}
Jonathan Ho, Ajay Jain, and Pieter Abbeel.
\newblock Denoising {{Diffusion Probabilistic Models}}.
\newblock In \emph{Advances in {{Neural Information Processing Systems}}}, pages 6840--6851. Curran Associates, Inc., 2020.

\bibitem[Huang and LeCun(2006)]{LargescaleLearningSVMhuang2006}
Fu~Jie Huang and Y. LeCun.
\newblock Large-scale {{Learning}} with {{SVM}} and {{Convolutional}} for {{Generic Object Categorization}}.
\newblock In \emph{2006 {{IEEE Computer Society Conference}} on {{Computer Vision}} and {{Pattern Recognition}} ({{CVPR}}'06)}, pages 284--291, 2006.

\bibitem[Jarrett et~al.(2009)Jarrett, Kavukcuoglu, Ranzato, and LeCun]{WhatBestMultistagejarrett2009}
Kevin Jarrett, Koray Kavukcuoglu, Marc'Aurelio Ranzato, and Yann LeCun.
\newblock What is the best multi-stage architecture for object recognition?
\newblock In \emph{2009 {{IEEE}} 12th {{International Conference}} on {{Computer Vision}}}, pages 2146--2153, 2009.

\bibitem[Jiang et~al.(2021)Jiang, Zhang, and Timofte]{FlexibleBlindJPEGjiang2021}
Jiaxi Jiang, Kai Zhang, and Radu Timofte.
\newblock Towards {{Flexible Blind JPEG Artifacts Removal}}.
\newblock In \emph{Proceedings of the {{IEEE}}/{{CVF International Conference}} on {{Computer Vision}}}, pages 4997--5006, 2021.

\bibitem[Karras et~al.(2019)Karras, Laine, and Aila]{StyleBasedGeneratorArchitecturekarras2019}
Tero Karras, Samuli Laine, and Timo Aila.
\newblock A {{Style-Based Generator Architecture}} for {{Generative Adversarial Networks}}.
\newblock In \emph{Proceedings of the {{IEEE}}/{{CVF Conference}} on {{Computer Vision}} and {{Pattern Recognition}}}, pages 4401--4410, 2019.

\bibitem[Kawar et~al.(2022)Kawar, Elad, Ermon, and Song]{DenoisingDiffusionRestorationkawar2022}
Bahjat Kawar, Michael Elad, Stefano Ermon, and Jiaming Song.
\newblock Denoising {{Diffusion Restoration Models}}.
\newblock \emph{Advances in Neural Information Processing Systems}, 35:\penalty0 23593--23606, 2022.

\bibitem[Kim et~al.(2024)Kim, Park, Chung, and Ye]{RegularizationTextsLatentkim2024}
Jeongsol Kim, Geon~Yeong Park, Hyungjin Chung, and Jong~Chul Ye.
\newblock Regularization by {{Texts}} for {{Latent Diffusion Inverse Solvers}}, 2024.

\bibitem[Kingma and Ba(2017)]{AdamMethodStochastickingma2017}
Diederik~P. Kingma and Jimmy Ba.
\newblock Adam: {{A Method}} for {{Stochastic Optimization}}, 2017.

\bibitem[Kingma and Welling(2022)]{AutoEncodingVariationalBayeskingma2022}
Diederik~P. Kingma and Max Welling.
\newblock Auto-{{Encoding Variational Bayes}}, 2022.

\bibitem[Krizhevsky et~al.(2012)Krizhevsky, Sutskever, and Hinton]{ImageNetClassificationDeepkrizhevsky2012}
Alex Krizhevsky, Ilya Sutskever, and Geoffrey~E Hinton.
\newblock {{ImageNet Classification}} with {{Deep Convolutional Neural Networks}}.
\newblock In \emph{Advances in {{Neural Information Processing Systems}}}. Curran Associates, Inc., 2012.

\bibitem[Li et~al.(2023)Li, Zhang, Liang, Cao, Liu, Gong, Zhang, Tang, Liu, Demandolx, Ranjan, Timofte, and Van~Gool]{LSDIRLargeScaleli2023}
Yawei Li, Kai Zhang, Jingyun Liang, Jiezhang Cao, Ce Liu, Rui Gong, Yulun Zhang, Hao Tang, Yun Liu, Denis Demandolx, Rakesh Ranjan, Radu Timofte, and Luc Van~Gool.
\newblock {{LSDIR}}: {{A Large Scale Dataset}} for {{Image Restoration}}.
\newblock In \emph{2023 {{IEEE}}/{{CVF Conference}} on {{Computer Vision}} and {{Pattern Recognition Workshops}} ({{CVPRW}})}, pages 1775--1787, Vancouver, BC, Canada, 2023. IEEE.

\bibitem[Liang et~al.(2021)Liang, Cao, Sun, Zhang, Van~Gool, and Timofte]{SwinIRImageRestorationliang2021}
Jingyun Liang, Jiezhang Cao, Guolei Sun, Kai Zhang, Luc Van~Gool, and Radu Timofte.
\newblock {{SwinIR}}: {{Image Restoration Using Swin Transformer}}.
\newblock In \emph{Proceedings of the {{IEEE}}/{{CVF International Conference}} on {{Computer Vision}}}, pages 1833--1844, 2021.

\bibitem[Lin et~al.(2014)Lin, Maire, Belongie, Hays, Perona, Ramanan, Doll{\'a}r, and Zitnick]{MicrosoftCOCOCommonlin2014}
Tsung-Yi Lin, Michael Maire, Serge Belongie, James Hays, Pietro Perona, Deva Ramanan, Piotr Doll{\'a}r, and C.~Lawrence Zitnick.
\newblock Microsoft {{COCO}}: {{Common Objects}} in {{Context}}.
\newblock In \emph{Computer {{Vision}} -- {{ECCV}} 2014}, pages 740--755, Cham, 2014. Springer International Publishing.

\bibitem[Loshchilov and Hutter(2018)]{DecoupledWeightDecayloshchilov2018}
Ilya Loshchilov and Frank Hutter.
\newblock Decoupled {{Weight Decay Regularization}}.
\newblock In \emph{International {{Conference}} on {{Learning Representations}}}, 2018.

\bibitem[Luo et~al.(2024)Luo, Darrell, Wang, Goldman, and Holynski]{ReadoutGuidanceLearningluo2024}
Grace Luo, Trevor Darrell, Oliver Wang, Dan~B. Goldman, and Aleksander Holynski.
\newblock Readout {{Guidance}}: {{Learning Control}} from {{Diffusion Features}}.
\newblock In \emph{Proceedings of the {{IEEE}}/{{CVF Conference}} on {{Computer Vision}} and {{Pattern Recognition}}}, pages 8217--8227, 2024.

\bibitem[Man et~al.(2023)Man, Ohayon, Adrai, and Elad]{HighPerceptualQualityJPEGman2023}
Sean Man, Guy Ohayon, Theo Adrai, and Michael Elad.
\newblock High-{{Perceptual Quality JPEG Decoding}} via {{Posterior Sampling}}.
\newblock In \emph{Proceedings of the {{IEEE}}/{{CVF Conference}} on {{Computer Vision}} and {{Pattern Recognition}}}, pages 1272--1282, 2023.

\bibitem[Mardani et~al.(2023)Mardani, Song, Kautz, and Vahdat]{VariationalPerspectiveSolvingmardani2023}
Morteza Mardani, Jiaming Song, Jan Kautz, and Arash Vahdat.
\newblock A {{Variational Perspective}} on {{Solving Inverse Problems}} with {{Diffusion Models}}.
\newblock In \emph{The {{Twelfth International Conference}} on {{Learning Representations}}}, 2023.

\bibitem[Ohayon et~al.(2021)Ohayon, Adrai, Vaksman, Elad, and Milanfar]{HighPerceptualQualityohayon2021}
Guy Ohayon, Theo Adrai, Gregory Vaksman, Michael Elad, and Peyman Milanfar.
\newblock High {{Perceptual Quality Image Denoising With}} a {{Posterior Sampling CGAN}}.
\newblock In \emph{Proceedings of the {{IEEE}}/{{CVF International Conference}} on {{Computer Vision}}}, pages 1805--1813, 2021.

\bibitem[Radford et~al.(2021)Radford, Kim, Hallacy, Ramesh, Goh, Agarwal, Sastry, Askell, Mishkin, Clark, Krueger, and Sutskever]{LearningTransferableVisualradford2021}
Alec Radford, Jong~Wook Kim, Chris Hallacy, Aditya Ramesh, Gabriel Goh, Sandhini Agarwal, Girish Sastry, Amanda Askell, Pamela Mishkin, Jack Clark, Gretchen Krueger, and Ilya Sutskever.
\newblock Learning {{Transferable Visual Models From Natural Language Supervision}}.
\newblock In \emph{Proceedings of the 38th {{International Conference}} on {{Machine Learning}}}, pages 8748--8763. PMLR, 2021.

\bibitem[Rodgers(2000)]{InverseMethodsAtmosphericrodgers2000}
Clive~D Rodgers.
\newblock Inverse {{Methods}} for {{Atmospheric Sounding}}: {{Theory}} and {{Practice}}.
\newblock In \emph{Inverse Methods for Atmospheric Sounding: Theory and Practice}. WORLD SCIENTIFIC, 2000.

\bibitem[Romano et~al.(2017)Romano, Elad, and Milanfar]{LittleEngineThatromano2017}
Yaniv Romano, Michael Elad, and Peyman Milanfar.
\newblock The {{Little Engine That Could}}: {{Regularization}} by {{Denoising}} ({{RED}}).
\newblock \emph{SIAM Journal on Imaging Sciences}, 10\penalty0 (4):\penalty0 1804--1844, 2017.

\bibitem[Rombach et~al.(2022)Rombach, Blattmann, Lorenz, Esser, and Ommer]{HighResolutionImageSynthesisrombach2022}
Robin Rombach, Andreas Blattmann, Dominik Lorenz, Patrick Esser, and Bj{\"o}rn Ommer.
\newblock High-{{Resolution Image Synthesis With Latent Diffusion Models}}.
\newblock In \emph{Proceedings of the {{IEEE}}/{{CVF Conference}} on {{Computer Vision}} and {{Pattern Recognition}}}, pages 10684--10695, 2022.

\bibitem[Rout et~al.(2023)Rout, Raoof, Daras, Caramanis, Dimakis, and Shakkottai]{SolvingLinearInverserout2023}
Litu Rout, Negin Raoof, Giannis Daras, Constantine Caramanis, Alexandros~G. Dimakis, and Sanjay Shakkottai.
\newblock Solving {{Linear Inverse Problems Provably}} via {{Posterior Sampling}} with {{Latent Diffusion Models}}, 2023.

\bibitem[Rout et~al.(2024)Rout, Chen, Kumar, Caramanis, Shakkottai, and Chu]{FirstOrderTweedieSolvingrout2024}
Litu Rout, Yujia Chen, Abhishek Kumar, Constantine Caramanis, Sanjay Shakkottai, and Wen-Sheng Chu.
\newblock Beyond {{First-Order Tweedie}}: {{Solving Inverse Problems}} using {{Latent Diffusion}}.
\newblock In \emph{Proceedings of the {{IEEE}}/{{CVF Conference}} on {{Computer Vision}} and {{Pattern Recognition}}}, pages 9472--9481, 2024.

\bibitem[Saharia et~al.(2022)Saharia, Chan, Saxena, Li, Whang, Denton, Ghasemipour, Gontijo~Lopes, Karagol~Ayan, Salimans, Ho, Fleet, and Norouzi]{PhotorealisticTextImageDiffusionsaharia2022}
Chitwan Saharia, William Chan, Saurabh Saxena, Lala Li, Jay Whang, Emily~L. Denton, Kamyar Ghasemipour, Raphael Gontijo~Lopes, Burcu Karagol~Ayan, Tim Salimans, Jonathan Ho, David~J. Fleet, and Mohammad Norouzi.
\newblock Photorealistic {{Text-to-Image Diffusion Models}} with {{Deep Language Understanding}}.
\newblock \emph{Advances in Neural Information Processing Systems}, 35:\penalty0 36479--36494, 2022.

\bibitem[SG161222()]{StablediffusionapiRealisticvision51Huggingsg161222}
Evgeny SG161222.
\newblock Stablediffusionapi/realistic-vision-51 {$\cdot$} {{Hugging Face}}.
\newblock https://huggingface.co/stablediffusionapi/realistic-vision-51.

\bibitem[Simonyan and Zisserman(2015)]{VeryDeepConvolutionalsimonyan2015}
Karen Simonyan and Andrew Zisserman.
\newblock Very {{Deep Convolutional Networks}} for {{Large-Scale Image Recognition}}, 2015.

\bibitem[{Sohl-Dickstein} et~al.(2015){Sohl-Dickstein}, Weiss, Maheswaranathan, and Ganguli]{DeepUnsupervisedLearningsohl-dickstein2015}
Jascha {Sohl-Dickstein}, Eric Weiss, Niru Maheswaranathan, and Surya Ganguli.
\newblock Deep {{Unsupervised Learning}} using {{Nonequilibrium Thermodynamics}}.
\newblock In \emph{Proceedings of the 32nd {{International Conference}} on {{Machine Learning}}}, pages 2256--2265. PMLR, 2015.

\bibitem[Song et~al.(2023)Song, Kwon, Zhang, Hu, Qu, and Shen]{SolvingInverseProblemssong2023}
Bowen Song, Soo~Min Kwon, Zecheng Zhang, Xinyu Hu, Qing Qu, and Liyue Shen.
\newblock Solving {{Inverse Problems}} with {{Latent Diffusion Models}} via {{Hard Data Consistency}}.
\newblock In \emph{The {{Twelfth International Conference}} on {{Learning Representations}}}, 2023.

\bibitem[Song and Ermon(2019{\natexlab{a}})]{GenerativeModelingEstimatingsong2019}
Yang Song and Stefano Ermon.
\newblock Generative {{Modeling}} by {{Estimating Gradients}} of the {{Data Distribution}}.
\newblock In \emph{Advances in {{Neural Information Processing Systems}}}. Curran Associates, Inc., 2019{\natexlab{a}}.

\bibitem[Song and Ermon(2019{\natexlab{b}})]{GenerativeModelingEstimatingsong2019a}
Yang Song and Stefano Ermon.
\newblock Generative {{Modeling}} by {{Estimating Gradients}} of the {{Data Distribution}}.
\newblock \emph{Advances in Neural Information Processing Systems}, 32, 2019{\natexlab{b}}.

\bibitem[Song et~al.(2020)Song, {Sohl-Dickstein}, Kingma, Kumar, Ermon, and Poole]{ScoreBasedGenerativeModelingsong2020}
Yang Song, Jascha {Sohl-Dickstein}, Diederik~P. Kingma, Abhishek Kumar, Stefano Ermon, and Ben Poole.
\newblock Score-{{Based Generative Modeling}} through {{Stochastic Differential Equations}}.
\newblock In \emph{International {{Conference}} on {{Learning Representations}}}, 2020.

\bibitem[Song et~al.(2021)Song, Shen, Xing, and Ermon]{SolvingInverseProblemssong2021}
Yang Song, Liyue Shen, Lei Xing, and Stefano Ermon.
\newblock Solving {{Inverse Problems}} in {{Medical Imaging}} with {{Score-Based Generative Models}}.
\newblock In \emph{International {{Conference}} on {{Learning Representations}}}, 2021.

\bibitem[Venkatakrishnan et~al.(2013)Venkatakrishnan, Bouman, and Wohlberg]{PlugPlayPriorsModelvenkatakrishnan2013}
Singanallur~V. Venkatakrishnan, Charles~A. Bouman, and Brendt Wohlberg.
\newblock Plug-and-{{Play}} priors for model based reconstruction.
\newblock In \emph{2013 {{IEEE Global Conference}} on {{Signal}} and {{Information Processing}}}, pages 945--948, 2013.

\bibitem[Vincent(2011)]{ConnectionScoreMatchingvincent2011}
Pascal Vincent.
\newblock A {{Connection Between Score Matching}} and {{Denoising Autoencoders}}.
\newblock \emph{Neural Computation}, 23\penalty0 (7):\penalty0 1661--1674, 2011.

\bibitem[Wang et~al.(2024)Wang, Bai, Tan, Wang, Fan, Bai, Chen, Liu, Wang, Ge, Fan, Dang, Du, Ren, Men, Liu, Zhou, Zhou, and Lin]{Qwen2VLEnhancingVisionLanguagewang2024}
Peng Wang, Shuai Bai, Sinan Tan, Shijie Wang, Zhihao Fan, Jinze Bai, Keqin Chen, Xuejing Liu, Jialin Wang, Wenbin Ge, Yang Fan, Kai Dang, Mengfei Du, Xuancheng Ren, Rui Men, Dayiheng Liu, Chang Zhou, Jingren Zhou, and Junyang Lin.
\newblock Qwen2-{{VL}}: {{Enhancing Vision-Language Model}}'s {{Perception}} of the {{World}} at {{Any Resolution}}, 2024.

\bibitem[Wang et~al.(2018)Wang, Yu, Wu, Gu, Liu, Dong, Qiao, and Change~Loy]{ESRGANEnhancedSuperResolutionwang2018}
Xintao Wang, Ke Yu, Shixiang Wu, Jinjin Gu, Yihao Liu, Chao Dong, Yu Qiao, and Chen Change~Loy.
\newblock {{ESRGAN}}: {{Enhanced Super-Resolution Generative Adversarial Networks}}.
\newblock In \emph{Proceedings of the {{European Conference}} on {{Computer Vision}} ({{ECCV}}) {{Workshops}}}, pages 0--0, 2018.

\bibitem[Wang et~al.(2022{\natexlab{a}})Wang, Yu, and Zhang]{ZeroShotImageRestorationwang2022}
Yinhuai Wang, Jiwen Yu, and Jian Zhang.
\newblock Zero-{{Shot Image Restoration Using Denoising Diffusion Null-Space Model}}.
\newblock In \emph{The {{Eleventh International Conference}} on {{Learning Representations}}}, 2022{\natexlab{a}}.

\bibitem[Wang et~al.(2022{\natexlab{b}})Wang, Zhang, Chen, Wang, and Luo]{RestoreFormerHighQualityBlindwang2022}
Zhouxia Wang, Jiawei Zhang, Runjian Chen, Wenping Wang, and Ping Luo.
\newblock {{RestoreFormer}}: {{High-Quality Blind Face Restoration From Undegraded Key-Value Pairs}}.
\newblock In \emph{Proceedings of the {{IEEE}}/{{CVF Conference}} on {{Computer Vision}} and {{Pattern Recognition}}}, pages 17512--17521, 2022{\natexlab{b}}.

\bibitem[Zeyde et~al.(2012)Zeyde, Elad, and Protter]{SingleImageScaleUsingzeyde2012}
Roman Zeyde, Michael Elad, and Matan Protter.
\newblock On {{Single Image Scale-Up Using Sparse-Representations}}.
\newblock In \emph{Curves and {{Surfaces}}}, pages 711--730, Berlin, Heidelberg, 2012. Springer.

\bibitem[Zhang et~al.(2017)Zhang, Zuo, Chen, Meng, and Zhang]{GaussianDenoiserResidualzhang2017}
Kai Zhang, Wangmeng Zuo, Yunjin Chen, Deyu Meng, and Lei Zhang.
\newblock Beyond a {{Gaussian Denoiser}}: {{Residual Learning}} of {{Deep CNN}} for {{Image Denoising}}.
\newblock \emph{IEEE Transactions on Image Processing}, 26\penalty0 (7):\penalty0 3142--3155, 2017.

\bibitem[Zhang et~al.(2018)Zhang, Isola, Efros, Shechtman, and Wang]{UnreasonableEffectivenessDeepzhang2018}
Richard Zhang, Phillip Isola, Alexei~A. Efros, Eli Shechtman, and Oliver Wang.
\newblock The {{Unreasonable Effectiveness}} of {{Deep Features}} as a {{Perceptual Metric}}.
\newblock In \emph{Proceedings of the {{IEEE Conference}} on {{Computer Vision}} and {{Pattern Recognition}}}, pages 586--595, 2018.

\bibitem[Zhang et~al.(2018 https://github.com/richzhang/PerceptualSimilarity)Zhang, Isola, Efros, Shechtman, and Wang]{PerceptualSimilarityzhang2018}
Richard Zhang, Phillip Isola, Alexei~A Efros, Eli Shechtman, and Oliver Wang.
\newblock {{PerceptualSimilarity}}, 2018 https://github.com/richzhang/PerceptualSimilarity.

\bibitem[Zhu et~al.(2023)Zhu, Zhang, Liang, Cao, Wen, Timofte, and Van~Gool]{DenoisingDiffusionModelszhu2023}
Yuanzhi Zhu, Kai Zhang, Jingyun Liang, Jiezhang Cao, Bihan Wen, Radu Timofte, and Luc Van~Gool.
\newblock Denoising {{Diffusion Models}} for {{Plug-and-Play Image Restoration}}.
\newblock In \emph{Proceedings of the {{IEEE}}/{{CVF Conference}} on {{Computer Vision}} and {{Pattern Recognition}}}, pages 1219--1229, 2023.

\end{thebibliography}

}

\clearpage
\clearpage
\appendix
\setcounter{page}{1}
\onecolumn
\maketitlesupplementary

\section{Flaws of bi-domain LDM restoration}\label{app:discuss_option_1}

The MMSE denoiser employed in the diffusion process of LDMs is designed to denoise latents, not images, while the observation $y$ resides in the measurement space.
Thus, the approximation of DPS,
\begin{equation}\label{eq:DPS-app}
    \nabla_{x_t} \ln{p(y|x_t)} \approx -\frac{1}{\sigma_y^2}\nabla_{x_t} \|y-\mathcal{A}(\hat{x}^t_0)\|_2^2,
\end{equation}
is no longer applicable.
As mentioned in \cref{sec:method}, one possible solution is to decode the latents to the pixel (measurement) space during the restoration process, allowing, for example, the use of 
\begin{equation}\label{eq:DPS-app-LDM}
    \nabla_{z_t} \ln{p(y|z_t)} \approx -\frac{1}{\sigma_y^2}\nabla_{z_t} \|y-\mathcal{A}(\mathcal{D}(\hat{z}^t_0))\|_2^2.
\end{equation} 
We categorize such solutions as part of the ``bi-domain'' family, as they compute gradients in both the pixel and latent domains.
To the best of our knowledge, all existing methods leveraging LDMs for inverse problems (apart from SILO) fall within this category.
In this section, we demonstrate the underlying flaws in solutions belonging to this family.

During the training process, the decoder is exposed only to clean latents, $z_0$.
Moreover, we do not require that its Jacobian be informative or well-behaved.
This leads to two main problems when using the decoder during restoration. First, the decoder is applied to out-of-distribution (OOD) latents compared to its training data, as the latents in the restoration process, $\hat{z}^t_0$, are MMSE denoised, thus coming from a different distribution.
Second, differentiating through the decoder transforms the score-likelihood gradient from the pixel to the latent space, introducing a possibly uninformative Jacobian to the backpropagation process.
These two problems are tightly related to each other. An OOD latent leads to an unpredictable Jacobian, further destabilizing the differentiation process of the likelihood.

To demonstrate these problems, we focus on LDPS~\cite{SolvingInverseProblemssong2023} and PSLD~\cite{SolvingLinearInverserout2023} as representatives of the bi-domain family.
LDPS (\cref{eq:LDPS}) is the starting ground for all other methods in this family, and PSLD (\cref{eq:psld}) is an extension of it.
As we see in \cref{fig:app_coco,fig:app_gb,fig:app_ip,fig:app_jpeg,fig:app_sr4,fig:app_sr8_noise,fig:app_sr8}, the reconstructions of LDPS and its derivatives often suffer from the presence of ``blob'' artifacts and noise patterns.
To investigate this matter, we record the gradients
\begin{equation}\label{eq:ldps_grad_z_0}
    \nabla_{\hat{z}^t_0} \|y-\mathcal{A}(\mathcal{D}(\hat{z}^t_0))\|_2^2
\end{equation}
during the restoration process of LDPS and PSLD. 
Note the subtle difference between  \cref{eq:ldps_grad_z_0} and \cref{eq:DPS-app-LDM};  the two differ only in the Jacobian of the denoiser, which is irrelevant to our analysis as it is independent of the decoder. %

In \cref{fig:option_1_demon}, we see that from the early stages of the restoration process, these gradients contain a patch that causes the latent to change in a way that does not correlate with the measurement. 
This effect prevails throughout the diffusion process, leading to a closely-related defect in the resulting image.
Qualitatively, from $t \approx 500$ onward, the gradients exhibit a noise pattern that dominates the signal, leading to a similar noise pattern in the reconstructed image. %
This behavior is inherent to the use of the decoder in the way practiced by LDPS and PSLD. 
Adding projections and a regularization could improve the restoration, but only to some extent. 
For example, PSLD attempts to mitigate this by guiding the latents to areas the encoder-decoder handles better, yet similar artifacts are still presented as seen in \cref{fig:option_1_demon}.
Another example is ReSample, which performs likelihood optimization in pixel space followed by an encoding step in parts of the restoration process to avoid those blob artifacts (Appendix B of ~\cite{SolvingInverseProblemssong2023}).

In summary, differentiating through the decoder might corrupt the information required for faithful reconstruction.
This motivates us to avoid using the decoder altogether during the restoration process.

\begin{figure*}[hbt]
  \centering
 \includegraphics[width=\linewidth]{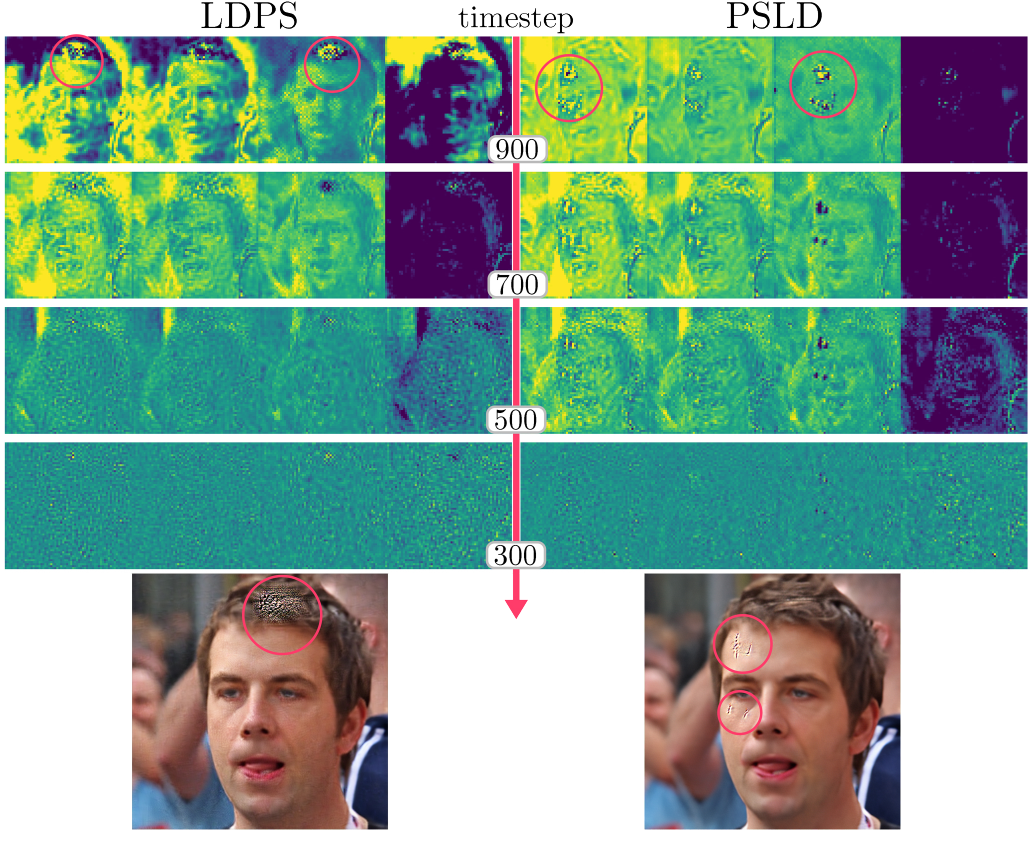}
\caption{\textbf{The decoder's artifacts.}
We present the likelihood's gradient through the decoder (\cref{eq:ldps_grad_z_0}) at different timesteps $t$. Each gradient is presented as four single-channel images, clipped to $[-3,3]$ and scaled by $10^2$ for better visualization. At the bottom, we present the resulting image of the restoration process. The decoder's Jacobian introduces ``blob'' artifacts and noise patterns that are present in the final restorations.
}
\label{fig:option_1_demon} 
\end{figure*}

\section{Diverse reconstructions using SILO}

Similar to DPS~\cite{DiffusionPosteriorSamplingchung2022}, SILO is a stochastic solver of inverse problems.
To demonstrate the effect of this stochasticity, we present in \cref{fig:stochastic} reconstruction examples using SILO for the box inpainting task.
We present multiple reconstruction per input, each is the result of a different random seed. We see large variability in the reconstructions, while keeping the consistency intact.
\begin{figure*}[bt]
  \centering
 \includegraphics[width=\linewidth]{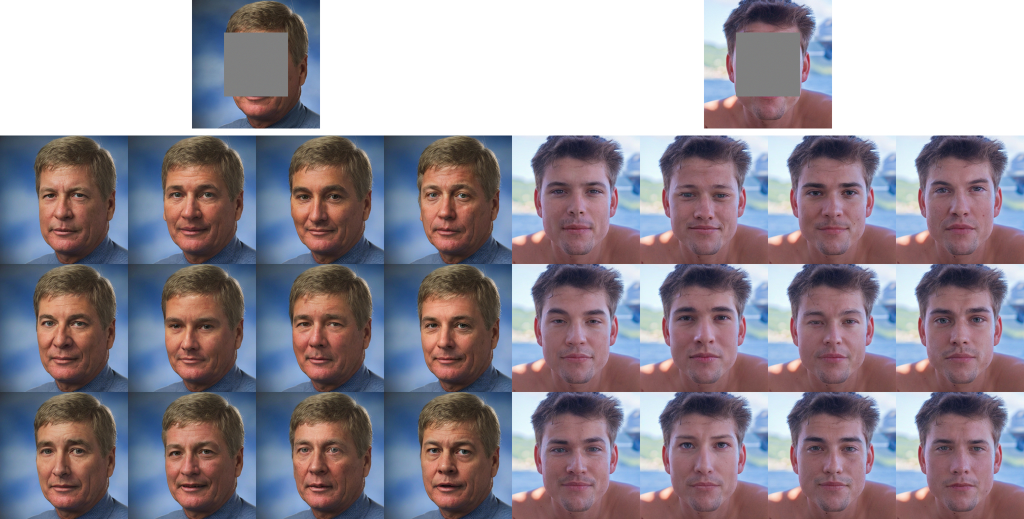}
\caption{\textbf{Diverse reconstructions.} We present two measurements from FFHQ corrupted by a box mask at the top. Below each measurement are several reconstructions using SILO. The reconstructions' hyperparameters are identical over all the images except for the random seed used. We see great variability in the details reconstructed inside the missing box, all while being consistent outside of it.}
\label{fig:stochastic} 
\end{figure*}
\section{Implementation details}\label{app:implementation}

\subsection{General details}

The degradations described in \cref{sec:Degradations} are done using the original code base\footnote{github.com/DPS2022/diffusion-posterior-sampling} of DPS \cite{DiffusionPosteriorSamplingchung2022}.
For JPEG, we use the publicly available implementation from kornia\footnote{kornia.readthedocs.io/en/latest/enhance.html\#kornia.enhance.\\jpeg\_codec\_differentiable}.
For the diffusion models, we use Stable Diffusion v1.5\footnote{huggingface.co/botp/stable-diffusion-v1-5} (denoted as SD or SD-v1.5) and Realistic Vision v5.1\footnote{huggingface.co/stablediffusionapi/realistic-vision-v51} (denoted as RV or Rv-v5.1).
In order to generate the HQ captions from \cref{sec:hq_prompt_prior_ablation}, we use Qwen2-VL-7B-Instruct \footnote{https://huggingface.co/Qwen/Qwen2-VL-7B-Instruct}.

\subsection{Metrics}
The metrics we use are divided into 2 groups: distortion and perception metrics.

\paragraph{Distortion Metrics.}
Distortion metrics are calculated between two images.
\noindent \textbf{PSNR} evaluates how close a reconstruction, $\hat{x}$, is to the original image, $x$,
\begin{equation}
    \label{eq:psnr}
    \text{PSNR}(x,\hat{x}) = 10 \text{log}_{10}{\left(\frac{2^2}{\text{mean}\left(\|x-\hat{x} \|^2_2\right)}\right)}.
\end{equation}
Since the images can have values in the range $[-1,1]$, $2$ is used as the data range to calculate the PSNR. The denominator transfers 
$\|x-\hat{x} \|^2_2$ to a per-pixel error via the $\text{`mean'}$ operation.  
Similarly, \textbf{CPSNR} evaluates whether a reconstruction, $\hat{x}$, is consistent with the measurement, meaning it could have been the underlying signal that created it. We define it by
\begin{equation}
\label{eq:cpsnr}
\text{CPSNR}(x,\hat{x}) = \text{PSNR}(\mathcal{A}(x),\mathcal{A}(\hat{x})).   
\end{equation}
\textbf{LPIPS} measures the perceptual similarity between two images and can be computed using different neural networks. 
As recommended by \cite{PerceptualSimilarityzhang2018}, we use AlexNet \cite{ImageNetClassificationDeepkrizhevsky2012} for evaluation, as it is preferred over VGG \cite{VeryDeepConvolutionalsimonyan2015}.
Unlike prior work, which primarily reports LPIPS-VGG, we present LPIPS-Alex in \cref{sec:results} and include both LPIPS-Alex and LPIPS-VGG in \cref{app:add_results} for completeness.

\paragraph{Perception Metrics.}
Perception metrics assess the divergence between the distribution of real images, $p_x$, and the distribution of reconstructed images, $p_{\hat{x}}$.
When these distributions are close, it indicates that our algorithm approximately samples from the real image distribution.
The perception measures we provide are Fréchet Inception Distance (\textbf{FID}) \cite{GANsTrainedTwoheusel2017}, and Kernel Inception Distance (\textbf{KID}) \cite{DemystifyingMMDGANsbinkowski2018} multiplied by a factor of $10^3$.

\subsection{Hyperparameters of results in paper}

In this subsection, we detail the exact hyperparameters used for all results presented in the paper.
Reproducing a reconstruction involves using the eight variables listed below along with the same dataset.
The information provided here, combined with the details in the tables or figures in the paper, ensures that all necessary information for reproducing the results is available.
\begin{itemize}
\item \textbf{seed:} The random seed to the process is fixed to $1,000$ throughout the paper unless otherwise mentioned.

\item \textbf{Model:} Either SD or RV, as defined in \cref{app:implementation}.

\item \textbf{idx:} The index of the image from the given dataset, when the count starts at 0.

\item \textbf{prompt:} If the dataset is FFHQ, then the options are null, general and HQ. They are defined in \cref{sec:results} and the HQ prompts are given in the supplementary material. 
If the dataset is COCO, we only use a null prompt. One can use ``A high quality photo'' as a general prompt, but we did not experiment with it.

\item \textbf{CFG:} Classifier-free guidance \cite{ClassifierFreeDiffusionGuidanceho2021} allows for reconstruction that adhere more to the given prompt. We use either 1 (equal to not performing CFG) or 4.

\item $\bm{\mathcal{A}:}$ The degradation operator, as defined in \cref{sec:Degradations}.
In the appendix, we use the following abbreviations: GB for Gaussian blur, IP for inpaining, JP for JPEG, SR8 for Super-resolution $\times8$, and SR4 for Super-resolution $\times4$.

\item $\bm{\sigma_y:}$ The amount of noise added to create the measurements was either $0.01$ or $0.03$ in all experiments. Note that in our code, these values are doubled ($0.02$ and $0.06$, respectively) because the noise is added after the images are normalized to the range $[-1, 1]$.

\item $\bm{\eta:}$ This parameter determines the scale (\ie, step size) of the guidance term.
Higher scales result in more consistent reconstructions at the expense of perceptual quality.
We set $\eta=0.5$ for all tasks, except for inpainting, where $\eta=1$ is used.
\end{itemize}
For all the results of SILO (denoted as ``Ours'') in \cref{tab:results_of_sr4_and_gb,tab:results_of_sr8_and_ip} we use a CFG of 1.
For \cref{tab:coco}, we use, $\sigma_y=0.01$, $\eta=1$, null prompt and CFG of 1. For \cref{tab:cnn}, the settings are the same as for \cref{tab:results_of_sr8_and_ip}.%
The sampling parameters of \cref{fig:first} are presented in \cref{tab:first_hp}, of \cref{fig:ffhq comparison} in \cref{tab:ffhq_comparison_hp} and of \cref{fig:coco_fig} in \cref{tab:coco_fig_hp}.
\begin{table}[tb]
    \centering
    \begin{tabular}{cccccccc}
         Row & idx& $\mathcal{A}$ & $\sigma_y$  & $\eta$ & prompt & CFG & model\\
         \hline
         1&629& IP & 0.01 & 1 & general & 1 & RV\\
         2&647& SR$8$ & 0.03 & 0.5 & general & 4 & RV\\
         3&982& JP & 0.01 & 0.5 & general & 1 & RV \\
    \end{tabular}
    \caption{Hyperparameters used in \cref{fig:first}}
    \label{tab:first_hp}
\end{table}

\begin{table}[tb]
    \centering
    \begin{tabular}{ccccccc}
         Row &idx& $\mathcal{A}$ & $\sigma_y$  & $\eta$ & prompt & CFG \\
         \hline
         1&713& IP & 0.01 & 1 & general & 1\\
         2&880& SR $8$ & 0.01 & 0.5 & general & 1 \\
         3&485& GB & 0.01 & 0.5 & general & 1 \\
    \end{tabular}
    \caption{Hyperparameters used in \cref{fig:ffhq comparison}, the model is specified at the top of the figure.}
    \label{tab:ffhq_comparison_hp}
\end{table}

\begin{table}[tb]
    \centering
    \begin{tabular}{cccccccc}
         Row &idx& $\mathcal{A}$ & $\sigma_y$  & $\eta$ & prompt & CFG & model \\
         \hline
         1&130& IP  & 0.01 & 1 & null & 1 & SD\\
         2&491& IP  & 0.01 & 1 & null & 1 & SD\\
    \end{tabular}
    \caption{Hyperparameters used in \cref{fig:coco_fig}.}
    \label{tab:coco_fig_hp}
\end{table}

\subsection{Architecture of CNN operator}
In \cref{tab:cnn,sec:results}, we show that even when $H_\theta$ is a simple CNN, SILO still performs well on the Super-resolution ${\times}8$ task.
The architecture of the CNN is depicted in \cref{fig:cnn_arch}, and the implementation is provided in the supplementary files.

\begin{figure*}[bt]
  \centering
 \includegraphics[width=0.85\linewidth]{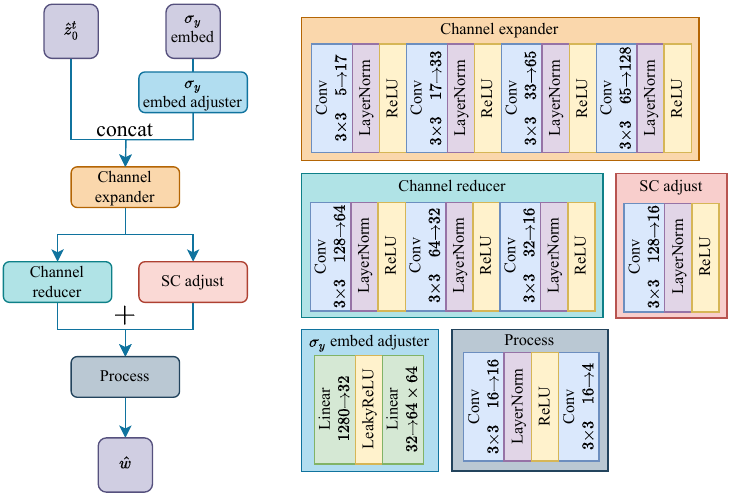}
\caption{Architecture of the CNN used to get the results of the CNN-RV row in \cref{tab:cnn}}
\label{fig:cnn_arch} 
\end{figure*}

\section{Comparison to other methods}
In this section, we describe how SILO was compared to ReSample, PSLD, GML, and LDPS.
SILO utilizes SD to generate reconstructions at a resolution of $512 \times 512$.
PSLD, GML, and LDPS natively support this resolution and diffusion prior, as implemented in the PSLD GitHub repository~\footnote{github.com/LituRout/PSLD}.
Resample use the LDM-VQ4, trained on FFHQ~\footnote{github.com/CompVis/latent-diffusion/tree/main?tab=readme-ov-file\#model-zoo} as the diffusion prior, which generates images of size $256 \times 256$.
Hence, to give a fair comparison to ReSample, we had to adapt their publicly available code.

\paragraph{PSLD.}
We made no modifications to the PSLD code, except for adapting the data-loading process to enable sampling from the COCO dataset.
The hyperparameters used were identical to those provided in the official repository for each task.
For tasks not explicitly included (JPEG and SR $\times8$), we applied the same hyperparameters as those used for the SR $\times4$ task.

\paragraph{GML-DPS.}
We used the same implementation as for PSLD, replacing the PSLD step (\cref{eq:psld}) with the GML step (\cref{eq:gml}).
The hyperparameters were identical to those used for PSLD.

\paragraph{LDPS.}
We used the same implementation as for PSLD, except that the PSLD step (\cref{eq:psld}) was omitted entirely.
This removes the associated computational requirements.
The hyperparameters remained the same as those used for PSLD.

\paragraph{ReSample.}
As mentioned earlier, the reported results for ReSample are based on $256 \times 256$ reconstructions, generated using a different diffusion prior than SD, which was trained specifically on face images.
To implement ReSample-SD, we started with the publicly available ReSample codebase~\footnote{github.com/soominkwon/resample}.
We replaced the LDM-VQ4 denoiser with the SD one, updated the Autoencoder to match the one used for SD (consistent with all other methods), and adjusted the data-loading process to handle larger images.
The hyperparameters used were identical to those in the original codebase.
We acknowledge that the results of ReSample-SD differ from the reported results in ReSample, particularly for box inpainting. This discrepancy could stem from changes in the diffusion prior and image size, as well as suboptimal hyperparameters (due to these changes). %
The authors of the original paper were contacted to discuss the discrepancies we encountered.
We should note that reconstruction time remains a significant factor -- ReSample is notably slower than SILO, which achieves speedups of $10 \times$ and $18 \times$ for SR $\times 8$ and JPEG tasks, respectively.
\section{Training}

The training scheme for the degradation operator is illustrated in \cref{fig:training}, and the training scripts are provided as supplementary material.
As a reminder, we employed a Readout Guidance~\cite{ReadoutGuidanceLearningluo2024} network for most of our experiments and included a comparison where $H_\theta$ is implemented as a CNN in \cref{tab:cnn}.
Training procedures for both setups are presented.

\subsection{Training a RG operator}

When training a Readout Guidance operator $H_\theta$, the inputs are features extracted from the denoising network.
We followed the same settings as described in RG~\cite{ReadoutGuidanceLearningluo2024}.
The learning rate was set to $2{\times}10^{-4}$, using the AdamW optimizer~\footnote{pytorch.org/docs/stable/generated/torch.optim.AdamW}~\cite{DecoupledWeightDecayloshchilov2018}.
Training was conducted on a single NVIDIA A100-SXM4-80GB card with a batch size of 16, for $7.5{\times}10^4$ steps, taking approximately 28 hours.
Training loss \vs the number of optimization steps for the Super-resolution $\times 8$ operator is shown in \cref{fig:c330fpxy_loss_plot}.
We did not optimize the training process at all.

\begin{figure}[bt]
  \centering
 \includegraphics[width=\linewidth]{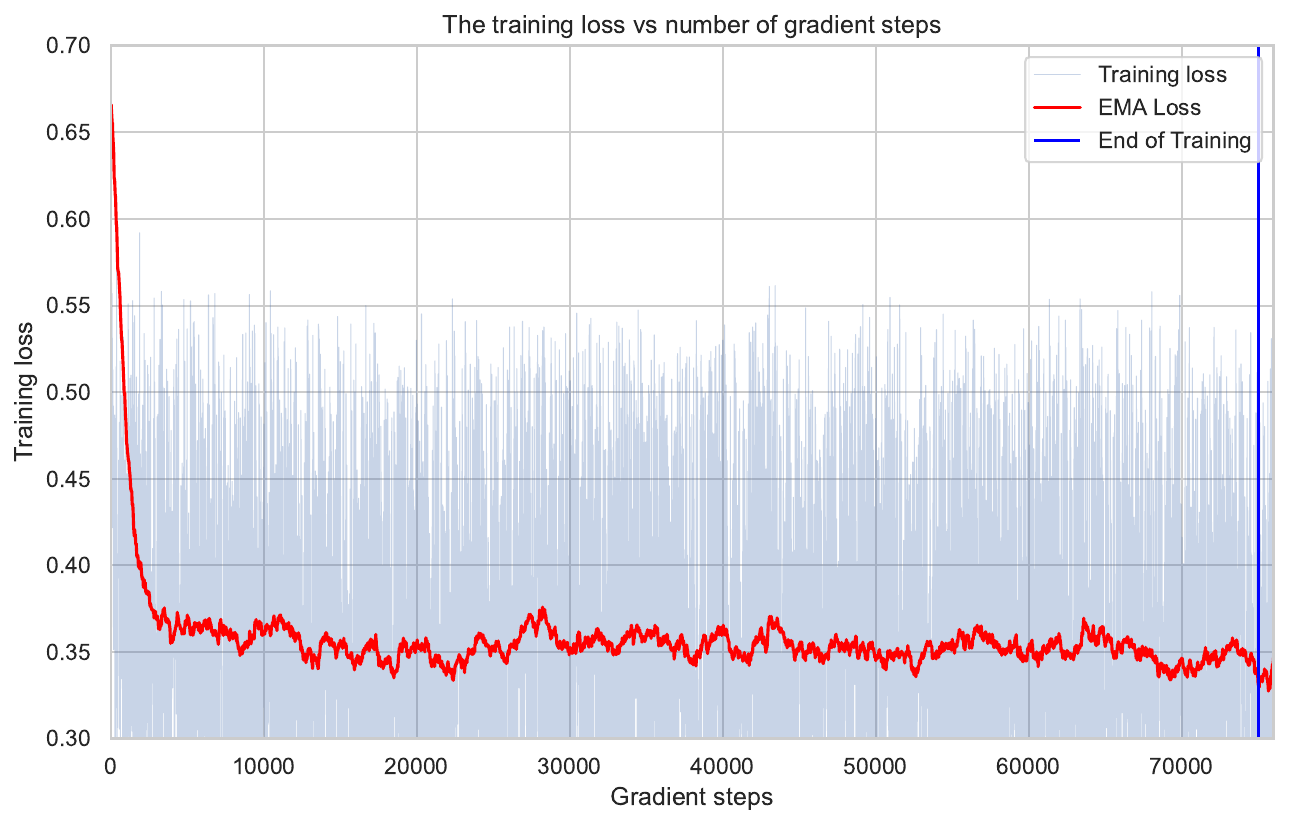}
\caption{Training loss when learning a RG $H_\theta$ to mimic the Super-resolution $\times 8$ operator.}
\label{fig:c330fpxy_loss_plot} 
\end{figure}

\subsection{Training a CNN operator}

The CNN-based $H_\theta$ takes a latent $z$ as input and produces $\hat{w}$ as output.
The training scheme for the CNN is similar to that depicted in \cref{fig:training}, with the key difference being the omission of noise addition and denoiser usage during training.
The learning rate was set to $10^{-3}$, using the Adam optimizer~\footnote{pytorch.org/docs/stable/generated/torch.optim.Adam}~\cite{AdamMethodStochastickingma2017}.
Training was conducted on a single NVIDIA L40S card with a batch size of 16, for $10^5$ steps, requiring approximately 28 hours.
Training loss \vs the number of optimization steps for the Super-resolution $\times 8$ operator is shown in \cref{fig:icy-energy-46_loss_plot}. We did not optimize the training process or network architecture at all.

\begin{figure}[!h]
  \centering
 \includegraphics[width=\linewidth]{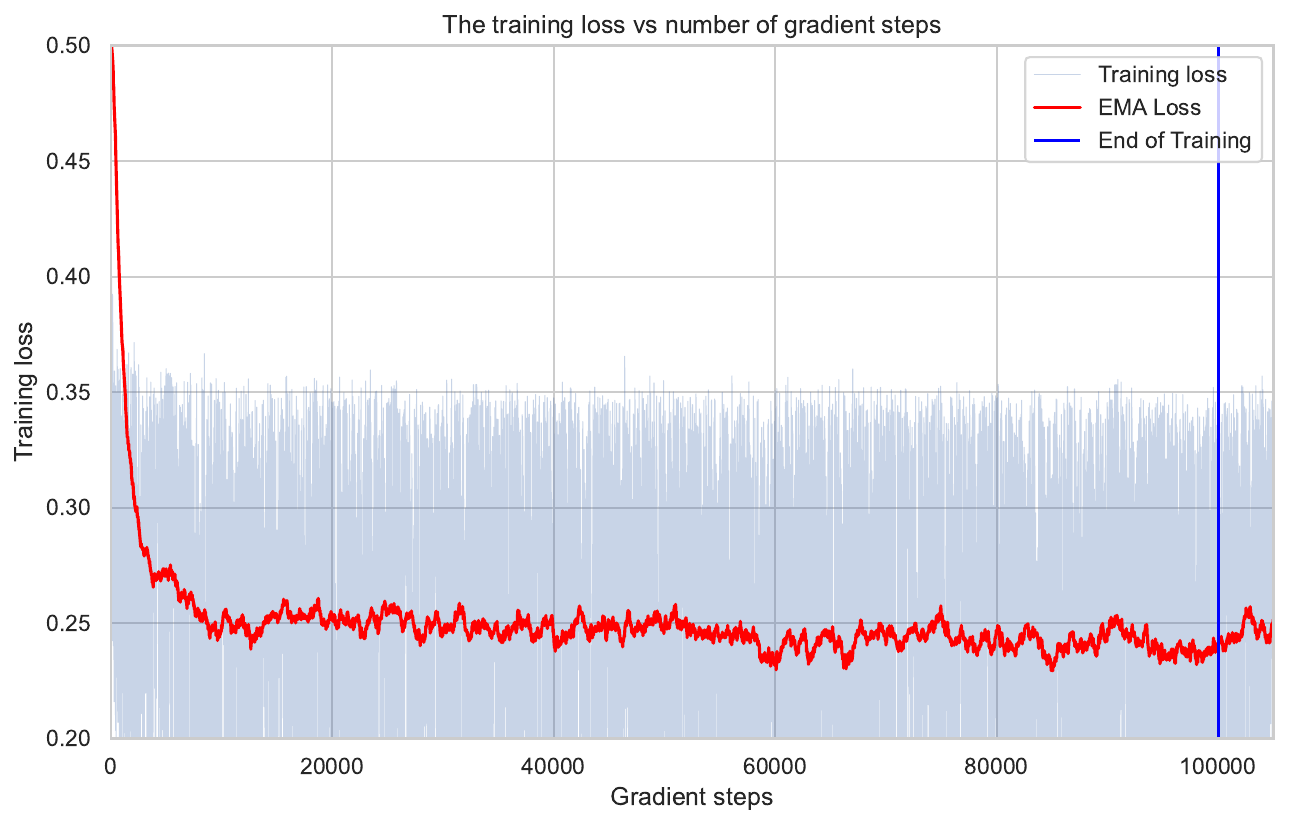}
\caption{Training loss when learning a CNN $H_\theta$ to mimic the Super-resolution $\times 8$ operator.}
\label{fig:icy-energy-46_loss_plot} 
\end{figure}

\section{Additional results}\label{app:add_results}
We reprint the results in \cref{tab:results_of_sr4_and_gb,tab:results_of_sr8_and_ip,tab:ablation_noise,tab:coco} in \cref{tab:results_of_sr4_and_gb_supp,tab:results_of_sr8_and_ip_supp,tab:ablation_noise_supp,tab:coco_supp,tab:results_of_jpeg_supp}, this time including LPIPS-VGG and CPSNR for completeness.
In \cref{fig:app_coco,fig:app_gb,fig:app_ip,fig:app_jpeg,fig:app_sr4,fig:app_sr8_noise,fig:app_sr8}, we provide additional  reconstructions examples of SILO, ReSample, PSLD, GML, and LDPS. 
These images are sampled with the settings used for \cref{tab:results_of_sr4_and_gb,tab:results_of_sr8_and_ip}.

\begingroup
\def\totalwidth{1}
\def\numdegredations{1}
\def\nummetrics{6}
\FPeval{\totalcolumns}{round(\numdegredations*\nummetrics,0)}
\FPeval{\colwidth}{\totalwidth/\totalcolumns}

\newcommand{\rankhl}[4]{%
  \ifnum\numexpr\pdfstrcmp{#1}{#2}=0
    \cellcolor{first}#1%
  \else\ifnum\numexpr\pdfstrcmp{#1}{#3}=0
    \cellcolor{second}#1%
  \else\ifnum\numexpr\pdfstrcmp{#1}{#4}=0
    \cellcolor{third}#1%
  \else
    #1%
  \fi\fi\fi
}

\newcommand{\rankhlrev}[4]{%
  \ifnum\numexpr\pdfstrcmp{#1}{#4}=0
    \cellcolor{first}#1%
  \else\ifnum\numexpr\pdfstrcmp{#1}{#3}=0
    \cellcolor{second}#1%
  \else\ifnum\numexpr\pdfstrcmp{#1}{#2}=0
    \cellcolor{third}#1%
  \else
    #1%
  \fi\fi\fi
}

\edef\firstTime{0}\edef\secondTime{0}\edef\thirdTime{0}
\edef\firstPSNRa{26.20}\edef\secondPSNRa{26.19}\edef\thirdPSNRa{25.14}
\edef\firstCPSNRa{35.52}\edef\secondCPSNRa{34.87}\edef\thirdCPSNRa{34.30}
\edef\firstLPIPSa{0.359}\edef\secondLPIPSa{0.279}\edef\thirdLPIPSa{0.252}
\edef\firstLPIPSVa{0.423}\edef\secondLPIPSVa{0.377}\edef\thirdLPIPSVa{0.357}
\edef\firstFIDa{39.61}\edef\secondFIDa{32.77}\edef\thirdFIDa{30.28}
\edef\firstKIDa{12.80}\edef\secondKIDa{8.00}\edef\thirdKIDa{6.04}

\begin{table}[hbt]
    \centering
    \setlength{\tabcolsep}{4pt}
    \small
    \begin{tabular}{l *{\totalcolumns}{c}}
        
        Method &
        PSNR &
        CPSNR &
        LPIPS-A &
        LPIPS-V &
        FID &
        KID \cr
        \hline
        
        Ours (RV)& 
        \rankhl{25.27}{\firstPSNRa}{\secondPSNRa}{\thirdPSNRa}&
        \rankhl{30.47}{\firstCPSNRa}{\secondCPSNRa}{\thirdCPSNRa}&
        \rankhlrev{0.252}{\firstLPIPSa}{\secondLPIPSa}{\thirdLPIPSa}&
        \rankhlrev{0.357}{\firstLPIPSVa}{\secondLPIPSVa}{\thirdLPIPSVa}&
        \rankhlrev{30.28}{\firstFIDa}{\secondFIDa}{\thirdFIDa}&
        \rankhlrev{6.04}{\firstKIDa}{\secondKIDa}{\thirdKIDa}
        \\

        Ours (SD)& 
        \rankhl{25.21}{\firstPSNRa}{\secondPSNRa}{\thirdPSNRa}&
        \rankhl{30.07}{\firstCPSNRa}{\secondCPSNRa}{\thirdCPSNRa}&
        \rankhlrev{0.279}{\firstLPIPSa}{\secondLPIPSa}{\thirdLPIPSa}&
        \rankhlrev{0.377}{\firstLPIPSVa}{\secondLPIPSVa}{\thirdLPIPSVa}&
        \rankhlrev{32.77}{\firstFIDa}{\secondFIDa}{\thirdFIDa}&
        \rankhlrev{8.00}{\firstKIDa}{\secondKIDa}{\thirdKIDa}
        \\

        ReSample& 
        \rankhl{16.18}{\firstPSNRa}{\secondPSNRa}{\thirdPSNRa}&
        \rankhl{31.22}{\firstCPSNRa}{\secondCPSNRa}{\thirdCPSNRa}&
        \rankhlrev{0.724}{\firstLPIPSa}{\secondLPIPSa}{\thirdLPIPSa}&
        \rankhlrev{0.740}{\firstLPIPSVa}{\secondLPIPSVa}{\thirdLPIPSVa}&
        \rankhlrev{235.8}{\firstFIDa}{\secondFIDa}{\thirdFIDa}&
        \rankhlrev{253.1}{\firstKIDa}{\secondKIDa}{\thirdKIDa}
        
        \\
        PSLD& 
        \rankhl{24.37}{\firstPSNRa}{\secondPSNRa}{\thirdPSNRa}&
        \rankhl{35.52}{\firstCPSNRa}{\secondCPSNRa}{\thirdCPSNRa}&
        \rankhlrev{0.359}{\firstLPIPSa}{\secondLPIPSa}{\thirdLPIPSa}&
        \rankhlrev{0.493}{\firstLPIPSVa}{\secondLPIPSVa}{\thirdLPIPSVa}&
        \rankhlrev{61.99}{\firstFIDa}{\secondFIDa}{\thirdFIDa}&
        \rankhlrev{33.23}{\firstKIDa}{\secondKIDa}{\thirdKIDa}
        \\

        GML-DPS& 
        \rankhl{26.19}{\firstPSNRa}{\secondPSNRa}{\thirdPSNRa}&
        \rankhl{34.30}{\firstCPSNRa}{\secondCPSNRa}{\thirdCPSNRa}&
        \rankhlrev{0.354}{\firstLPIPSa}{\secondLPIPSa}{\thirdLPIPSa}&
        \rankhlrev{0.425}{\firstLPIPSVa}{\secondLPIPSVa}{\thirdLPIPSVa}&
        \rankhlrev{41.06}{\firstFIDa}{\secondFIDa}{\thirdFIDa}&
        \rankhlrev{13.69}{\firstKIDa}{\secondKIDa}{\thirdKIDa}
        \\
        
        LDPS& 
        \rankhl{26.20}{\firstPSNRa}{\secondPSNRa}{\thirdPSNRa}&
        \rankhl{34.87}{\firstCPSNRa}{\secondCPSNRa}{\thirdCPSNRa}&
        \rankhlrev{0.359}{\firstLPIPSa}{\secondLPIPSa}{\thirdLPIPSa}&
        \rankhlrev{0.423}{\firstLPIPSVa}{\secondLPIPSVa}{\thirdLPIPSVa}&
        \rankhlrev{39.61}{\firstFIDa}{\secondFIDa}{\thirdFIDa}&
        \rankhlrev{12.80}{\firstKIDa}{\secondKIDa}{\thirdKIDa}
        \\
        
    \end{tabular}
    \caption{Comparison of inverse problem solvers using LDMs on the FFHQ dataset, for SR $\times8$ with $\sigma_y=0.03$. }
    \label{tab:ablation_noise_supp}
\end{table}
\endgroup

\begingroup
\def\totalwidth{1}
\def\numdegredations{1}
\def\nummetrics{6}
\FPeval{\totalcolumns}{round(\numdegredations*\nummetrics,0)}
\FPeval{\colwidth}{\totalwidth/\totalcolumns}

\newcommand{\rankhl}[4]{%
  \ifnum\numexpr\pdfstrcmp{#1}{#2}=0
    \cellcolor{first}#1%
  \else\ifnum\numexpr\pdfstrcmp{#1}{#3}=0
    \cellcolor{second}#1%
  \else\ifnum\numexpr\pdfstrcmp{#1}{#4}=0
    \cellcolor{third}#1%
  \else
    #1%
  \fi\fi\fi
}

\newcommand{\rankhlrev}[4]{%
  \ifnum\numexpr\pdfstrcmp{#1}{#4}=0
    \cellcolor{first}#1%
  \else\ifnum\numexpr\pdfstrcmp{#1}{#3}=0
    \cellcolor{second}#1%
  \else\ifnum\numexpr\pdfstrcmp{#1}{#2}=0
    \cellcolor{third}#1%
  \else
    #1%
  \fi\fi\fi
}

\edef\firstTime{0}\edef\secondTime{0}\edef\thirdTime{0}
\edef\firstPSNRa{18.51}\edef\secondPSNRa{18.30}\edef\thirdPSNRa{18.25}
\edef\firstCPSNRa{33.66}\edef\secondCPSNRa{27.92}\edef\thirdCPSNRa{27.27}
\edef\firstLPIPSa{0.297}\edef\secondLPIPSa{0.221}\edef\thirdLPIPSa{0.214}
\edef\firstLPIPSVa{0.368}\edef\secondLPIPSVa{0.302}\edef\thirdLPIPSVa{0.286}
\edef\firstFIDa{88.16}\edef\secondFIDa{48.96}\edef\thirdFIDa{45.59}
\edef\firstKIDa{21.99}\edef\secondKIDa{3.74}\edef\thirdKIDa{2.15}

\begin{table}[hbt]
    \centering
    \setlength{\tabcolsep}{4pt}
    \small
    \begin{tabular}{l *{\totalcolumns}{c}}
        
        Method &
        PSNR &
        CPSNR &
        LPIPS-A &
        LPIPS-V &
        FID &
        KID \cr
        \hline
        
        Ours (RV)& 
        \rankhl{18.51}{\firstPSNRa}{\secondPSNRa}{\thirdPSNRa}&
        \rankhl{27.92}{\firstCPSNRa}{\secondCPSNRa}{\thirdCPSNRa}&
        \rankhlrev{0.214}{\firstLPIPSa}{\secondLPIPSa}{\thirdLPIPSa}&
        \rankhlrev{0.286}{\firstLPIPSVa}{\secondLPIPSVa}{\thirdLPIPSVa}&
        \rankhlrev{48.96}{\firstFIDa}{\secondFIDa}{\thirdFIDa}&
        \rankhlrev{3.74}{\firstKIDa}{\secondKIDa}{\thirdKIDa}
        \\

        Ours (SD)& 
        \rankhl{18.30}{\firstPSNRa}{\secondPSNRa}{\thirdPSNRa}&
        \rankhl{27.12}{\firstCPSNRa}{\secondCPSNRa}{\thirdCPSNRa}&
        \rankhlrev{0.221}{\firstLPIPSa}{\secondLPIPSa}{\thirdLPIPSa}&
        \rankhlrev{0.302}{\firstLPIPSVa}{\secondLPIPSVa}{\thirdLPIPSVa}&
        \rankhlrev{45.59}{\firstFIDa}{\secondFIDa}{\thirdFIDa}&
        \rankhlrev{2.15}{\firstKIDa}{\secondKIDa}{\thirdKIDa}
        \\

        ReSample& 
        \rankhl{16.53}{\firstPSNRa}{\secondPSNRa}{\thirdPSNRa}&
        \rankhl{33.66}{\firstCPSNRa}{\secondCPSNRa}{\thirdCPSNRa}&
        \rankhlrev{0.297}{\firstLPIPSa}{\secondLPIPSa}{\thirdLPIPSa}&
        \rankhlrev{0.368}{\firstLPIPSVa}{\secondLPIPSVa}{\thirdLPIPSVa}&
        \rankhlrev{104.37}{\firstFIDa}{\secondFIDa}{\thirdFIDa}&
        \rankhlrev{54.16}{\firstKIDa}{\secondKIDa}{\thirdKIDa}
        
        \\
        PSLD& 
        \rankhl{18.24}{\firstPSNRa}{\secondPSNRa}{\thirdPSNRa}&
        \rankhl{27.27}{\firstCPSNRa}{\secondCPSNRa}{\thirdCPSNRa}&
        \rankhlrev{0.454}{\firstLPIPSa}{\secondLPIPSa}{\thirdLPIPSa}&
        \rankhlrev{0.513}{\firstLPIPSVa}{\secondLPIPSVa}{\thirdLPIPSVa}&
        \rankhlrev{90.38}{\firstFIDa}{\secondFIDa}{\thirdFIDa}&
        \rankhlrev{24.40}{\firstKIDa}{\secondKIDa}{\thirdKIDa}
        \\

        GML-DPS& 
        \rankhl{18.25}{\firstPSNRa}{\secondPSNRa}{\thirdPSNRa}&
        \rankhl{27.27}{\firstCPSNRa}{\secondCPSNRa}{\thirdCPSNRa}&
        \rankhlrev{0.453}{\firstLPIPSa}{\secondLPIPSa}{\thirdLPIPSa}&
        \rankhlrev{0.513}{\firstLPIPSVa}{\secondLPIPSVa}{\thirdLPIPSVa}&
        \rankhlrev{88.16}{\firstFIDa}{\secondFIDa}{\thirdFIDa}&
        \rankhlrev{21.99}{\firstKIDa}{\secondKIDa}{\thirdKIDa}
        \\
        
        LDPS& 
        \rankhl{18.26}{\firstPSNRa}{\secondPSNRa}{\thirdPSNRa}&
        \rankhl{26.94}{\firstCPSNRa}{\secondCPSNRa}{\thirdCPSNRa}&
        \rankhlrev{0.474}{\firstLPIPSa}{\secondLPIPSa}{\thirdLPIPSa}&
        \rankhlrev{0.513}{\firstLPIPSVa}{\secondLPIPSVa}{\thirdLPIPSVa}&
        \rankhlrev{92.67}{\firstFIDa}{\secondFIDa}{\thirdFIDa}&
        \rankhlrev{24.98}{\firstKIDa}{\secondKIDa}{\thirdKIDa}
        \\
        
    \end{tabular}
    \caption{Comparison of inverse problem solvers using latent diffusion on 1,000 images from the COCO dataset with inpainting.}
    \label{tab:coco_supp}
\end{table}
\endgroup

\begingroup
\def\totalwidth{1}
\def\numdegredations{2}
\def\nummetrics{6}
\FPeval{\totalcolumns}{round(\numdegredations*\nummetrics +1 ,0)}
\FPeval{\colwidth}{\totalwidth/\totalcolumns}

\newcommand{\rankhl}[4]{%
  \ifnum\numexpr\pdfstrcmp{#1}{#2}=0
    \cellcolor{first}#1%
  \else\ifnum\numexpr\pdfstrcmp{#1}{#3}=0
    \cellcolor{second}#1%
  \else\ifnum\numexpr\pdfstrcmp{#1}{#4}=0
    \cellcolor{third}#1%
  \else
    #1%
  \fi\fi\fi
}

\newcommand{\rankhlrev}[4]{%
  \ifnum\numexpr\pdfstrcmp{#1}{#4}=0
    \cellcolor{first}#1%
  \else\ifnum\numexpr\pdfstrcmp{#1}{#3}=0
    \cellcolor{second}#1%
  \else\ifnum\numexpr\pdfstrcmp{#1}{#2}=0
    \cellcolor{third}#1%
  \else
    #1%
  \fi\fi\fi
}

\edef\firstTime{148}\edef\secondTime{149}\edef\thirdTime{331}
\edef\firstPSNRa{27.01}\edef\secondPSNRa{26.89}\edef\thirdPSNRa{26.28}
\edef\firstCPSNRa{41.23}\edef\secondCPSNRa{40.91}\edef\thirdCPSNRa{38.73}
\edef\firstLPIPSa{0.320}\edef\secondLPIPSa{0.253}\edef\thirdLPIPSa{0.226}
\edef\firstLPIPSVa{0.399}\edef\secondLPIPSVa{0.344}\edef\thirdLPIPSVa{0.327}
\edef\firstFIDa{38.50}\edef\secondFIDa{30.71}\edef\thirdFIDa{27.10}
\edef\firstKIDa{12.94}\edef\secondKIDa{8.47}\edef\thirdKIDa{5.23}

\edef\firstPSNRb{22.51}\edef\secondPSNRb{22.23}\edef\thirdPSNRb{20.64}
\edef\firstCPSNRb{36.11}\edef\secondCPSNRb{30.09}\edef\thirdCPSNRb{30.04}
\edef\firstLPIPSb{0.273}\edef\secondLPIPSb{0.151}\edef\thirdLPIPSb{0.139}
\edef\firstLPIPSVb{0.359}\edef\secondLPIPSVb{0.258}\edef\thirdLPIPSVb{0.239}
\edef\firstFIDb{49.56}\edef\secondFIDb{21.04}\edef\thirdFIDb{18.98}
\edef\firstKIDb{16.02}\edef\secondKIDb{4.32}\edef\thirdKIDb{1.80}

\begin{table*}[bt]
    \centering
    \setlength{\tabcolsep}{4pt}
    \small
    \begin{tabular}{l *{\totalcolumns}{c}}
        \multicolumn{2}{c}{} &
	\multicolumn{\nummetrics}{c}{Super-Resolution ${\times}8$} &
	\multicolumn{\nummetrics}{c}{Inpainting}
	    \\
     
        \cmidrule(lr){3-8}
        \cmidrule(lr){9-14}
        
        Method &
        Time [sec]&
        PSNR &
        CPSNR &
        LPIPS-A &
        LPIPS-V &
        FID &
        KID &

        PSNR &
        CPSNR &
        LPIPS-A &
        LPIPS-V &
        FID &
        KID 
        
        \\

        \hline
        
        Ours (RV)& 
        \rankhl{149}{\firstTime}{\secondTime}{\thirdTime}&
        \rankhl{26.28}{\firstPSNRa}{\secondPSNRa}{\thirdPSNRa}&
        \rankhl{32.60}{\firstCPSNRa}{\secondCPSNRa}{\thirdCPSNRa}&
        \rankhlrev{0.226}{\firstLPIPSa}{\secondLPIPSa}{\thirdLPIPSa}&
        \rankhlrev{0.327}{\firstLPIPSVa}{\secondLPIPSVa}{\thirdLPIPSVa}&
        \rankhlrev{27.10}{\firstFIDa}{\secondFIDa}{\thirdFIDa}&
        \rankhlrev{5.23}{\firstKIDa}{\secondKIDa}{\thirdKIDa}&
        
        \rankhl{22.51}{\firstPSNRb}{\secondPSNRb}{\thirdPSNRb}&
        \rankhl{29.75}{\firstCPSNRb}{\secondCPSNRb}{\thirdCPSNRb}&
        \rankhlrev{0.139}{\firstLPIPSb}{\secondLPIPSb}{\thirdLPIPSb}&
        \rankhlrev{0.239}{\firstLPIPSVb}{\secondLPIPSVb}{\thirdLPIPSVb}&
        \rankhlrev{18.98}{\firstFIDb}{\secondFIDb}{\thirdFIDb}&
        \rankhlrev{1.80}{\firstKIDb}{\secondKIDb}{\thirdKIDb} 

        \\

        Ours (SD)& 
        \rankhl{148}{\firstTime}{\secondTime}{\thirdTime}&
        \rankhl{26.13}{\firstPSNRa}{\secondPSNRa}{\thirdPSNRa}&
        \rankhl{32.41}{\firstCPSNRa}{\secondCPSNRa}{\thirdCPSNRa}&
        \rankhlrev{0.253}{\firstLPIPSa}{\secondLPIPSa}{\thirdLPIPSa}&
        \rankhlrev{0.344}{\firstLPIPSVa}{\secondLPIPSVa}{\thirdLPIPSVa}&
        \rankhlrev{30.71}{\firstFIDa}{\secondFIDa}{\thirdFIDa}&
        \rankhlrev{8.47}{\firstKIDa}{\secondKIDa}{\thirdKIDa}&
        
        \rankhl{22.23}{\firstPSNRb}{\secondPSNRb}{\thirdPSNRb}&
        \rankhl{29.10}{\firstCPSNRb}{\secondCPSNRb}{\thirdCPSNRb}&
        \rankhlrev{0.151}{\firstLPIPSb}{\secondLPIPSb}{\thirdLPIPSb}&
        \rankhlrev{0.258}{\firstLPIPSVb}{\secondLPIPSVb}{\thirdLPIPSVb}&
        \rankhlrev{21.04}{\firstFIDb}{\secondFIDb}{\thirdFIDb}&
        \rankhlrev{4.32}{\firstKIDb}{\secondKIDb}{\thirdKIDb}

        \\

        ReSample& 
        \rankhl{1418}{\firstTime}{\secondTime}{\thirdTime}&
        \rankhl{22.80}{\firstPSNRa}{\secondPSNRa}{\thirdPSNRa}&
        \rankhl{41.23}{\firstCPSNRa}{\secondCPSNRa}{\thirdCPSNRa}&
        \rankhlrev{0.575}{\firstLPIPSa}{\secondLPIPSa}{\thirdLPIPSa}&
        \rankhlrev{0.603}{\firstLPIPSVa}{\secondLPIPSVa}{\thirdLPIPSVa}&
        \rankhlrev{131.75}{\firstFIDa}{\secondFIDa}{\thirdFIDa}&
        \rankhlrev{118.57}{\firstKIDa}{\secondKIDa}{\thirdKIDa}&
        
        \rankhl{16.91}{\firstPSNRb}{\secondPSNRb}{\thirdPSNRb}&
        \rankhl{36.11}{\firstCPSNRb}{\secondCPSNRb}{\thirdCPSNRb}&
        \rankhlrev{0.273}{\firstLPIPSb}{\secondLPIPSb}{\thirdLPIPSb}&
        \rankhlrev{0.359}{\firstLPIPSVb}{\secondLPIPSVb}{\thirdLPIPSVb}&
        \rankhlrev{146.08}{\firstFIDb}{\secondFIDb}{\thirdFIDb}&
        \rankhlrev{119.34}{\firstKIDb}{\secondKIDb}{\thirdKIDb}

        \\
        PSLD& 
        \rankhl{390}{\firstTime}{\secondTime}{\thirdTime}&
        \rankhl{25.08}{\firstPSNRa}{\secondPSNRa}{\thirdPSNRa}&
        \rankhl{40.91}{\firstCPSNRa}{\secondCPSNRa}{\thirdCPSNRa}&
        \rankhlrev{0.320}{\firstLPIPSa}{\secondLPIPSa}{\thirdLPIPSa}&
        \rankhlrev{0.419}{\firstLPIPSVa}{\secondLPIPSVa}{\thirdLPIPSVa}&
        \rankhlrev{41.58}{\firstFIDa}{\secondFIDa}{\thirdFIDa}&
        \rankhlrev{14.90}{\firstKIDa}{\secondKIDa}{\thirdKIDa}&
        
        \rankhl{20.58}{\firstPSNRb}{\secondPSNRb}{\thirdPSNRb}&
        \rankhl{30.04}{\firstCPSNRb}{\secondCPSNRb}{\thirdCPSNRb}&
        \rankhlrev{0.357}{\firstLPIPSb}{\secondLPIPSb}{\thirdLPIPSb}&
        \rankhlrev{0.445}{\firstLPIPSVb}{\secondLPIPSVb}{\thirdLPIPSVb}&
        \rankhlrev{50.84}{\firstFIDb}{\secondFIDb}{\thirdFIDb}&
        \rankhlrev{17.23}{\firstKIDb}{\secondKIDb}{\thirdKIDb}

        \\

        GML-DPS& 
        \rankhl{389}{\firstTime}{\secondTime}{\thirdTime}&
        \rankhl{27.01}{\firstPSNRa}{\secondPSNRa}{\thirdPSNRa}&
        \rankhl{38.04}{\firstCPSNRa}{\secondCPSNRa}{\thirdCPSNRa}&
        \rankhlrev{0.327}{\firstLPIPSa}{\secondLPIPSa}{\thirdLPIPSa}&
        \rankhlrev{0.399}{\firstLPIPSVa}{\secondLPIPSVa}{\thirdLPIPSVa}&
        \rankhlrev{38.71}{\firstFIDa}{\secondFIDa}{\thirdFIDa}&
        \rankhlrev{12.99}{\firstKIDa}{\secondKIDa}{\thirdKIDa}&
        
        \rankhl{20.64}{\firstPSNRb}{\secondPSNRb}{\thirdPSNRb}&
        \rankhl{30.09}{\firstCPSNRb}{\secondCPSNRb}{\thirdCPSNRb}&
        \rankhlrev{0.356}{\firstLPIPSb}{\secondLPIPSb}{\thirdLPIPSb}&
        \rankhlrev{0.443}{\firstLPIPSVb}{\secondLPIPSVb}{\thirdLPIPSVb}&
        \rankhlrev{49.89}{\firstFIDb}{\secondFIDb}{\thirdFIDb}&
        \rankhlrev{16.54}{\firstKIDb}{\secondKIDb}{\thirdKIDb}

        \\
        
        LDPS& 
        \rankhl{331}{\firstTime}{\secondTime}{\thirdTime}&
        \rankhl{26.89}{\firstPSNRa}{\secondPSNRa}{\thirdPSNRa}&
        \rankhl{38.73}{\firstCPSNRa}{\secondCPSNRa}{\thirdCPSNRa}&
        \rankhlrev{0.343}{\firstLPIPSa}{\secondLPIPSa}{\thirdLPIPSa}&
        \rankhlrev{0.404}{\firstLPIPSVa}{\secondLPIPSVa}{\thirdLPIPSVa}&
        \rankhlrev{38.50}{\firstFIDa}{\secondFIDa}{\thirdFIDa}&
        \rankhlrev{12.94}{\firstKIDa}{\secondKIDa}{\thirdKIDa}&
        
        \rankhl{20.58}{\firstPSNRb}{\secondPSNRb}{\thirdPSNRb}&
        \rankhl{29.81}{\firstCPSNRb}{\secondCPSNRb}{\thirdCPSNRb}&
        \rankhlrev{0.368}{\firstLPIPSb}{\secondLPIPSb}{\thirdLPIPSb}&
        \rankhlrev{0.440}{\firstLPIPSVb}{\secondLPIPSVb}{\thirdLPIPSVb}&
        \rankhlrev{49.56}{\firstFIDb}{\secondFIDb}{\thirdFIDb}&
        \rankhlrev{16.02}{\firstKIDb}{\secondKIDb}{\thirdKIDb}

    \end{tabular}
    \caption{Comparison of inverse problem solvers using latent diffusion on the FFHQ dataset.
    }
    \label{tab:results_of_sr8_and_ip_supp}
\end{table*}
\endgroup

\begingroup
\def\totalwidth{1}
\def\numdegredations{2}
\def\nummetrics{6}
\FPeval{\totalcolumns}{round(\numdegredations*\nummetrics +1,0)}
\FPeval{\colwidth}{\totalwidth/\totalcolumns}

\newcommand{\rankhl}[4]{%
  \ifnum\numexpr\pdfstrcmp{#1}{#2}=0
    \cellcolor{first}#1%
  \else\ifnum\numexpr\pdfstrcmp{#1}{#3}=0
    \cellcolor{second}#1%
  \else\ifnum\numexpr\pdfstrcmp{#1}{#4}=0
    \cellcolor{third}#1%
  \else
    #1%
  \fi\fi\fi
}

\newcommand{\rankhlrev}[4]{%
  \ifnum\numexpr\pdfstrcmp{#1}{#4}=0
    \cellcolor{first}#1%
  \else\ifnum\numexpr\pdfstrcmp{#1}{#3}=0
    \cellcolor{second}#1%
  \else\ifnum\numexpr\pdfstrcmp{#1}{#2}=0
    \cellcolor{third}#1%
  \else
    #1%
  \fi\fi\fi
}

\edef\firstTime{148}\edef\secondTime{149}\edef\thirdTime{331}
\edef\firstPSNRa{28.74}\edef\secondPSNRa{28.63}\edef\thirdPSNRa{28.00}
\edef\firstCPSNRa{50.66}\edef\secondCPSNRa{44.18}\edef\thirdCPSNRa{44.07}
\edef\firstLPIPSa{0.253}\edef\secondLPIPSa{0.236}\edef\thirdLPIPSa{0.222}
\edef\firstLPIPSVa{0.359}\edef\secondLPIPSVa{0.327}\edef\thirdLPIPSVa{0.311}
\edef\firstFIDa{30.33}\edef\secondFIDa{29.61}\edef\thirdFIDa{28.34}
\edef\firstKIDa{10.74}\edef\secondKIDa{9.68}\edef\thirdKIDa{8.21}

\edef\firstPSNRb{29.34}\edef\secondPSNRb{29.06}\edef\thirdPSNRb{28.23}
\edef\firstCPSNRb{42.74}\edef\secondCPSNRb{38.76}\edef\thirdCPSNRb{36.69}
\edef\firstLPIPSb{0.247}\edef\secondLPIPSb{0.200}\edef\thirdLPIPSb{0.182}
\edef\firstLPIPSVb{0.335}\edef\secondLPIPSVb{0.306}\edef\thirdLPIPSVb{0.291}
\edef\firstFIDb{29.63}\edef\secondFIDb{26.51}\edef\thirdFIDb{23.82}
\edef\firstKIDb{9.05}\edef\secondKIDb{7.34}\edef\thirdKIDb{5.10}

\begin{table*}[hbt]
    \centering
    \setlength{\tabcolsep}{4pt}
    \small
    \begin{tabular}{l *{\totalcolumns}{c}}
        \multicolumn{2}{c}{} &
	    \multicolumn{\nummetrics}{c}{Gaussian blur} &
	    \multicolumn{\nummetrics}{c}{Super-Resolution $\times4$} 
	    \\
        
        \cmidrule(lr){3-8}
        \cmidrule(lr){9-14}
        
        Method &
        Time [sec]&
        PSNR &
        CPSNR &
        LPIPS-A &
        LPIPS-V &
        FID &
        KID &
        PSNR &
        CPSNR &
        LPIPS-A &
        LPIPS-V &
        FID&
        KID 
        \\

        \hline
        
        Ours (RV)& 
        \rankhl{149}{\firstTime}{\secondTime}{\thirdTime}&
        \rankhl{26.70}{\firstPSNRa}{\secondPSNRa}{\thirdPSNRa}&
        \rankhl{32.17}{\firstCPSNRa}{\secondCPSNRa}{\thirdCPSNRa}&
        
        \rankhlrev{0.222}{\firstLPIPSa}{\secondLPIPSa}{\thirdLPIPSa}&
        \rankhlrev{0.311}{\firstLPIPSVa}{\secondLPIPSVa}{\thirdLPIPSVa}&
        
        \rankhlrev{28.34}{\firstFIDa}{\secondFIDa}{\thirdFIDa}&
        \rankhlrev{8.21}{\firstKIDa}{\secondKIDa}{\thirdKIDa}&
        
        \rankhl{27.03}{\firstPSNRb}{\secondPSNRb}{\thirdPSNRb}&
        \rankhl{30.49}{\firstCPSNRb}{\secondCPSNRb}{\thirdCPSNRb}&
        \rankhlrev{0.182}{\firstLPIPSb}{\secondLPIPSb}{\thirdLPIPSb}&
        \rankhlrev{0.291}{\firstLPIPSVb}{\secondLPIPSVb}{\thirdLPIPSVb}&
        \rankhlrev{23.82}{\firstFIDb}{\secondFIDb}{\thirdFIDb}&
        \rankhlrev{5.10}{\firstKIDb}{\secondKIDb}{\thirdKIDb}
        \\

        Ours (SD)& 
        \rankhl{148}{\firstTime}{\secondTime}{\thirdTime}&
        \rankhl{26.55}{\firstPSNRa}{\secondPSNRa}{\thirdPSNRa}&
        \rankhl{31.35}{\firstCPSNRa}{\secondCPSNRa}{\thirdCPSNRa}&
        \rankhlrev{0.236}{\firstLPIPSa}{\secondLPIPSa}{\thirdLPIPSa}&
        \rankhlrev{0.327}{\firstLPIPSVa}{\secondLPIPSVa}{\thirdLPIPSVa}&
        \rankhlrev{30.33}{\firstFIDa}{\secondFIDa}{\thirdFIDa}&
        \rankhlrev{9.68}{\firstKIDa}{\secondKIDa}{\thirdKIDa}&
        
        \rankhl{26.95}{\firstPSNRb}{\secondPSNRb}{\thirdPSNRb}&
        \rankhl{30.23}{\firstCPSNRb}{\secondCPSNRb}{\thirdCPSNRb}&
        \rankhlrev{0.200}{\firstLPIPSb}{\secondLPIPSb}{\thirdLPIPSb}&
        \rankhlrev{0.306}{\firstLPIPSVb}{\secondLPIPSVb}{\thirdLPIPSVb}&
        \rankhlrev{26.51}{\firstFIDb}{\secondFIDb}{\thirdFIDb}&
        \rankhlrev{7.34}{\firstKIDb}{\secondKIDb}{\thirdKIDb}
        \\

        ReSample& 
        \rankhl{1418}{\firstTime}{\secondTime}{\thirdTime}&
        \rankhl{27.92}{\firstPSNRa}{\secondPSNRa}{\thirdPSNRa}&
        \rankhl{50.66}{\firstCPSNRa}{\secondCPSNRa}{\thirdCPSNRa}&
        \rankhlrev{0.253}{\firstLPIPSa}{\secondLPIPSa}{\thirdLPIPSa}&
        \rankhlrev{0.411}{\firstLPIPSVa}{\secondLPIPSVa}{\thirdLPIPSVa}&
        \rankhlrev{29.61}{\firstFIDa}{\secondFIDa}{\thirdFIDa}&
        \rankhlrev{10.74}{\firstKIDa}{\secondKIDa}{\thirdKIDa}&
        
        \rankhl{24.62}{\firstPSNRb}{\secondPSNRb}{\thirdPSNRb}&
        \rankhl{42.74}{\firstCPSNRb}{\secondCPSNRb}{\thirdCPSNRb}&
        \rankhlrev{0.433}{\firstLPIPSb}{\secondLPIPSb}{\thirdLPIPSb}&
        \rankhlrev{0.504}{\firstLPIPSVb}{\secondLPIPSVb}{\thirdLPIPSVb}&
        \rankhlrev{45.02}{\firstFIDb}{\secondFIDb}{\thirdFIDb}&
        \rankhlrev{25.50}{\firstKIDb}{\secondKIDb}{\thirdKIDb}
        \\
        PSLD& 
        \rankhl{390}{\firstTime}{\secondTime}{\thirdTime}&
        \rankhl{28.63}{\firstPSNRa}{\secondPSNRa}{\thirdPSNRa}&
        \rankhl{44.18}{\firstCPSNRa}{\secondCPSNRa}{\thirdCPSNRa}&
        \rankhlrev{0.288}{\firstLPIPSa}{\secondLPIPSa}{\thirdLPIPSa}&
        \rankhlrev{0.372}{\firstLPIPSVa}{\secondLPIPSVa}{\thirdLPIPSVa}&
        \rankhlrev{38.44}{\firstFIDa}{\secondFIDa}{\thirdFIDa}&
        \rankhlrev{12.23}{\firstKIDa}{\secondKIDa}{\thirdKIDa}&
        
        \rankhl{28.23}{\firstPSNRb}{\secondPSNRb}{\thirdPSNRb}&
        \rankhl{38.76}{\firstCPSNRb}{\secondCPSNRb}{\thirdCPSNRb}&
        \rankhlrev{0.249}{\firstLPIPSb}{\secondLPIPSb}{\thirdLPIPSb}&
        \rankhlrev{0.355}{\firstLPIPSVb}{\secondLPIPSVb}{\thirdLPIPSVb}&
        \rankhlrev{29.63}{\firstFIDb}{\secondFIDb}{\thirdFIDb}&
        \rankhlrev{10.11}{\firstKIDb}{\secondKIDb}{\thirdKIDb}
        \\

        GML-DPS& 
        \rankhl{389}{\firstTime}{\secondTime}{\thirdTime}&
        \rankhl{28.74}{\firstPSNRa}{\secondPSNRa}{\thirdPSNRa}&
        \rankhl{44.07}{\firstCPSNRa}{\secondCPSNRa}{\thirdCPSNRa}&
        \rankhlrev{0.309}{\firstLPIPSa}{\secondLPIPSa}{\thirdLPIPSa}&
        \rankhlrev{0.359}{\firstLPIPSVa}{\secondLPIPSVa}{\thirdLPIPSVa}&
        \rankhlrev{42.68}{\firstFIDa}{\secondFIDa}{\thirdFIDa}&
        \rankhlrev{16.58}{\firstKIDa}{\secondKIDa}{\thirdKIDa}&
        
        \rankhl{29.34}{\firstPSNRb}{\secondPSNRb}{\thirdPSNRb}&
        \rankhl{36.69}{\firstCPSNRb}{\secondCPSNRb}{\thirdCPSNRb}&
        \rankhlrev{0.247}{\firstLPIPSb}{\secondLPIPSb}{\thirdLPIPSb}&
        \rankhlrev{0.335}{\firstLPIPSVb}{\secondLPIPSVb}{\thirdLPIPSVb}&
        \rankhlrev{30.71}{\firstFIDb}{\secondFIDb}{\thirdFIDb}&
        \rankhlrev{9.05}{\firstKIDb}{\secondKIDb}{\thirdKIDb}
        \\
        
        LDPS& 
        \rankhl{331}{\firstTime}{\secondTime}{\thirdTime}&
        \rankhl{28.00}{\firstPSNRa}{\secondPSNRa}{\thirdPSNRa}&
        \rankhl{42.69}{\firstCPSNRa}{\secondCPSNRa}{\thirdCPSNRa}&
        \rankhlrev{0.327}{\firstLPIPSa}{\secondLPIPSa}{\thirdLPIPSa}&
        \rankhlrev{0.378}{\firstLPIPSVa}{\secondLPIPSVa}{\thirdLPIPSVa}&
        \rankhlrev{47.38}{\firstFIDa}{\secondFIDa}{\thirdFIDa}&
        \rankhlrev{19.95}{\firstKIDa}{\secondKIDa}{\thirdKIDa}&
        
        \rankhl{29.06}{\firstPSNRb}{\secondPSNRb}{\thirdPSNRb}&
        \rankhl{36.39}{\firstCPSNRb}{\secondCPSNRb}{\thirdCPSNRb}&
        \rankhlrev{0.281}{\firstLPIPSb}{\secondLPIPSb}{\thirdLPIPSb}&
        \rankhlrev{0.362}{\firstLPIPSVb}{\secondLPIPSVb}{\thirdLPIPSVb}&
        \rankhlrev{34.44}{\firstFIDb}{\secondFIDb}{\thirdFIDb}&
        \rankhlrev{11.69}{\firstKIDb}{\secondKIDb}{\thirdKIDb}
        \\

    \end{tabular}
    \caption{Comparison of inverse problem solvers using latent diffusion on the FFHQ dataset.
    }
    \label{tab:results_of_sr4_and_gb_supp}
\end{table*}
\endgroup

\begingroup
\def\totalwidth{1}
\def\numdegredations{1}
\def\nummetrics{6}
\FPeval{\totalcolumns}{round(\numdegredations*\nummetrics +1 ,0)}
\FPeval{\colwidth}{\totalwidth/\totalcolumns}

\newcommand{\rankhl}[4]{%
  \ifnum\numexpr\pdfstrcmp{#1}{#2}=0
    \cellcolor{first}#1%
  \else\ifnum\numexpr\pdfstrcmp{#1}{#3}=0
    \cellcolor{second}#1%
  \else\ifnum\numexpr\pdfstrcmp{#1}{#4}=0
    \cellcolor{third}#1%
  \else
    #1%
  \fi\fi\fi
}

\newcommand{\rankhlrev}[4]{%
  \ifnum\numexpr\pdfstrcmp{#1}{#4}=0
    \cellcolor{first}#1%
  \else\ifnum\numexpr\pdfstrcmp{#1}{#3}=0
    \cellcolor{second}#1%
  \else\ifnum\numexpr\pdfstrcmp{#1}{#2}=0
    \cellcolor{third}#1%
  \else
    #1%
  \fi\fi\fi
}

\edef\firstTime{133}\edef\secondTime{138}\edef\thirdTime{402}

\edef\firstPSNRc{27.60}\edef\secondPSNRc{25.69}\edef\thirdPSNRc{25.52}
\edef\firstCPSNRc{38.74}\edef\secondCPSNRc{29.53}\edef\thirdCPSNRc{25.82}
\edef\firstLPIPSc{0.268}\edef\secondLPIPSc{0.212}\edef\thirdLPIPSc{0.203}
\edef\firstLPIPSVc{0.373}\edef\secondLPIPSVc{0.341}\edef\thirdLPIPSVc{0.326}
\edef\firstFIDc{33.69}\edef\secondFIDc{27.30}\edef\thirdFIDc{25.48}
\edef\firstKIDc{7.54}\edef\secondKIDc{6.02}\edef\thirdKIDc{4.21}

\begin{table*}[tb]
    \centering
    \setlength{\tabcolsep}{4pt}
    \small
    \begin{tabular}{l *{\totalcolumns}{c}}
        \multicolumn{2}{c}{} &
	\multicolumn{\nummetrics}{c}{JPEG} 
	    \\
     
        \cmidrule(lr){3-8}
        
        Method &
        Time [sec]&
        PSNR &
        CPSNR &
        LPIPS-A &
        LPIPS-V &
        FID &
        KID 

        \\

        \hline
        
        Ours (RV)& 
        
        \rankhl{138}{\firstTime}{\secondTime}{\thirdTime}&
        \rankhl{25.52}{\firstPSNRc}{\secondPSNRc}{\thirdPSNRc}&
        \rankhl{25.73}{\firstCPSNRc}{\secondCPSNRc}{\thirdCPSNRc}&
        \rankhlrev{0.203}{\firstLPIPSc}{\secondLPIPSc}{\thirdLPIPSc}&
        \rankhlrev{0.326}{\firstLPIPSVc}{\secondLPIPSVc}{\thirdLPIPSVc}&
        \rankhlrev{25.48}{\firstFIDc}{\secondFIDc}{\thirdFIDc}&
        \rankhlrev{4.21}{\firstKIDc}{\secondKIDc}{\thirdKIDc}
        \\

        Ours (SD)& 
        \rankhl{133}{\firstTime}{\secondTime}{\thirdTime}&
        \rankhl{25.40}{\firstPSNRc}{\secondPSNRc}{\thirdPSNRc}&
        \rankhl{25.60}{\firstCPSNRc}{\secondCPSNRc}{\thirdCPSNRc}&
        \rankhlrev{0.212}{\firstLPIPSc}{\secondLPIPSc}{\thirdLPIPSc}&
        \rankhlrev{0.341}{\firstLPIPSVc}{\secondLPIPSVc}{\thirdLPIPSVc}&
        \rankhlrev{27.30}{\firstFIDc}{\secondFIDc}{\thirdFIDc}&
        \rankhlrev{6.02}{\firstKIDc}{\secondKIDc}{\thirdKIDc}
        \\

        ReSample& 
        \rankhl{2438}{\firstTime}{\secondTime}{\thirdTime}&
        \rankhl{25.69}{\firstPSNRc}{\secondPSNRc}{\thirdPSNRc}&
        \rankhl{38.74}{\firstCPSNRc}{\secondCPSNRc}{\thirdCPSNRc}&
        \rankhlrev{0.456}{\firstLPIPSc}{\secondLPIPSc}{\thirdLPIPSc}&
        \rankhlrev{0.493}{\firstLPIPSVc}{\secondLPIPSVc}{\thirdLPIPSVc}&
        \rankhlrev{39.71}{\firstFIDc}{\secondFIDc}{\thirdFIDc}&
        \rankhlrev{20.17}{\firstKIDc}{\secondKIDc}{\thirdKIDc}
        \\
        PSLD& 
        \multicolumn{7}{c}{\cellcolor{invalid}cannot compute for nonlinear $\mathcal{A}$}

        \\

        GML-DPS& 
        \rankhl{402}{\firstTime}{\secondTime}{\thirdTime}&

        \rankhl{27.60}{\firstPSNRc}{\secondPSNRc}{\thirdPSNRc}&
        \rankhl{29.53}{\firstCPSNRc}{\secondCPSNRc}{\thirdCPSNRc}&
        \rankhlrev{0.268}{\firstLPIPSc}{\secondLPIPSc}{\thirdLPIPSc}&
        \rankhlrev{0.373}{\firstLPIPSVc}{\secondLPIPSVc}{\thirdLPIPSVc}&
        \rankhlrev{33.69}{\firstFIDc}{\secondFIDc}{\thirdFIDc}&
        \rankhlrev{7.54}{\firstKIDc}{\secondKIDc}{\thirdKIDc}
        \\
        
        LDPS& 
        \rankhl{412}{\firstTime}{\secondTime}{\thirdTime}&

        \rankhl{24.53}{\firstPSNRc}{\secondPSNRc}{\thirdPSNRc}&
        \rankhl{25.82}{\firstCPSNRc}{\secondCPSNRc}{\thirdCPSNRc}&
        \rankhlrev{0.373}{\firstLPIPSc}{\secondLPIPSc}{\thirdLPIPSc}&
        \rankhlrev{0.445}{\firstLPIPSVc}{\secondLPIPSVc}{\thirdLPIPSVc}&
        \rankhlrev{53.21}{\firstFIDc}{\secondFIDc}{\thirdFIDc}&
        \rankhlrev{17.71}{\firstKIDc}{\secondKIDc}{\thirdKIDc}

    \end{tabular}
    \caption{Comparison of inverse problem solvers using latent diffusion on the FFHQ dataset.
    In this table, the time values are for the JPEG decompression task.
    }
    \label{tab:results_of_jpeg_supp}
\end{table*}
\endgroup

\begingroup
\newcolumntype{M}[1]{>{\centering\arraybackslash}m{#1}}
\newcommand{\vcentered}[1]{\begin{tabular}{@{}l@{}} #1 \end{tabular}}
\setlength{\tabcolsep}{0pt} %
\renewcommand{\arraystretch}{0} %

\def\columns{8}
\def\totalwidth{1}

\FPeval{\colwidth}{\totalwidth/\columns}
\FPeval{\imgwidth}{\totalwidth/\columns *\columns}

\FPeval{\colwidth}{clip(\totalwidth/\columns)}

\begin{figure*}[h!]
    \centering
    \begin{tabular}{*{\columns}{M{\colwidth\linewidth}}}

        \footnotesize{$x$} &
        \footnotesize{$y$} &
        \footnotesize{Ours (RV)} &
        \footnotesize{Ours (SD)} &
        \footnotesize{ReSample} &
        \footnotesize{PSLD} &
        \footnotesize{GML} &
        \footnotesize{LDPS} 
        \\
	    
        \rule{0pt}{0.8ex}\\

        \multicolumn{\columns}{c}{\centered{\includegraphics[width=\imgwidth\linewidth]{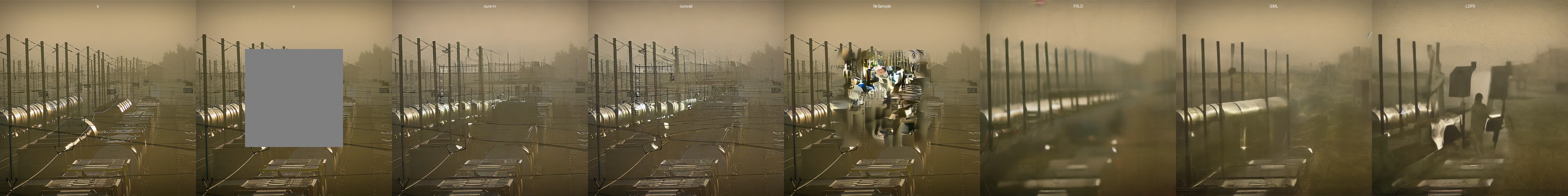}}} 
        \\
        \multicolumn{\columns}{c}{\centered{\includegraphics[width=\imgwidth\linewidth]{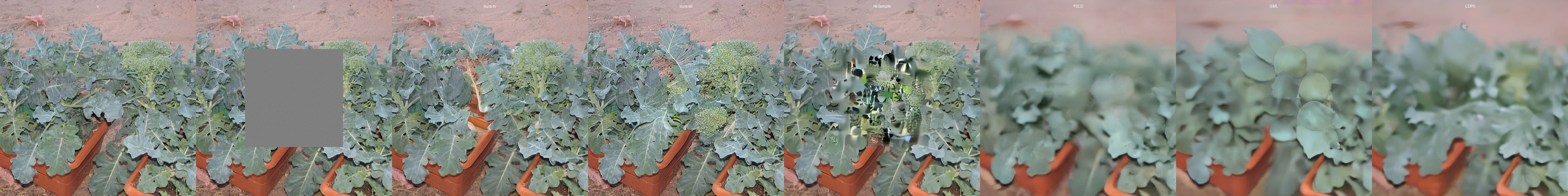}}} 
        \\
        \multicolumn{\columns}{c}{\centered{\includegraphics[width=\imgwidth \linewidth]{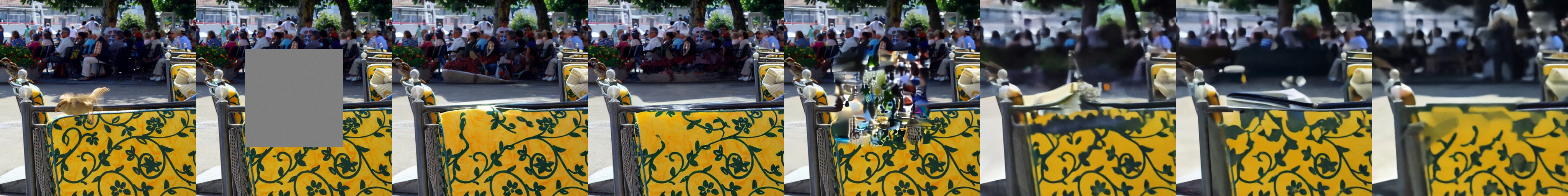}}} 
        \\
        \multicolumn{\columns}{c}{\centered{\includegraphics[width=\imgwidth \linewidth]{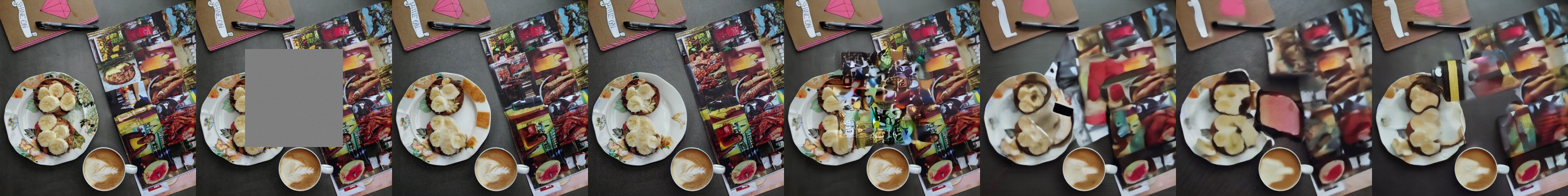}}} 

        \\
        \multicolumn{\columns}{c}{\centered{\includegraphics[width=\imgwidth \linewidth]{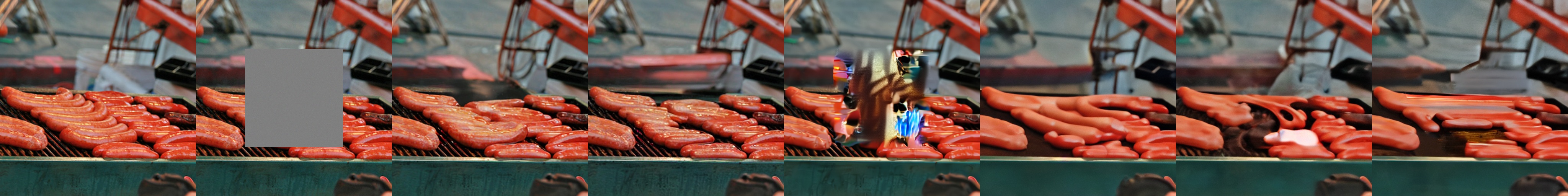}}} 
        \\
        \multicolumn{\columns}{c}{\centered{\includegraphics[width=\imgwidth \linewidth]{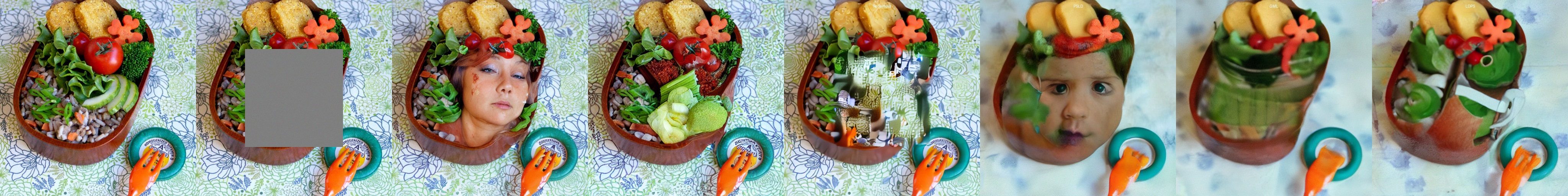}}}
        \\
        \multicolumn{\columns}{c}{\centered{\includegraphics[width=\imgwidth \linewidth]{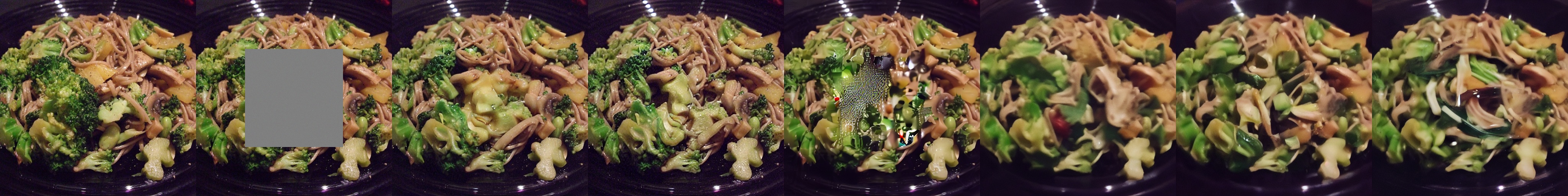}}}
        \\
        \multicolumn{\columns}{c}{\centered{\includegraphics[width=\imgwidth \linewidth]{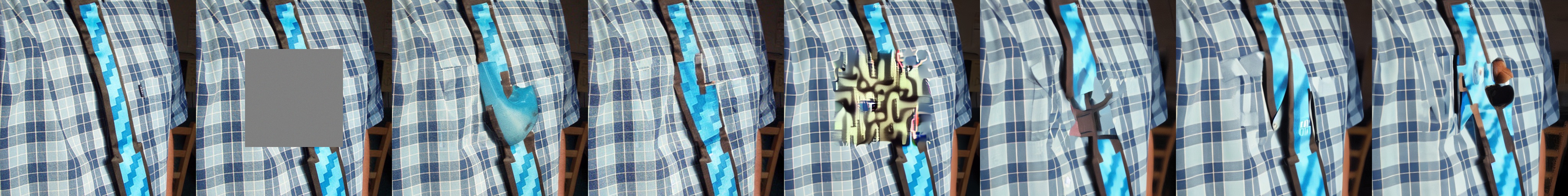}}}
        \\
        \multicolumn{\columns}{c}{\centered{\includegraphics[width=\imgwidth \linewidth]{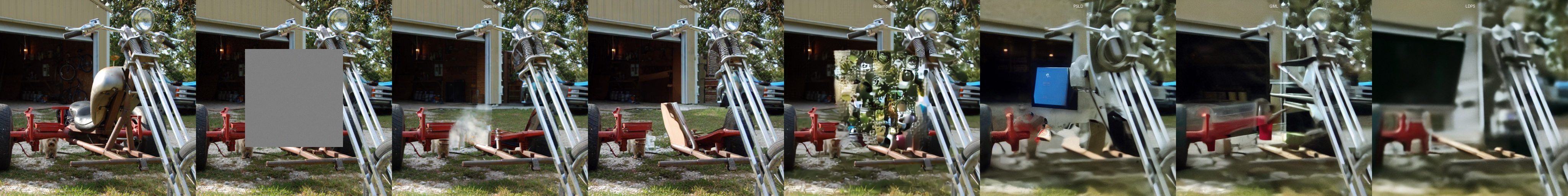}}}

    \end{tabular}
    \caption{Box inpainting with $\sigma_y=0.01$, COCO dataset. Additional results.}
    \label{fig:app_coco}
\end{figure*}
\endgroup

\begingroup
\newcolumntype{M}[1]{>{\centering\arraybackslash}m{#1}}
\newcommand{\vcentered}[1]{\begin{tabular}{@{}l@{}} #1 \end{tabular}}
\setlength{\tabcolsep}{0pt} %
\renewcommand{\arraystretch}{0} %

\def\columns{8}
\def\totalwidth{1}

\FPeval{\colwidth}{\totalwidth/\columns}
\FPeval{\imgwidth}{\totalwidth/\columns *\columns}

\FPeval{\colwidth}{clip(\totalwidth/\columns)}

\begin{figure*}[h!]
    \centering
    \begin{tabular}{*{\columns}{M{\colwidth\linewidth}}}

        \footnotesize{$x$} &
        \footnotesize{$y$} &
        \footnotesize{Ours (RV)} &
        \footnotesize{Ours (SD)} &
        \footnotesize{ReSample} &
        \footnotesize{PSLD} &
        \footnotesize{GML} &
        \footnotesize{LDPS} 
        \\
	    
        \rule{0pt}{0.8ex}\\

        \multicolumn{\columns}{c}{\centered{\includegraphics[width=\imgwidth\linewidth]{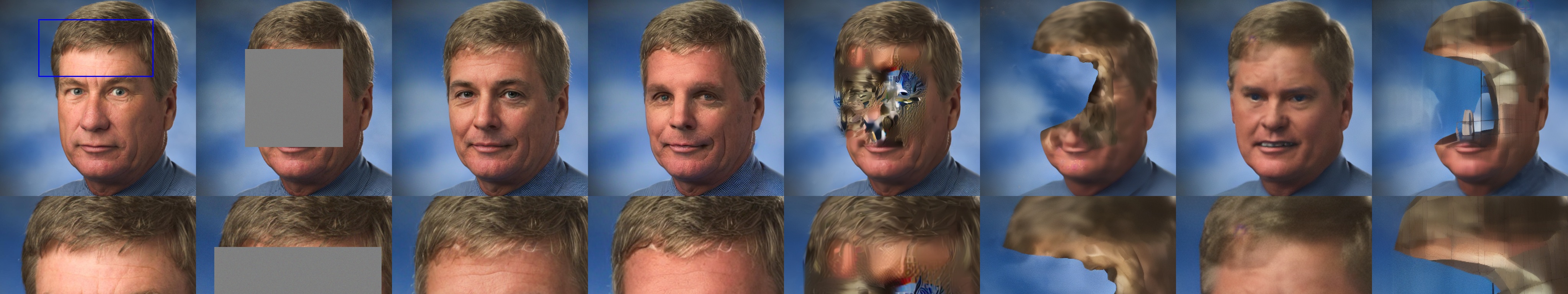}}} 
        \\
        \multicolumn{\columns}{c}{\centered{\includegraphics[width=\imgwidth\linewidth]{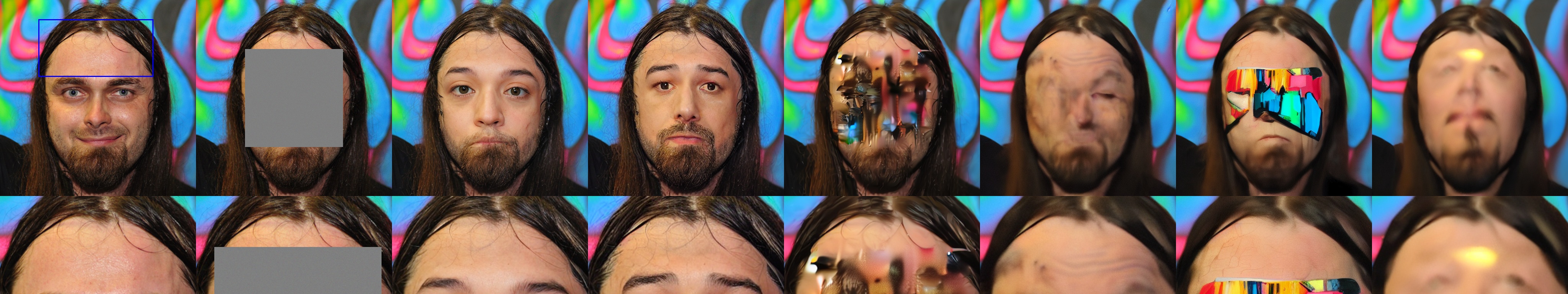}}} 
        \\
        \multicolumn{\columns}{c}{\centered{\includegraphics[width=\imgwidth \linewidth]{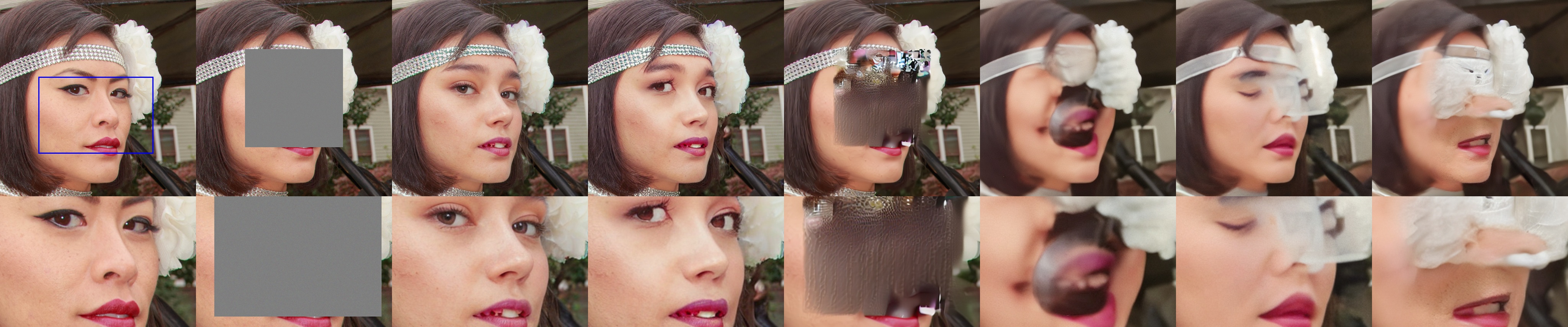}}} 
        \\
        \multicolumn{\columns}{c}{\centered{\includegraphics[width=\imgwidth \linewidth]{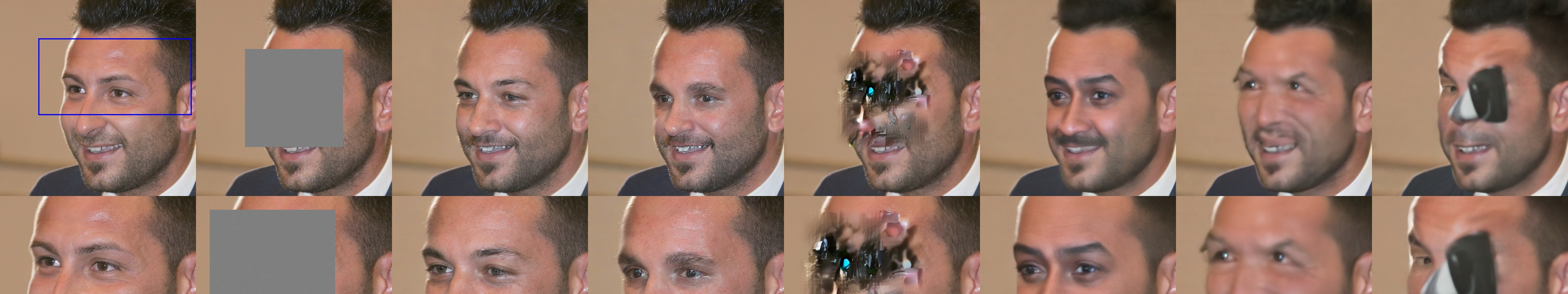}}} 

        \\
        \multicolumn{\columns}{c}{\centered{\includegraphics[width=\imgwidth \linewidth]{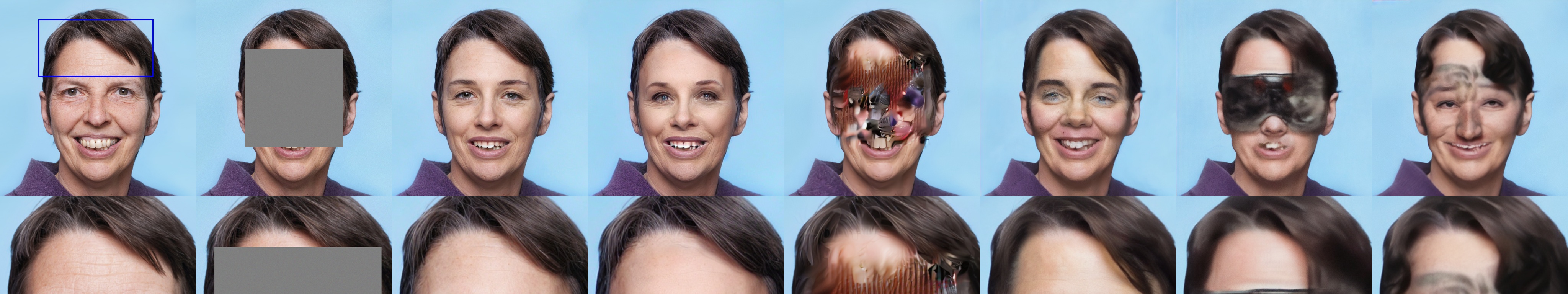}}} 
        \\
        \multicolumn{\columns}{c}{\centered{\includegraphics[width=\imgwidth \linewidth]{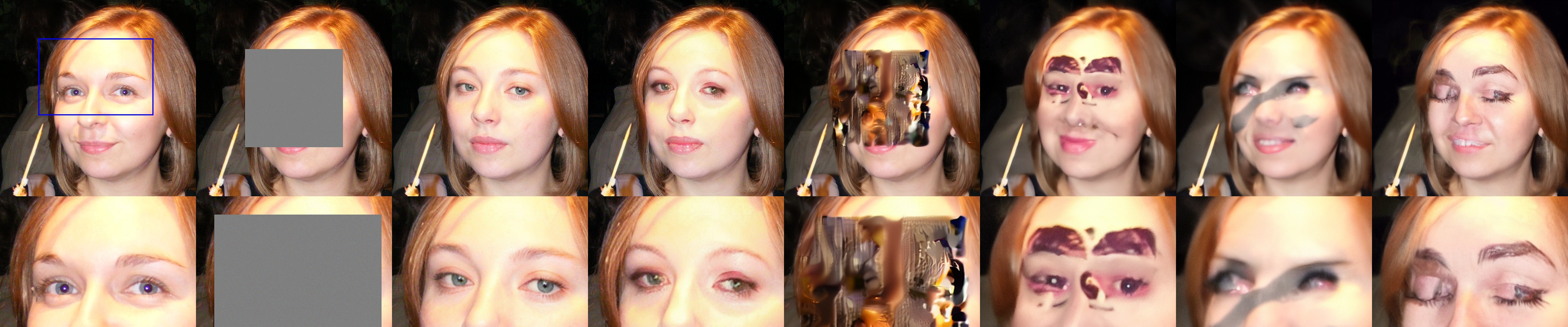}}}

    \end{tabular}
    \caption{Box inpainting with $\sigma_y=0.01$, FFHQ dataset. Additional results.}
    \label{fig:app_ip}
\end{figure*}
\endgroup

\begingroup
\newcolumntype{M}[1]{>{\centering\arraybackslash}m{#1}}
\newcommand{\vcentered}[1]{\begin{tabular}{@{}l@{}} #1 \end{tabular}}
\setlength{\tabcolsep}{0pt} %
\renewcommand{\arraystretch}{0} %

\def\columns{8}
\def\totalwidth{1}

\FPeval{\colwidth}{\totalwidth/\columns}
\FPeval{\imgwidth}{\totalwidth/\columns *\columns}

\FPeval{\colwidth}{clip(\totalwidth/\columns)}

\begin{figure*}[h!]
    \centering
    \begin{tabular}{*{\columns}{M{\colwidth\linewidth}}}

        \footnotesize{$x$} &
        \footnotesize{$y$} &
        \footnotesize{Ours (RV)} &
        \footnotesize{Ours (SD)} &
        \footnotesize{ReSample} &
        \footnotesize{PSLD} &
        \footnotesize{GML} &
        \footnotesize{LDPS} 
        \\
	    
        \rule{0pt}{0.8ex}\\

        \multicolumn{\columns}{c}{\centered{\includegraphics[width=\imgwidth\linewidth]{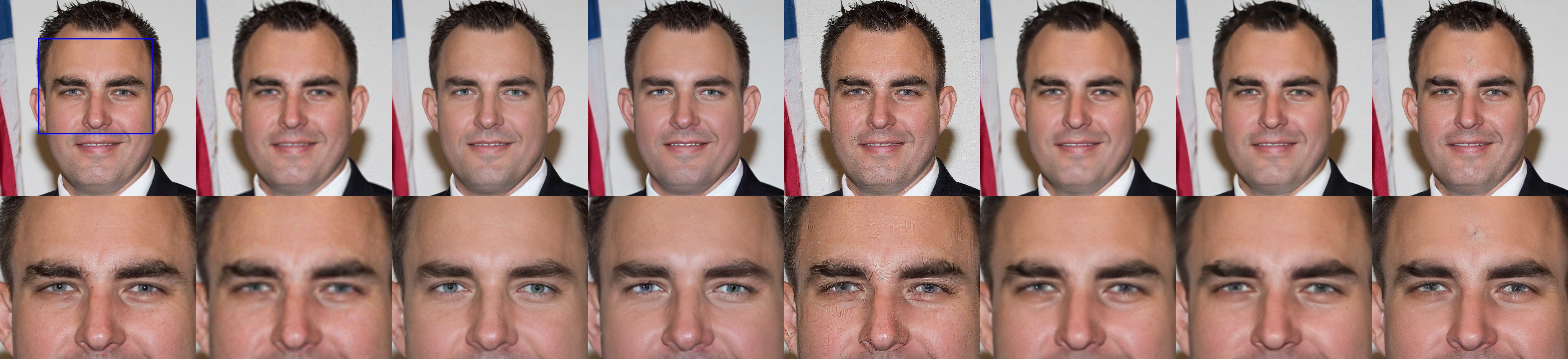}}} 
        \\
        \multicolumn{\columns}{c}{\centered{\includegraphics[width=\imgwidth\linewidth]{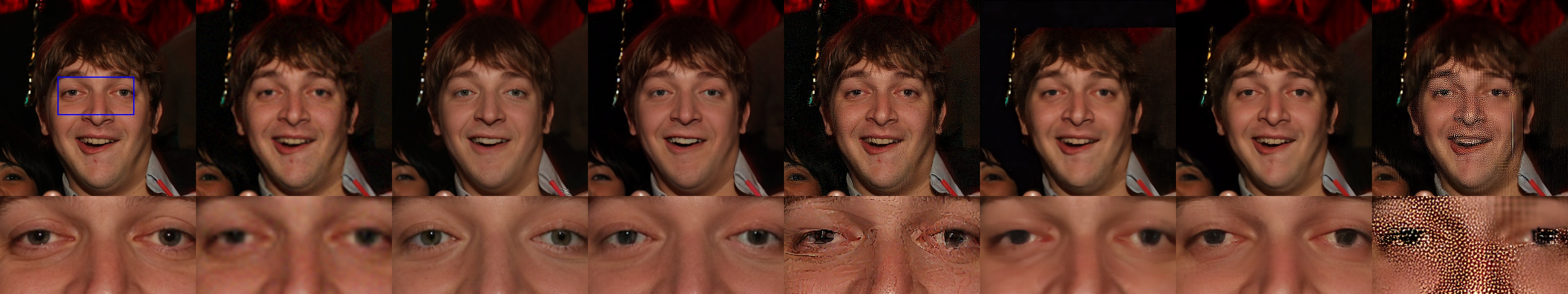}}} 
        \\
        \multicolumn{\columns}{c}{\centered{\includegraphics[width=\imgwidth \linewidth]{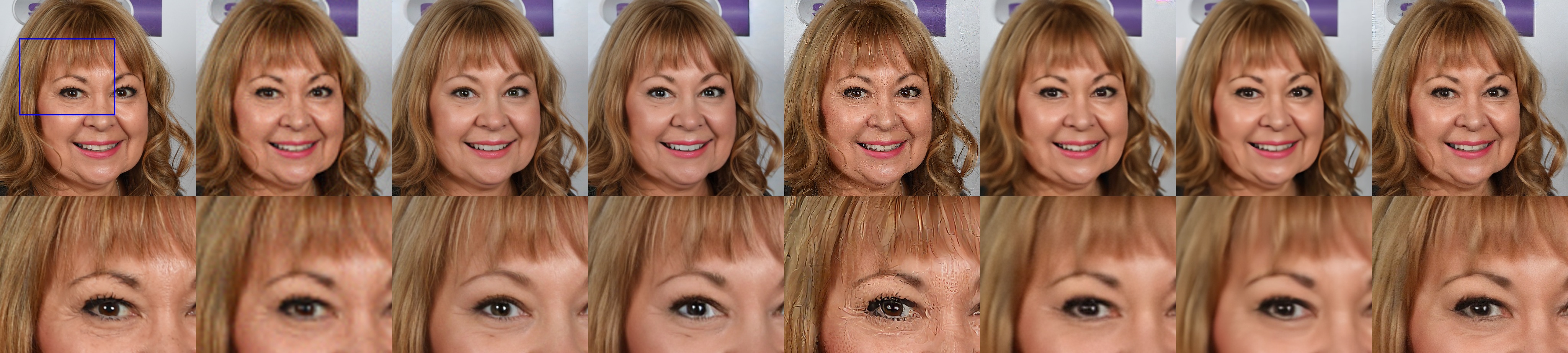}}} 
        \\
        \multicolumn{\columns}{c}{\centered{\includegraphics[width=\imgwidth \linewidth]{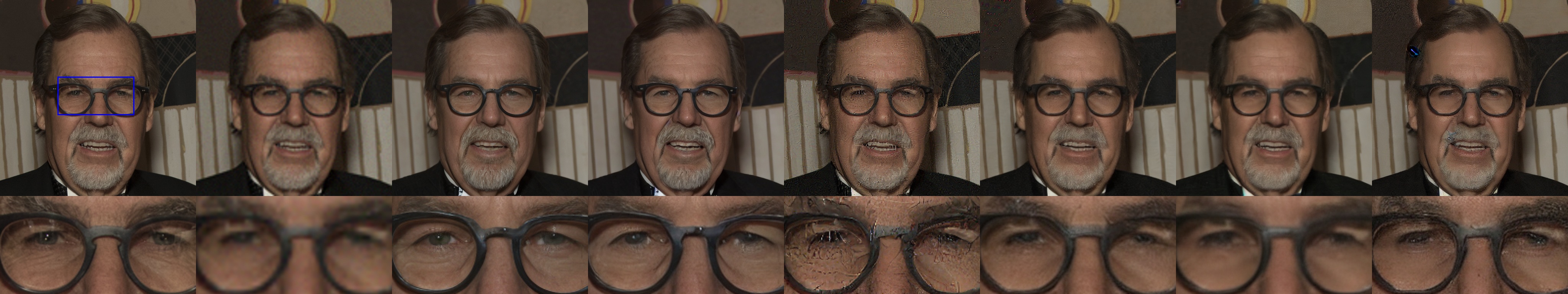}}} 

        \\
        \multicolumn{\columns}{c}{\centered{\includegraphics[width=\imgwidth \linewidth]{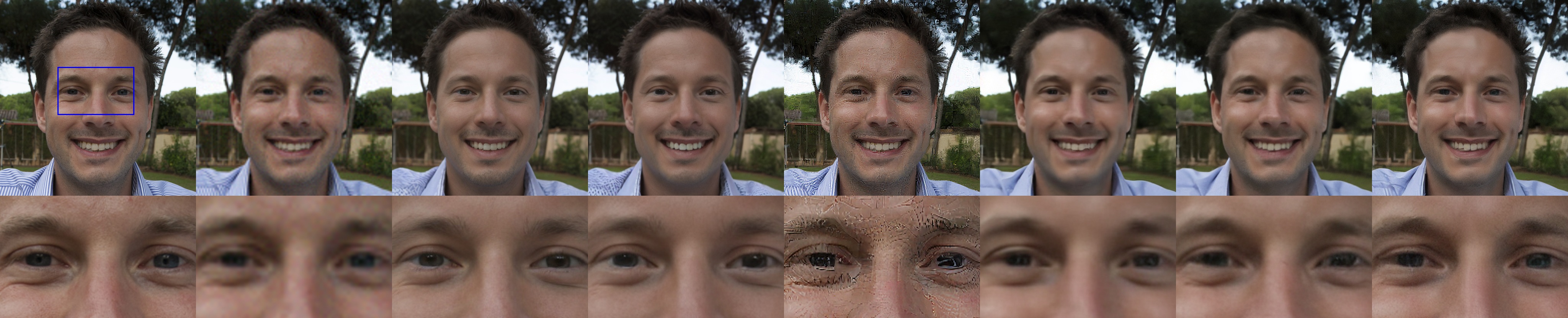}}} 
        \\
        \multicolumn{\columns}{c}{\centered{\includegraphics[width=\imgwidth \linewidth]{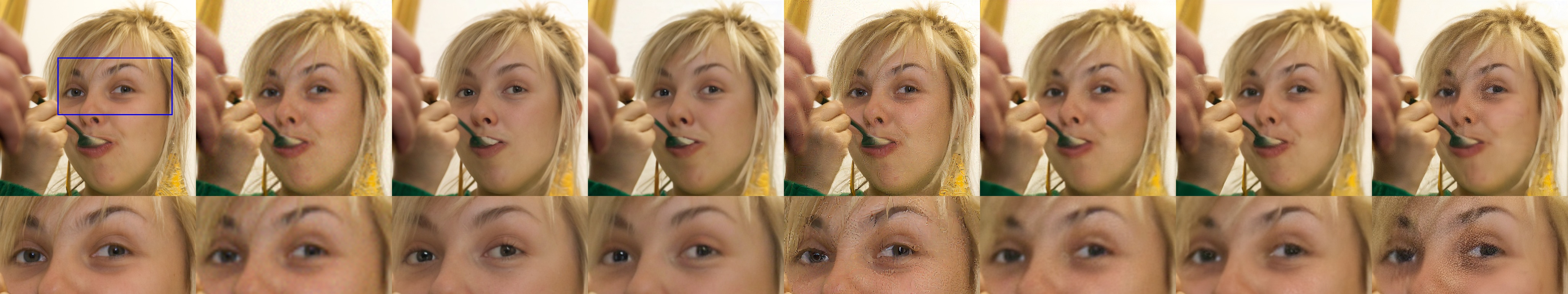}}}

    \end{tabular}
    \caption{SR${\times}4$ with $\sigma_y=0.01$, FFHQ dataset. Additional results.}
    \label{fig:app_sr4}
\end{figure*}
\endgroup

\begingroup
\newcolumntype{M}[1]{>{\centering\arraybackslash}m{#1}}
\newcommand{\vcentered}[1]{\begin{tabular}{@{}l@{}} #1 \end{tabular}}
\setlength{\tabcolsep}{0pt} %
\renewcommand{\arraystretch}{0} %

\def\columns{8}
\def\totalwidth{1}

\FPeval{\colwidth}{\totalwidth/\columns}
\FPeval{\imgwidth}{\totalwidth/\columns *\columns}

\FPeval{\colwidth}{clip(\totalwidth/\columns)}

\begin{figure*}[h!]
    \centering
    \begin{tabular}{*{\columns}{M{\colwidth\linewidth}}}

        \footnotesize{$x$} &
        \footnotesize{$y$} &
        \footnotesize{Ours (RV)} &
        \footnotesize{Ours (SD)} &
        \footnotesize{ReSample} &
        \footnotesize{PSLD} &
        \footnotesize{GML} &
        \footnotesize{LDPS} 
        \\
	    
        \rule{0pt}{0.8ex}\\

        \multicolumn{\columns}{c}{\centered{\includegraphics[width=\imgwidth\linewidth]{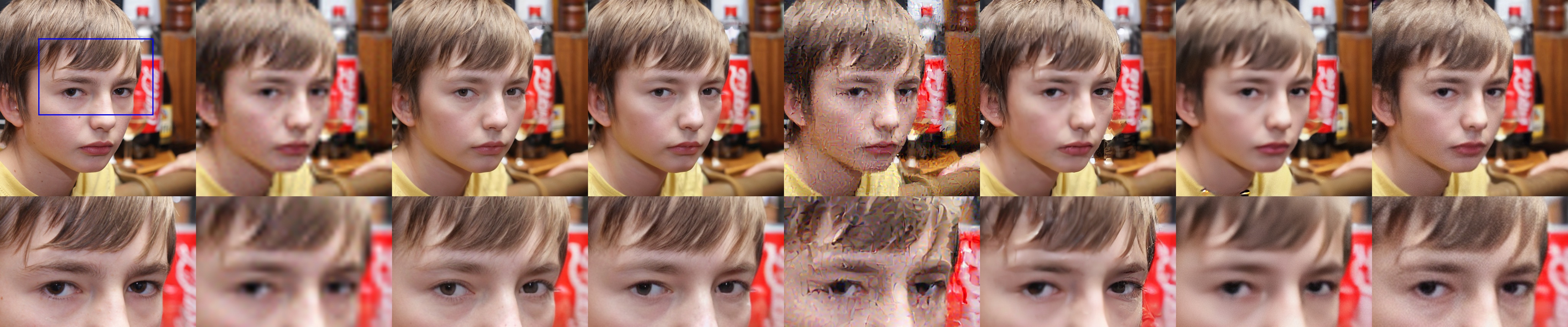}}} 
        \\
        \multicolumn{\columns}{c}{\centered{\includegraphics[width=\imgwidth\linewidth]{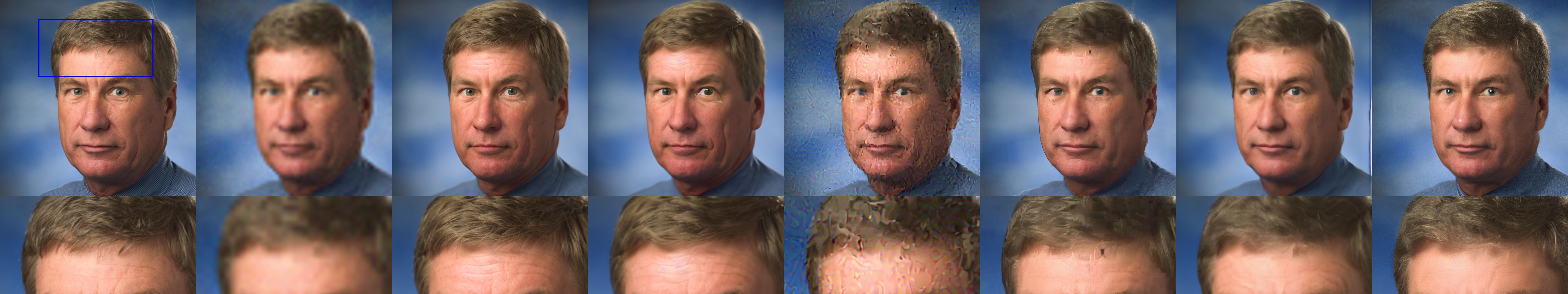}}} 
        \\
        \multicolumn{\columns}{c}{\centered{\includegraphics[width=\imgwidth \linewidth]{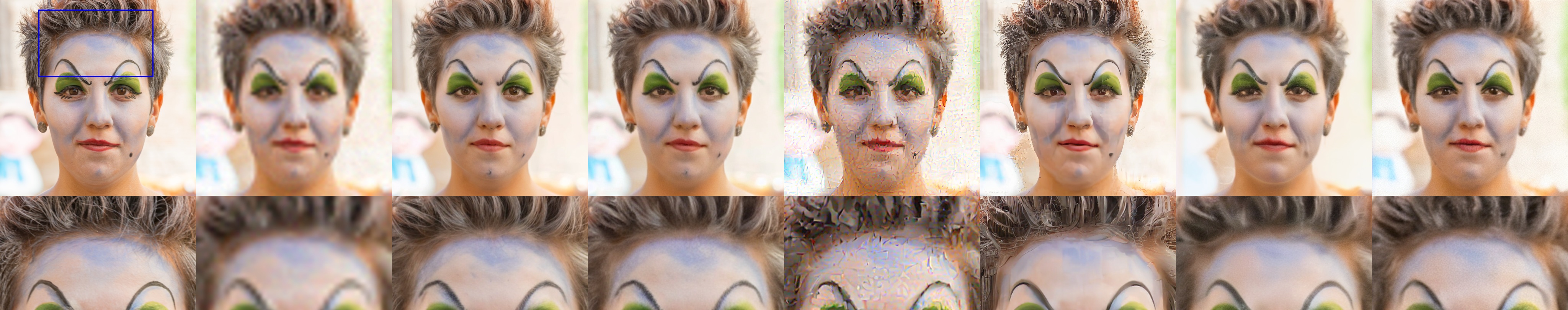}}} 
        \\
        \multicolumn{\columns}{c}{\centered{\includegraphics[width=\imgwidth \linewidth]{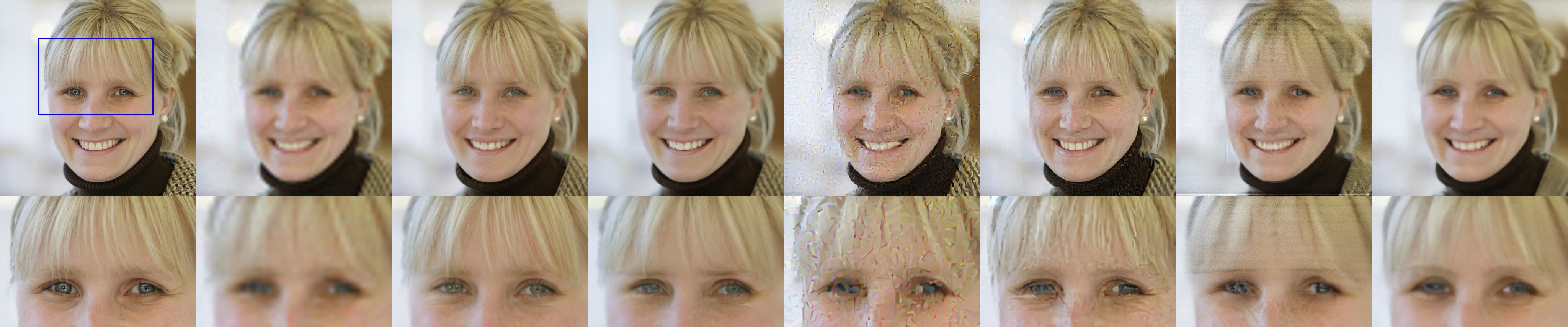}}} 

        \\
        \multicolumn{\columns}{c}{\centered{\includegraphics[width=\imgwidth \linewidth]{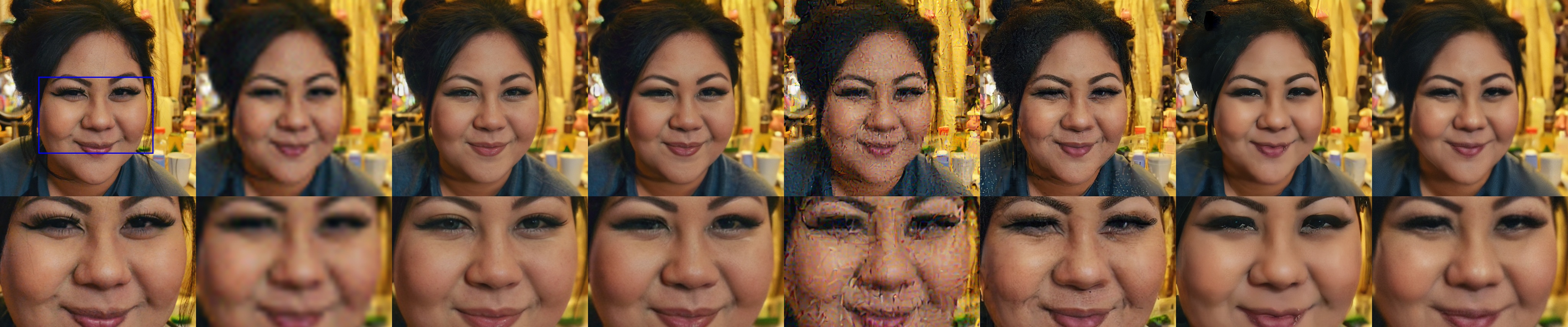}}} 
        \\
        \multicolumn{\columns}{c}{\centered{\includegraphics[width=\imgwidth \linewidth]{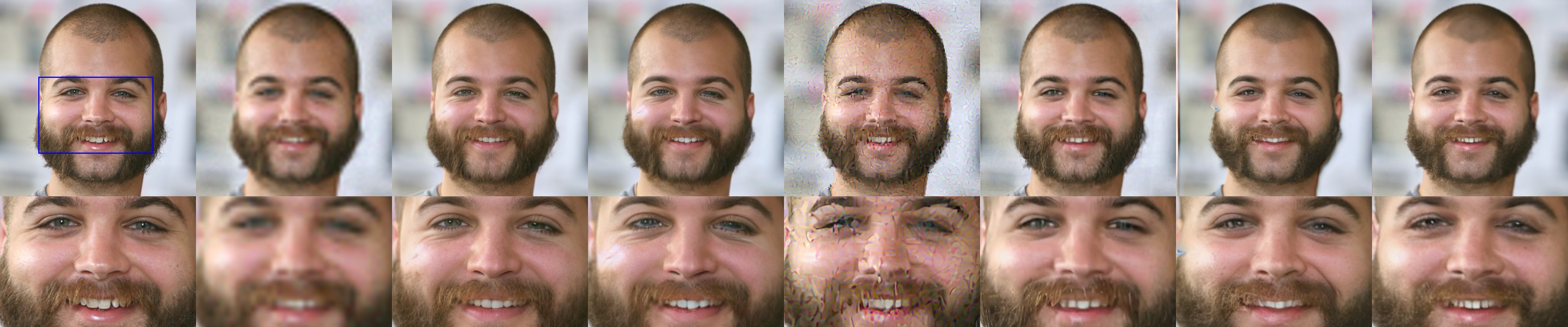}}}

    \end{tabular}
    \caption{SR${\times}8$, FFHQ dataset. Additional results.}
    \label{fig:app_sr8}
\end{figure*}
\endgroup

\begingroup
\newcolumntype{M}[1]{>{\centering\arraybackslash}m{#1}}
\newcommand{\vcentered}[1]{\begin{tabular}{@{}l@{}} #1 \end{tabular}}
\setlength{\tabcolsep}{0pt} %
\renewcommand{\arraystretch}{0} %

\def\columns{8}
\def\totalwidth{1}

\FPeval{\colwidth}{\totalwidth/\columns}
\FPeval{\imgwidth}{\totalwidth/\columns *\columns}

\FPeval{\colwidth}{clip(\totalwidth/\columns)}

\begin{figure*}[h!]
    \centering
    \begin{tabular}{*{\columns}{M{\colwidth\linewidth}}}

        \footnotesize{$x$} &
        \footnotesize{$y$} &
        \footnotesize{Ours (RV)} &
        \footnotesize{Ours (SD)} &
        \footnotesize{ReSample} &
        \footnotesize{PSLD} &
        \footnotesize{GML} &
        \footnotesize{LDPS} 
        \\
	    
        \rule{0pt}{0.8ex}\\

        \multicolumn{\columns}{c}{\centered{\includegraphics[width=\imgwidth\linewidth]{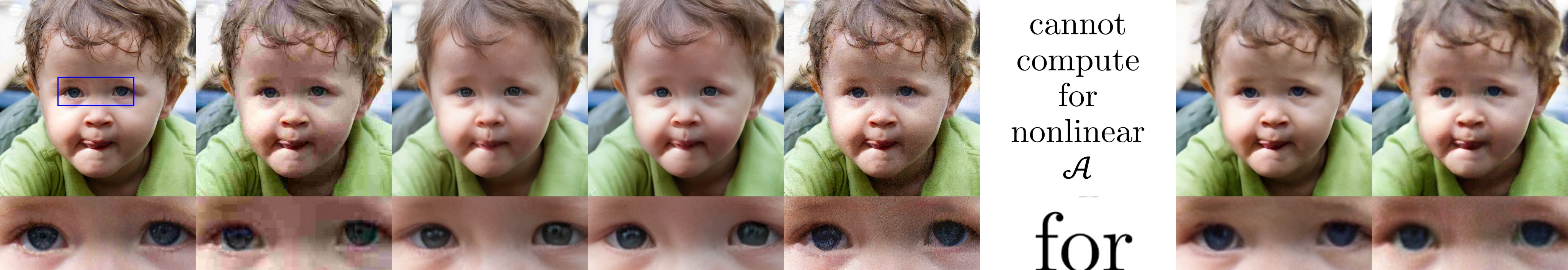}}} 
        \\
        \multicolumn{\columns}{c}{\centered{\includegraphics[width=\imgwidth\linewidth]{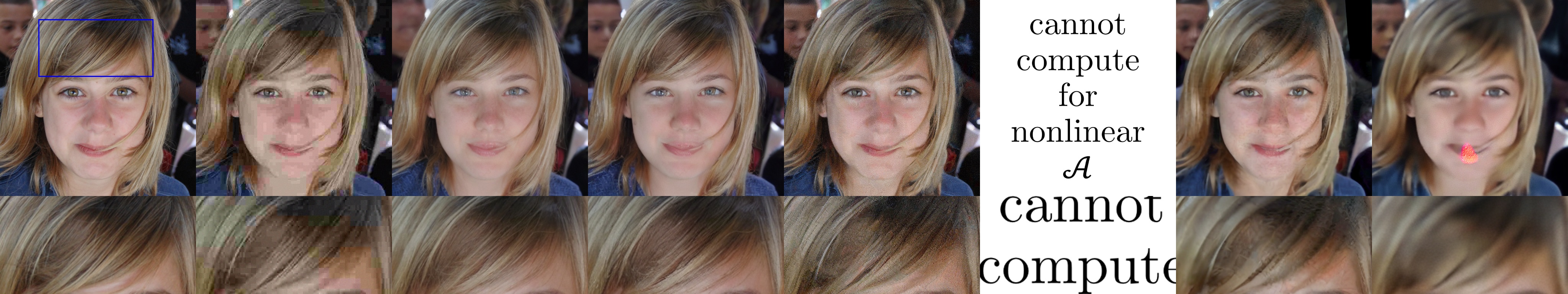}}} 
        \\
        \multicolumn{\columns}{c}{\centered{\includegraphics[width=\imgwidth \linewidth]{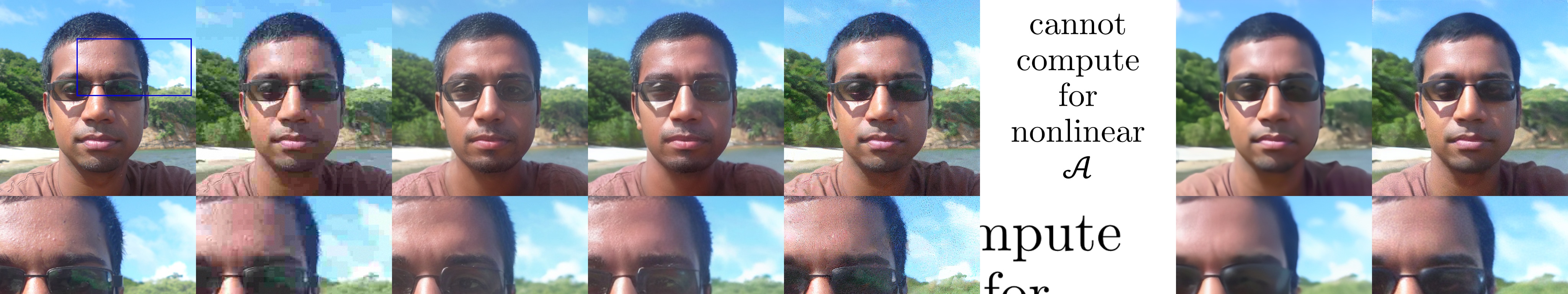}}} 
        \\
        \multicolumn{\columns}{c}{\centered{\includegraphics[width=\imgwidth \linewidth]{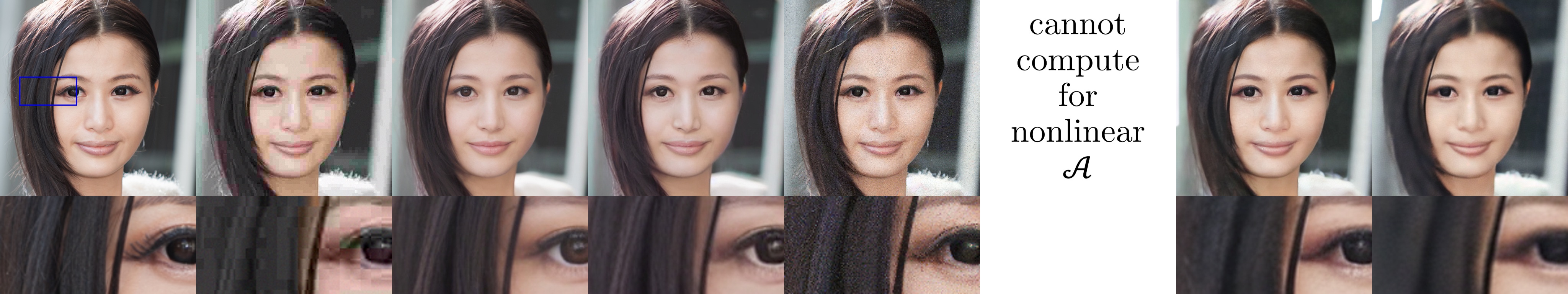}}} 

        \\
        \multicolumn{\columns}{c}{\centered{\includegraphics[width=\imgwidth \linewidth]{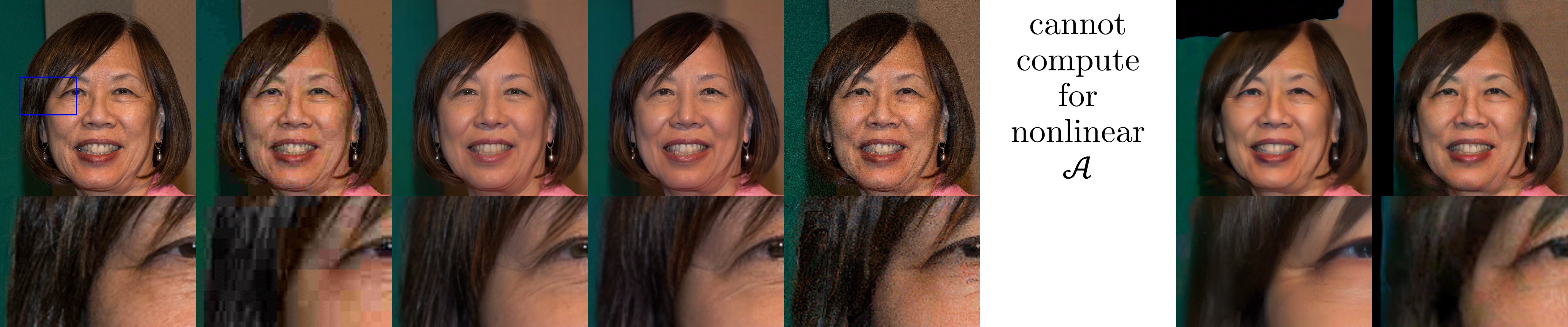}}} 
        \\
        \multicolumn{\columns}{c}{\centered{\includegraphics[width=\imgwidth \linewidth]{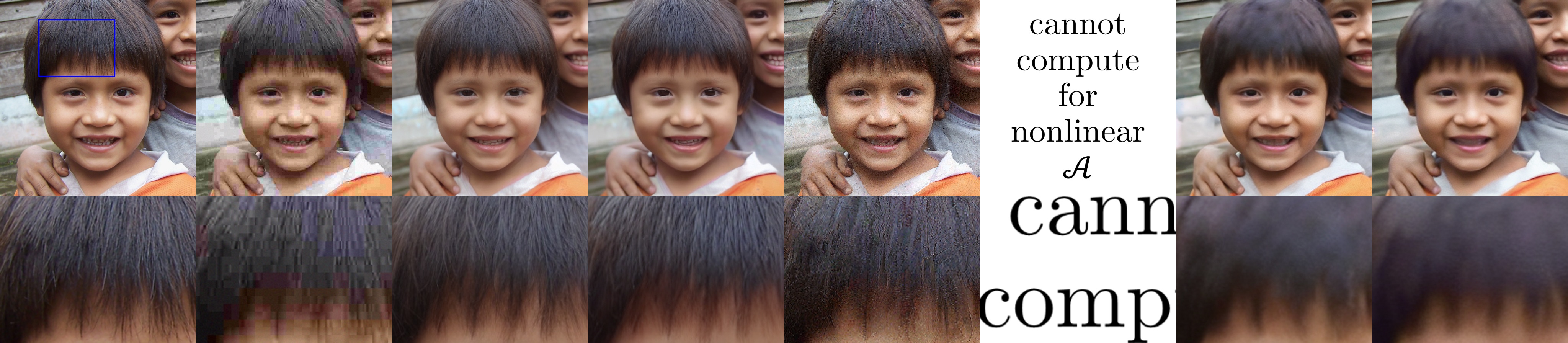}}}

    \end{tabular}
    \caption{JPEG with $\sigma_y=0.01$, FFHQ dataset. Additional results.}
    \label{fig:app_jpeg}
\end{figure*}
\endgroup

\begingroup
\newcolumntype{M}[1]{>{\centering\arraybackslash}m{#1}}
\newcommand{\vcentered}[1]{\begin{tabular}{@{}l@{}} #1 \end{tabular}}
\setlength{\tabcolsep}{0pt} %
\renewcommand{\arraystretch}{0} %

\def\columns{8}
\def\totalwidth{1}

\FPeval{\colwidth}{\totalwidth/\columns}
\FPeval{\imgwidth}{\totalwidth/\columns *\columns}

\FPeval{\colwidth}{clip(\totalwidth/\columns)}

\begin{figure*}[h!]
    \centering
    \begin{tabular}{*{\columns}{M{\colwidth\linewidth}}}

        \footnotesize{$x$} &
        \footnotesize{$y$} &
        \footnotesize{Ours (RV)} &
        \footnotesize{Ours (SD)} &
        \footnotesize{ReSample} &
        \footnotesize{PSLD} &
        \footnotesize{GML} &
        \footnotesize{LDPS} 
        \\
	    
        \rule{0pt}{0.8ex}\\

        \multicolumn{\columns}{c}{\centered{\includegraphics[width=\imgwidth\linewidth]{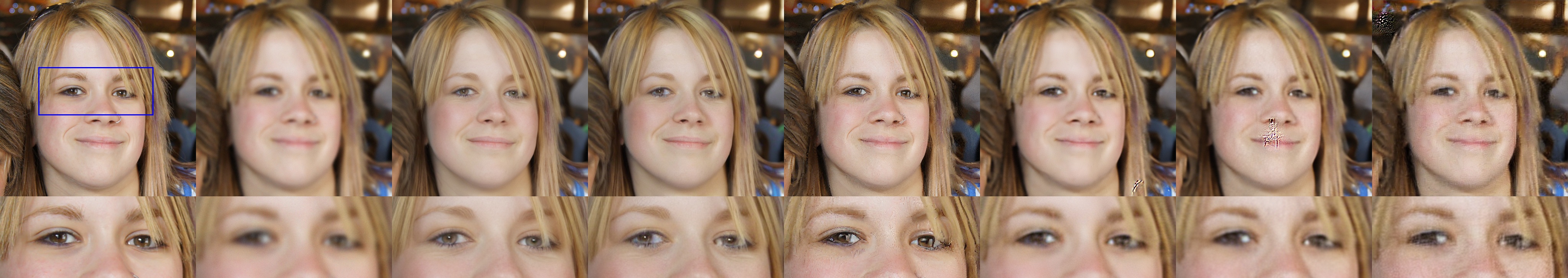}}} 
        \\
        \multicolumn{\columns}{c}{\centered{\includegraphics[width=\imgwidth\linewidth]{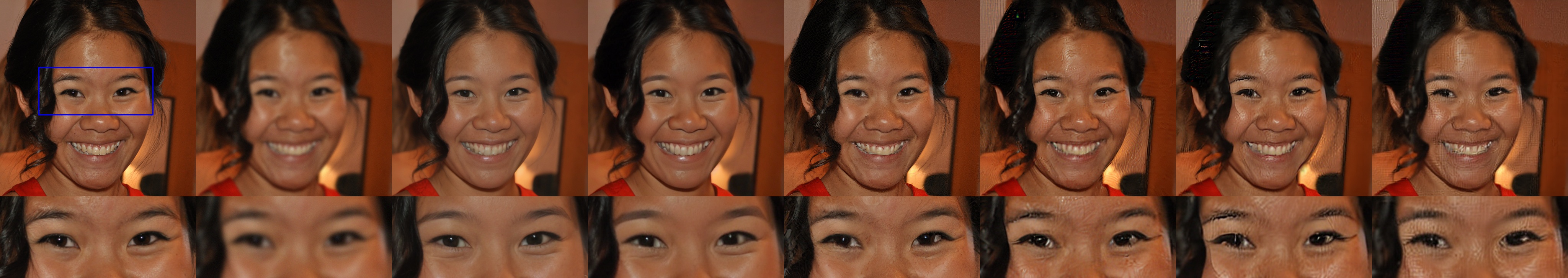}}} 
        \\
        \multicolumn{\columns}{c}{\centered{\includegraphics[width=\imgwidth \linewidth]{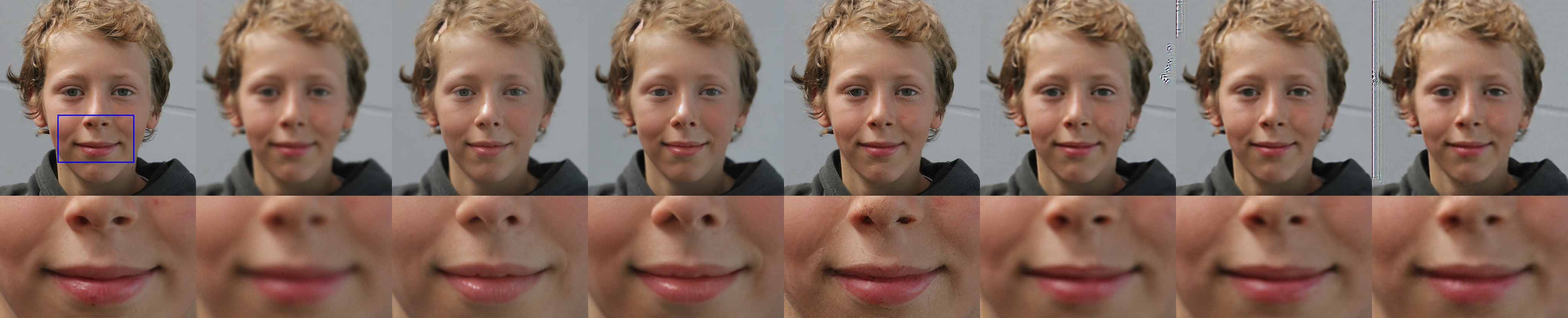}}} 
        \\
        \multicolumn{\columns}{c}{\centered{\includegraphics[width=\imgwidth \linewidth]{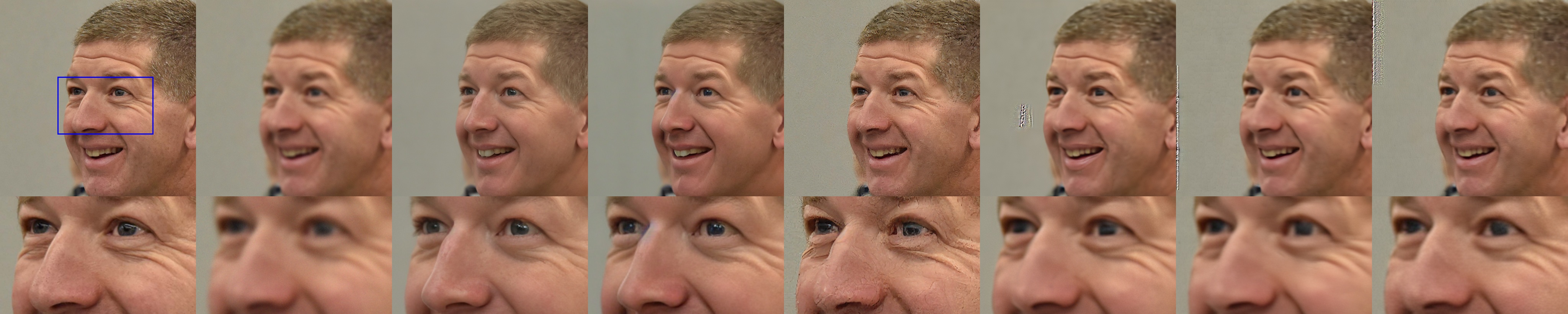}}} 

        \\
        \multicolumn{\columns}{c}{\centered{\includegraphics[width=\imgwidth \linewidth]{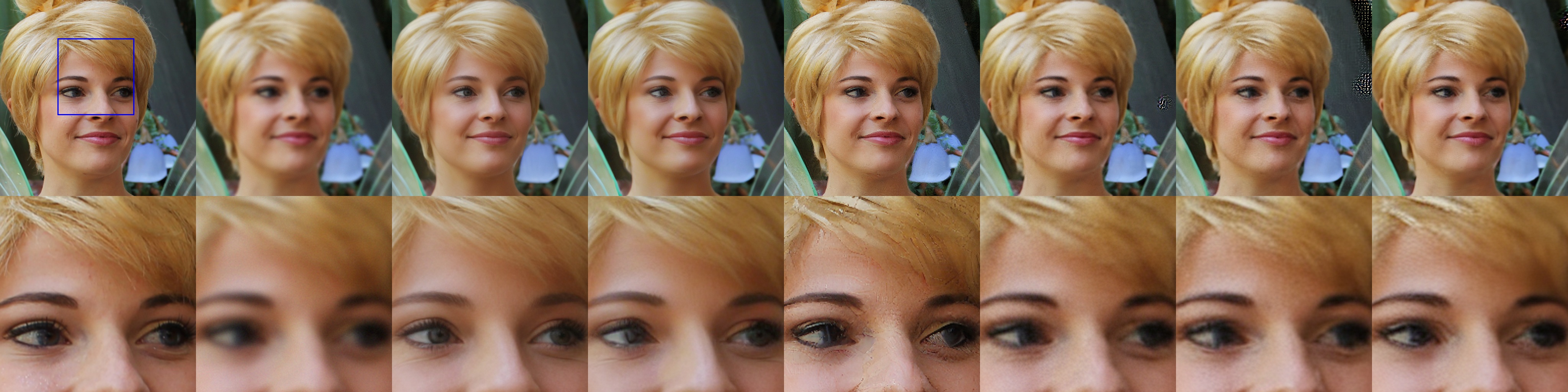}}} 
        \\
        \multicolumn{\columns}{c}{\centered{\includegraphics[width=\imgwidth \linewidth]{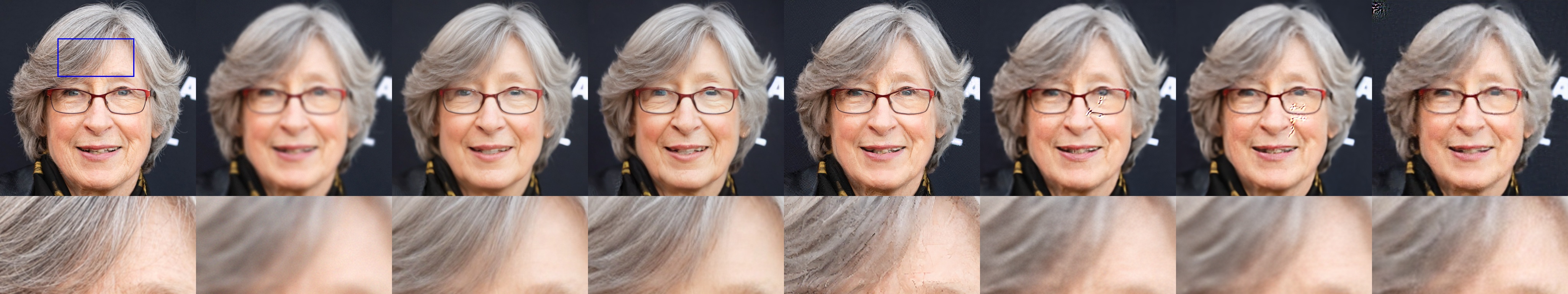}}}

    \end{tabular}
    \caption{Gaussian blur with $\sigma_y=0.01$, FFHQ dataset. Additional results.}
    \label{fig:app_gb}
\end{figure*}
\endgroup

\begingroup
\newcolumntype{M}[1]{>{\centering\arraybackslash}m{#1}}
\newcommand{\vcentered}[1]{\begin{tabular}{@{}l@{}} #1 \end{tabular}}
\setlength{\tabcolsep}{0pt} %
\renewcommand{\arraystretch}{0} %

\def\columns{8}
\def\totalwidth{1}

\FPeval{\colwidth}{\totalwidth/\columns}
\FPeval{\imgwidth}{\totalwidth/\columns *\columns}

\FPeval{\colwidth}{clip(\totalwidth/\columns)}

\begin{figure*}[h!]
    \centering
    \begin{tabular}{*{\columns}{M{\colwidth\linewidth}}}

        \footnotesize{$x$} &
        \footnotesize{$y$} &
        \footnotesize{Ours (RV)} &
        \footnotesize{Ours (SD)} &
        \footnotesize{ReSample} &
        \footnotesize{PSLD} &
        \footnotesize{GML} &
        \footnotesize{LDPS} 
        \\
	    
        \rule{0pt}{0.8ex}\\

        \multicolumn{\columns}{c}{\centered{\includegraphics[width=\imgwidth\linewidth]{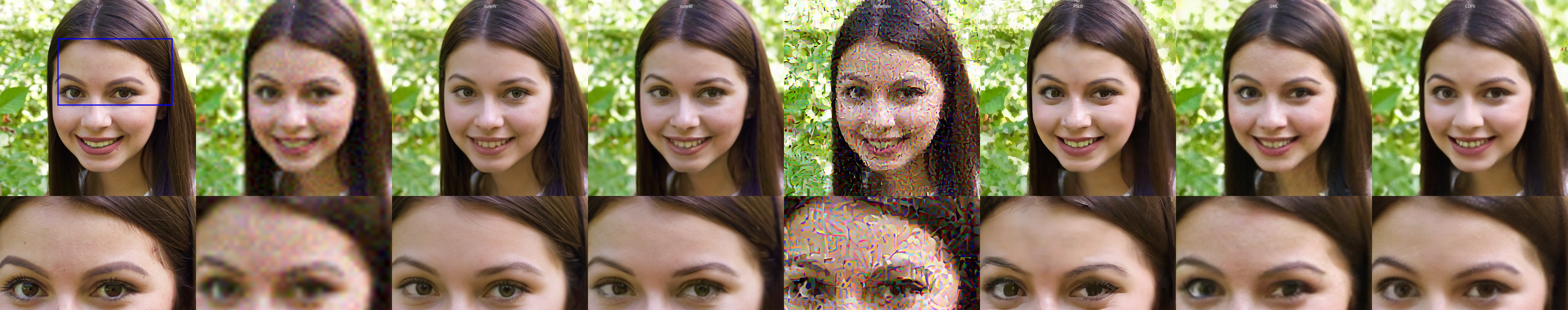}}} 
        \\
        \multicolumn{\columns}{c}{\centered{\includegraphics[width=\imgwidth\linewidth]{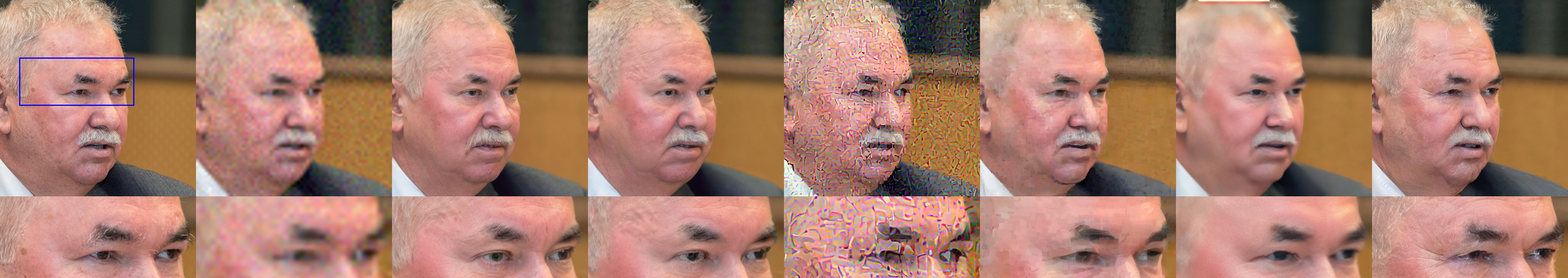}}} 
        \\
        \multicolumn{\columns}{c}{\centered{\includegraphics[width=\imgwidth \linewidth]{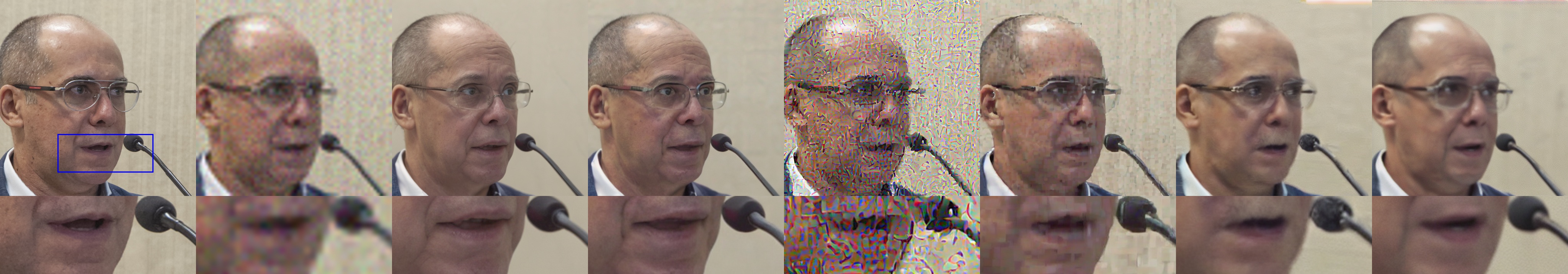}}} 
        \\
        \multicolumn{\columns}{c}{\centered{\includegraphics[width=\imgwidth \linewidth]{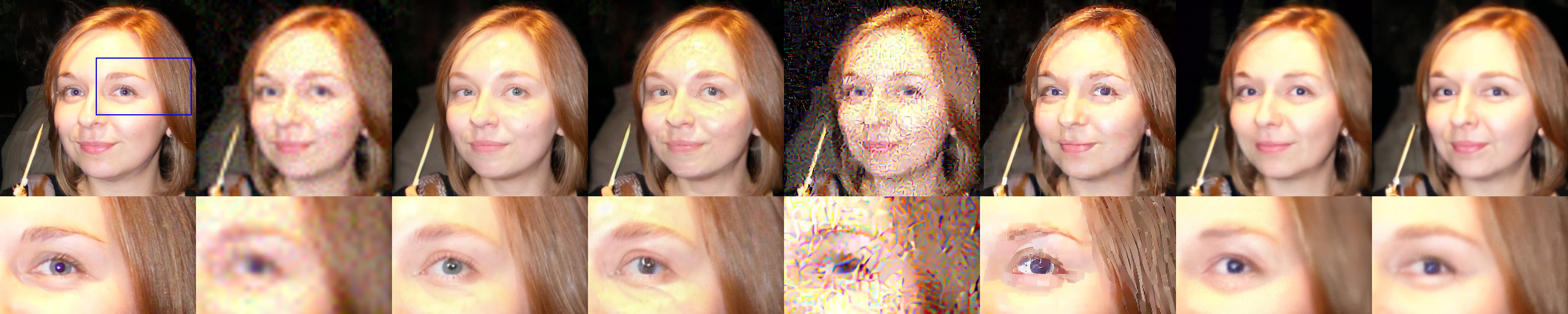}}} 

        \\
        \multicolumn{\columns}{c}{\centered{\includegraphics[width=\imgwidth \linewidth]{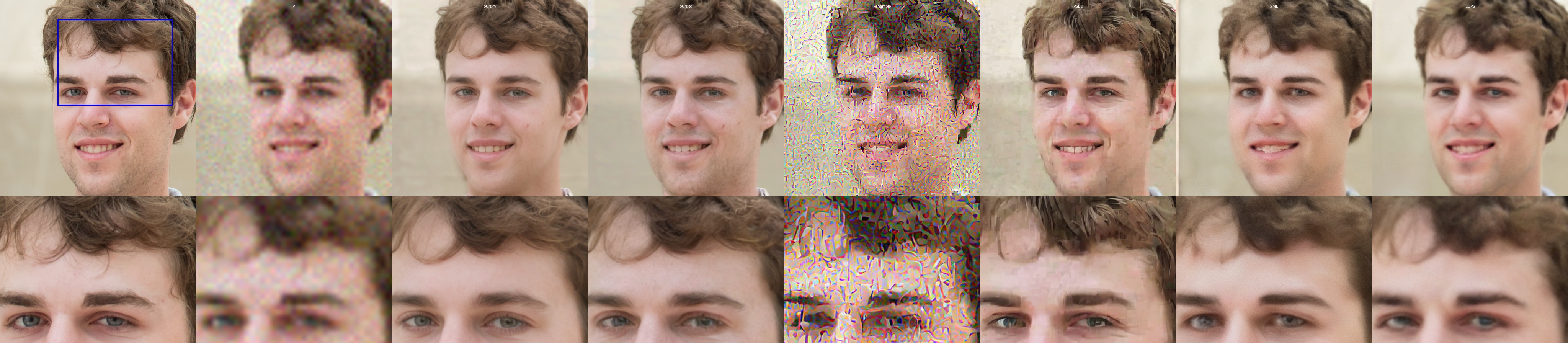}}} 
        \\
        \multicolumn{\columns}{c}{\centered{\includegraphics[width=\imgwidth \linewidth]{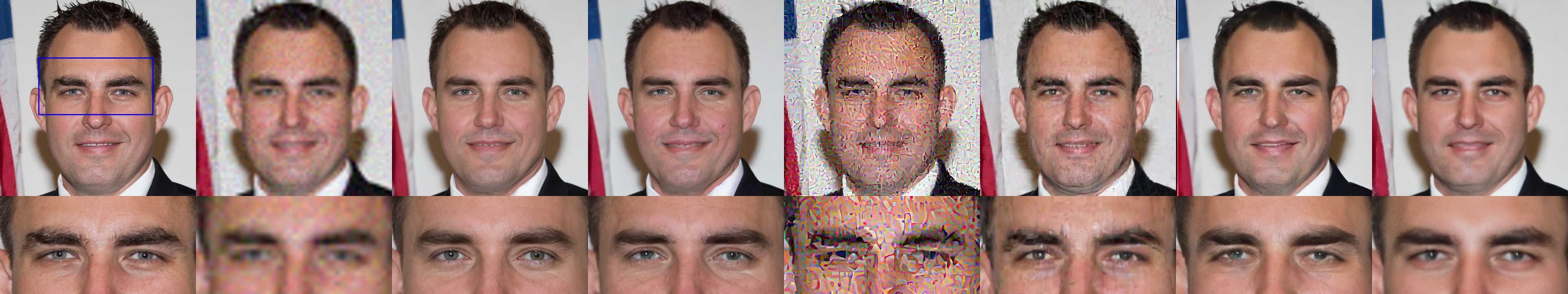}}}

    \end{tabular}
    \caption{SR${\times}8$ with $\sigma_y=0.03$, FFHQ dataset. Additional results.}
    \label{fig:app_sr8_noise}
\end{figure*}
\endgroup

\end{document}